%% file: sigcomm26_CR.tex
\documentclass[sigconf,9pt,authorversion]{acmart}

\usepackage[english]{babel}
\usepackage{blindtext}

\acmYear{2026}\copyrightyear{2026}
\setcopyright{cc}
\setcctype[4.0]{by}
\acmConference[SIGCOMM '26]{ACM SIGCOMM 2026 Conference}{August 17--21, 2026}{Denver, CO, USA}
\acmBooktitle{ACM SIGCOMM 2026 Conference (SIGCOMM '26), August 17--21, 2026, Denver, CO, USA}
\acmDOI{10.1145/3789240.3829148}
\acmISBN{979-8-4007-2467-1/26/08}

\usepackage{multirow}  

\usepackage{color}
\usepackage{tikz}
\usepackage[tight]{subfigure}
\usepackage{comment}
\usepackage{amsmath}
\usepackage{enumerate}
\usepackage{multirow}
\usepackage{hhline}

\usepackage{enumerate}
\usepackage{enumitem}

\usepackage{xurl}       

\usepackage{slashbox}

\usepackage{amsmath}
\usepackage{xspace}
\usepackage{algorithm}
\usepackage[font=small]{caption}
\usepackage[noend]{algpseudocode}
\usepackage[capitalize]{cleveref}
\makeatletter
\def\BState{\State\hskip-\ALG@thistlm}
\makeatother

\algdef{SE}[SUBALG]{Indent}{EndIndent}{}{\algorithmicend\ }%
\algtext*{Indent}
\algtext*{EndIndent}
\algrenewcommand\algorithmicindent{1.0em}%

\usepackage[normalem]{ulem}
\usepackage{xcolor}

\newif\ifcomm
\ifcomm
\else
\commfalse
\fi
\ifcomm
	\newcommand{\mycomm}[3]{{\footnotesize{{\color{#2} \textbf{[#1: #3]}}}}}
    \newcommand{\revise}[1]{\textcolor{blue}{#1}}
    \newcommand{\remove}[1]{{\color{orange}\sout{#1}}}
\else
    \newcommand{\mycomm}[3]{}
    \newcommand{\revise}[1]{#1}
    \newcommand{\remove}[1]{}
\fi

\newcommand{\ran}[1]{\mycomm{Ran}{blue}{#1}}

\newcommand{\Wenchen}[1]{\mycomm{Wenchen}{Wenchen}{#1}}

\newcommand{\norm}[1]{\left\lVert#1\right\rVert}

\newcommand{\floor}[1]{\left\lfloor#1\right\rfloor}

\newcommand{\parentheses}[1]{\left(#1\right)}

\newcommand{\set}[1]{\left\{#1\right\}}

\newcommand{\Var}[1]{\mathrm{Var}{#1}}

\definecolor{Wenchen}{RGB}{200,0,200}

\hypersetup{pdfstartview=FitH,pdfpagelayout=SinglePage}

\input{wenchen_macros}

\newcommand{\sysname}{\textsc{DynamiQ}\xspace}
\newcommand{\sys}{\sysname}

\renewcommand{\eg}{e.g.}

\begin{document}

\title{DynamiQ: Accelerating Gradient Synchronization using Compressed Multi-hop All-reduce}

\author{Wenchen Han}
\affiliation{%
  \institution{University College London}%
  \country{}%
}

\author{Shay Vargaftik} 
\affiliation{%
  \institution{VMware Research by Broadcom}%
  \country{}%
}

\author{Michael Mitzenmacher}
\affiliation{%
  \institution{Harvard University}%
  \country{}%
}


\author{Ran Ben Basat}
\affiliation{%
  \institution{University College London and Broadcom}%
  \country{}%
}

\ccsdesc[500]{Computer systems organization~Distributed architectures}
\ccsdesc[500]{Computing methodologies~Machine learning}

\keywords{Distributed training, all-reduce, gradient compression, quantization, algorithms, data-parallelism, large language models}

\begin{abstract}
Multi-hop all-reduce is the de facto backbone of large model training. As the training scale increases, the network often becomes a bottleneck, motivating the reduction of the volume of transmitted data. 
Accordingly, recent systems have demonstrated significant acceleration of the training process using gradient quantization.
However, these systems are not optimized for multi-hop aggregation, where entries are partially summed multiple times \mbox{along their aggregation topology.}

We present DynamiQ, a quantization framework that bridges the gap between quantization best practices and multi-hop aggregation. DynamiQ introduces novel techniques to better represent partial sums, co-designed with a decompress‑accumulate‑recompress fused kernel to facilitate fast execution.

%
We extend PyTorch DDP to support DynamiQ over NCCL P2P, and across different LLMs, tasks, and scales, we demonstrate consistent improvement of up to 34.2\% over the best among state-of-the-art methods such as Omni-Reduce, THC, and emerging standards such as MXFP4, MXFP6, and MXFP8. Further, DynamiQ is the only evaluated method that consistently reaches near-baseline accuracy (e.g., 99.9\% of the BF16 baseline) and does so while significantly accelerating the training.



\end{abstract}

\maketitle


\vspace{1cm}
\section{Introduction}

Distributed data parallel (DDP)~\cite{NIPS2012_6aca9700} is the standard paradigm for large language model (LLM) training and fine-tuning. Under this paradigm, the model is replicated across workers, each processing a different part of the data to compute a local gradient. These gradients are then synchronized (aggregated) via the network to obtain a global update. 
Gradient aggregation in LLM training commonly relies on multi-hop all-reduce schemes~\cite{li2020pytorch,collective-nccl,jax}, such as ring~\cite{ring} and butterfly~\cite{thakur2005optimization}.
With the growth of model sizes and the number of workers, gradient aggregation increasingly becomes a bottleneck~\cite{sapio2021scaling, desensi2024swing, tang2025dreamddp, hotnets2499problems, warraich2025optireduce,warraich2025optinic}. 
Recent practices of running multiple jobs in the same cluster, where jobs compete on network resources~\cite{cao2024crux, hwang2021elastic}, further intensify the bottleneck.




Gradient compression, which aims to reduce the volume of communicated gradient data, is therefore a natural and promising approach to accelerating gradient aggregation. Despite substantial prior work, we observe that state-of-the-art solutions~\cite{peng2023fp8, li2024thc, fei2021efficient, bai2021gradient, wang2023hi, wang2023cupcake} typically consider the (single-hop) parameter-server architecture~\cite{li2014scaling} where aggregation (after decompression) can be performed with higher precision without any bandwidth implications. In particular, they are not optimized for \textit{multihop} all-reduce, where gradients are \textit{partially summed} along their aggregation topology. In such a scheme, intermediate nodes face the choice of either recompressing the partial sum, thus degrading accuracy and eventually model performance, or increasing the number of bits used for its representation, leading to limited end-to-end speedups~\cite{han24hotnets,on-the-utility}. 
As we show in \Cref{sec:eval}, this limitation applies both to existing quantization and sparsification schemes (e.g., THC~\cite{li2024thc} and OmniReduce~\cite{fei2021efficient}), as well as recent microscaling \mbox{floating-point (FP) formats~\cite{peng2023fp8,rouhani2023microscaling,opencompute}.}

In this paper, we \revise{focus on data-parallelism and} introduce \sys, a compression framework \emph{tailored for multi-hop all-reduce}
\revise{
 and is applicable to both pre-training and fine-tuning.} 
\sys minimizes the compression error of partial sums under a bandwidth constraint, utilizing a fused decompress–
accumulate–recompress kernel to minimize memory bandwidth~\cite{memory-bound, memory-bound2, wang2025optimizing} and facilitate the overlap of compression with communication. 
The key to \sys's superior accuracy-bandwidth tradeoff lies in its two-phase method of quantizing different coordinates with different numbers of bits, depending on their magnitude \emph{in the aggregated gradient}.

\setlist[itemize]{
  leftmargin=1.5em,
  labelsep=0.3em
}
\setlist[enumerate]{
  leftmargin=1.5em,
  labelsep=0.3em
}

Ideally, we would quantize each coordinate with a number of bits based on its magnitude. \mbox{However, this imposes several challenges:}
\begin{itemize}
\item Communicating each entry’s quantized bit width would add prohibitive overhead.
\item Arbitrary quantized widths can break byte alignment, preventing efficient fused kernels.
\item Aperiodic widths harm memory coalescing. Moreover, offset metadata also significantly increases the memory and bandwidth overheads.
\item Varying the bit allocation along the aggregation path requires repacking and hinders performance, as this cannot be done efficiently in a fused kernel.
\end{itemize}



Instead, our framework works as follows.
\sys joins consecutive entries into small \emph{groups} (e.g., 16 entries), which are then further formed into \emph{super-groups} (e.g., 16 groups). Groups and super-groups share metadata; groups have a shared scale parameter, and all entries in a super-group use the same bitwidth. This approach provides a good balance between bit allocation flexibility (from varying the group size) and metadata overhead. 
\sys performs an initial lightweight all-reduce call to collect necessary statistics about super-groups. This enables all workers to agree on the bit allocation, which then remains fixed throughout the aggregation. 
Further, all workers then reorder the super-groups according to their bit allocation, which enables fused kernel invocations on sequential data in the main all-reduce. 

To further optimize the bandwidth-accuracy tradeoff, \sys uses advanced quantization techniques, including:
\begin{itemize}[leftmargin=*]
    \item \emph{Non-uniform quantization} —
    \sys normalizes the data and uses a pre-determined non-uniform set of quantization values that optimizes the per-entry multiplicative error. 
    Intuitively, this is achieved by using more quantization values that are closer to zero than larger ones, akin to floating-point formats. Specifically, we use the choice \mbox{of quantization values proposed by~\cite{mee}.}
    \item \emph{Negative correlation across workers} — \sys uses correlated rounding so the errors are more likely cancel out~\cite{suresh2022correlated}. Intuitively, \sys uses shared randomness to increase the probability that if one worker rounds upwards, another will round down, lowering the aggregation error.
\end{itemize}


We integrate \sys into PyTorch DDP~\cite{li2020pytorch} via a communication hook that runs over NCCL P2P~\cite{nccl}.  We evaluate \sys across different LLM fine-tuning workloads (BERT-large~\cite{devlin2018bert} Masked LM, LLaMA-1B~\cite{grattafiori2024llama3herdmodels} chat \& MMLU~\cite{hendrycks2020measuring}, Gemma-1B~\cite{team2025gemma} Chat) and all-reduce topologies (ring and butterfly). \sys improves the time-to-accuracy by up to 34.2\% over the best among OmniReduce~\cite{fei2021efficient}, THC~\cite{li2024thc} and modern floating point (FP) formats (MXFP4/MXFP6/MXFP8)~\cite{opencompute}. In several settings, \sys is the only method that reaches near-baseline accuracy (\ie, $99.9\%$ final accuracy relative to BF16) in all our workloads while accelerating training by 40.8\% compared with BF16. 
\revise{In simulated setups, we show that \sysname scales well to large DP-dimensions. For example, with a DP-dimension of $8192$, \sysname with $\bar{b}=6$ bits per coordinate is more accurate than MXFP8 ($\bar{b}=8.5$ per coordinate) by a large margin.}
\remove{We plan to open-source our code upon the publication of the work.}
\revise{Our code is available at~\cite{open-source}.}

\vspace{-0.1cm}
\section{Background}
\vspace{-0.05cm}

In this section, we provide necessary background on quantization. Quantization is the process of mapping a continuous or high-precision set of values to a smaller, discrete set of values, essentially reducing the number of bits used to represent a number. In the context of distributed gradient-based training frameworks (e.g., using a distributed SGD, ADAM~\cite{adam2014method}, or AdamW~\cite{loshchilov2017decoupled} for LLMs), at each training round, gradients from different workers must be aggregated to compute a global gradient. Thus, employing gradient quantization at the workers reduces communication. The challenge, however, is to make the quantization accurate and fast, so that it takes less time to reach the desired model accuracy, a metric known to \emph{time-to-accuracy}.


\vspace{-0.1cm}
\subsection{Unbiased quantization}
\vspace{-0.05cm}

An important property of gradient quantization is being \emph{unbiased}. Namely, given a gradient $X \in \mathbb{R}^d$ and its quantized estimate $\widehat 
X$, we would like to have $\mathbb{E}[\widehat X] = X$.

Intuitively, unbiasedness is desirable when averaging, as when some values are rounded up and others are rounded down, the errors cancel out in expectation.
Critically, under mild conditions, unbiased quantization ensures training convergence~\cite{alistarh2017qsgd}.

A fundamental method for unbiased quantization is \emph{stochastic quantization} (SQ). In SQ, given a scalar $x\in\mathbb R$ and two \emph{quantization values} $x_\downarrow, x_\uparrow$ where $x \in [x_\downarrow, x_\uparrow]$, we obtain $\mathbb{E}[\widehat x] = x$ by setting
$$\widehat{x} =
\begin{cases}
    x_{\uparrow}   & \text{w.p. } \dfrac{x - x_{\downarrow}}{\,x_{\uparrow}-x_{\downarrow}\,}\\
    x_{\downarrow} & \text{otherwise}
\end{cases}.$$ 

More generally, when quantizing a vector $X\in \mathbb{R}^d$ using a set of quantization values $Q\subset \mathbb{R}^d$, for each $x\in X$, we denote by $x_\downarrow = \max\set{q\in Q\mid q\le x}$ and $x_\uparrow = \min\set{q\in Q\mid q\ge x}$ its two closest values, and apply SQ as described above. Thus, each quantized value $\widehat x\in Q$ can be represented using $\log_2 |Q|$ bits using its quantization value index \mbox{(e.g., for 4 bits per coordinate, we can use $|Q| = 16$).}

\vspace{-0.1cm}
\subsection{Grouped quantization}\label{subsec:background-grouped-quantization}
\vspace{-0.05cm}
To dequantize a quantized vector, the receiver must know the set $Q$. While many previous works select a single $Q$ for an entire all-reduce call, we opt to choose one set for each \emph{group} (a consecutive sequence of e.g., 16 entries) as it allows \mbox{optimizing $Q$ for the specific entries. }

There are two main reasons why a per-group choice of $Q$ is more accurate. First, gradients often exhibit \emph{spatial locality}, where nearby entries tend to have similar magnitudes. Grouping, therefore, tailors $Q$ to a small range of values. Second, the gradient distribution itself is skewed, with a small number of coordinates (outliers) that can be orders of magnitude larger than others. These outliers have a disproportionate effect on the set $Q$ and the effectiveness of quantization.
Using per-group values $Q$ with small groups reduces the overall effect of these outliers. 

Intuitively, due to the above reasons, the smaller the groups are, the more accurate the quantization is, as we tailor $Q$ (and the size of $Q$) for each specific group.
However, since each group has an overhead (i.e., the encoding of its $Q$), having too many small groups inflates the required bandwidth, defeating the purpose of quantization.
To reduce the encoding overheads, one can use super-groups (e.g., of 16 consecutive groups each) \mbox{to share some of the metadata~\cite{GGUF}. }

\begin{figure}
    \centering

        \hspace{-0.2cm}
        \subfigure[LLaMA, group size 16]{
		\begin{minipage}[t]{0.48\linewidth}{
		\vspace{-0.00in}
		\begin{center}
		\includegraphics[width=\textwidth, ]{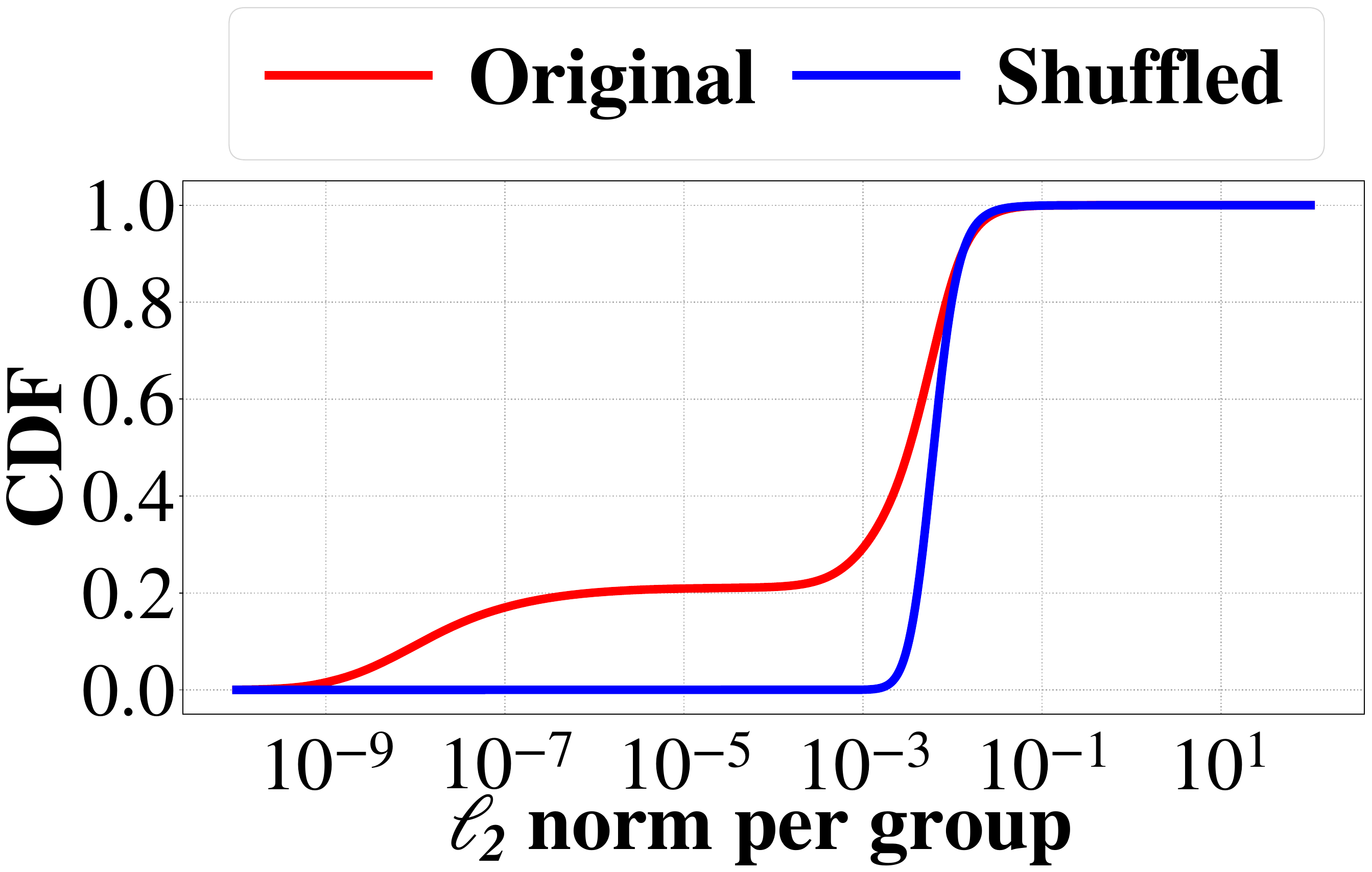}
		\end{center}
		}
		\label{subfig:demo-original-CDF-per-group-vs-shuffled}
		\end{minipage}
	}
    \hspace{-0.1cm}
        \subfigure[Gemma, group size 16]{
		\begin{minipage}[t]{0.48\linewidth}{
		\vspace{-0.00in}
		\begin{center}
		\includegraphics[width=\textwidth, ]{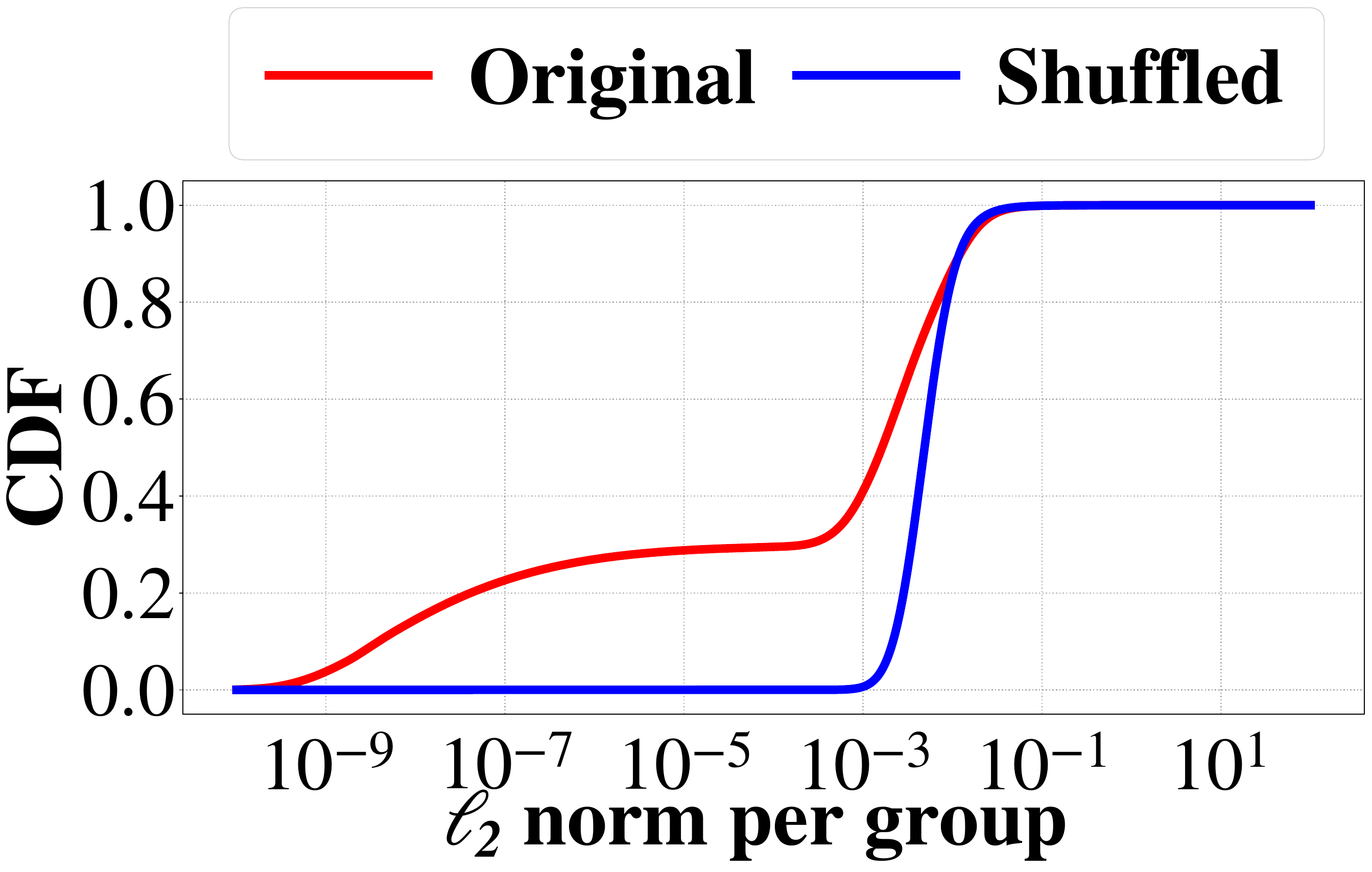}
		\end{center}
		}
		\label{subfig:demo-original-CDF-per-group-gemma}
		\end{minipage}
	}
    
    \hspace{-0.2cm}
        \subfigure[LLaMA, super-group size  256]{
		\begin{minipage}[t]{0.48\linewidth}{
		\vspace{-0.00in}
		\begin{center}
		\includegraphics[width=\textwidth, ]{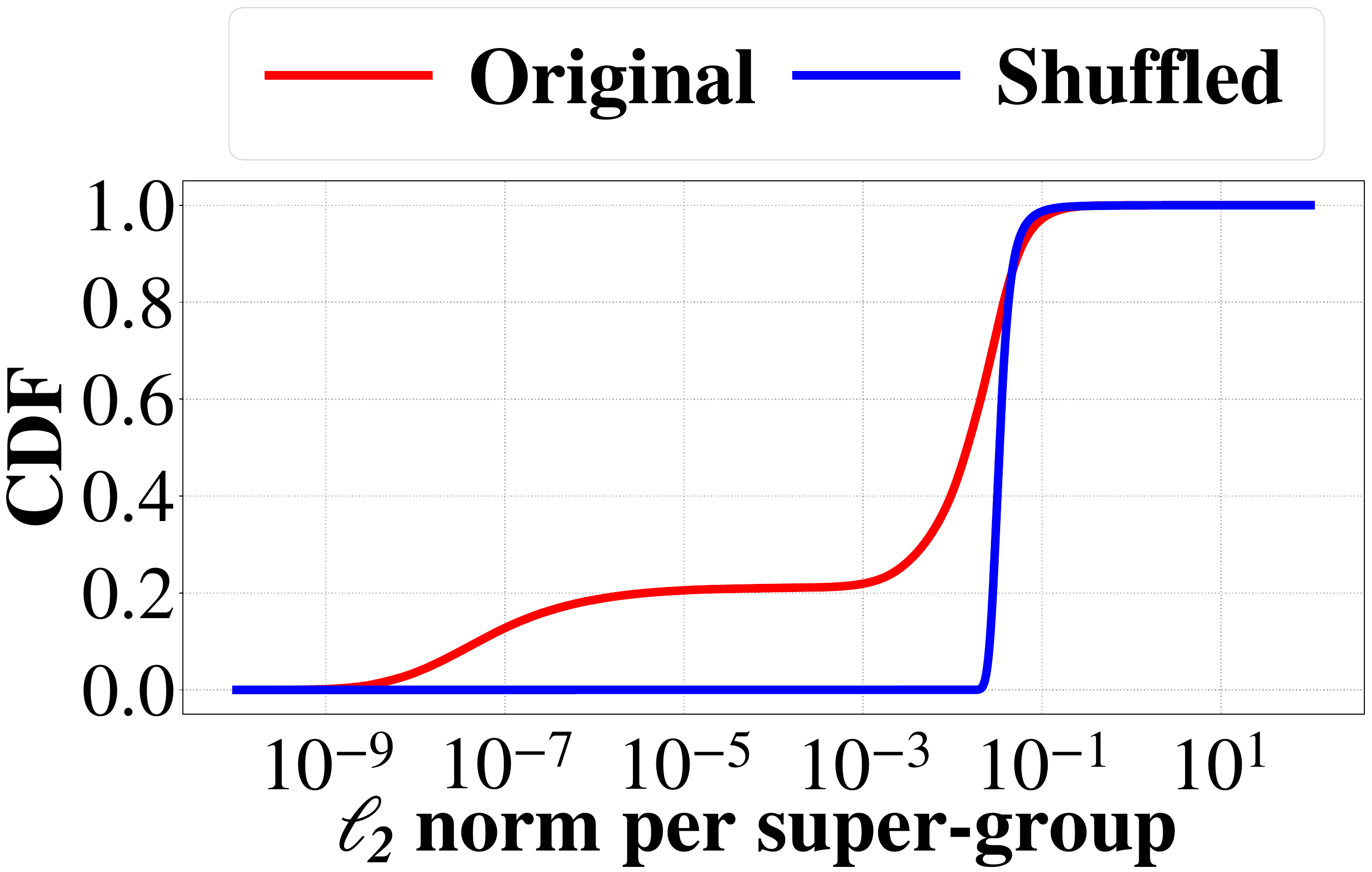}
		\end{center}
		}
		\label{subfig:demo-original-CDF-per-supergroup-vs-shuffled}
		\end{minipage}
	}
    \hspace{-0.1cm}
        \subfigure[Gemma, super-group size 256]{
		\begin{minipage}[t]{0.48\linewidth}{
		\vspace{-0.00in}
		\begin{center}
		\includegraphics[width=\textwidth, ]{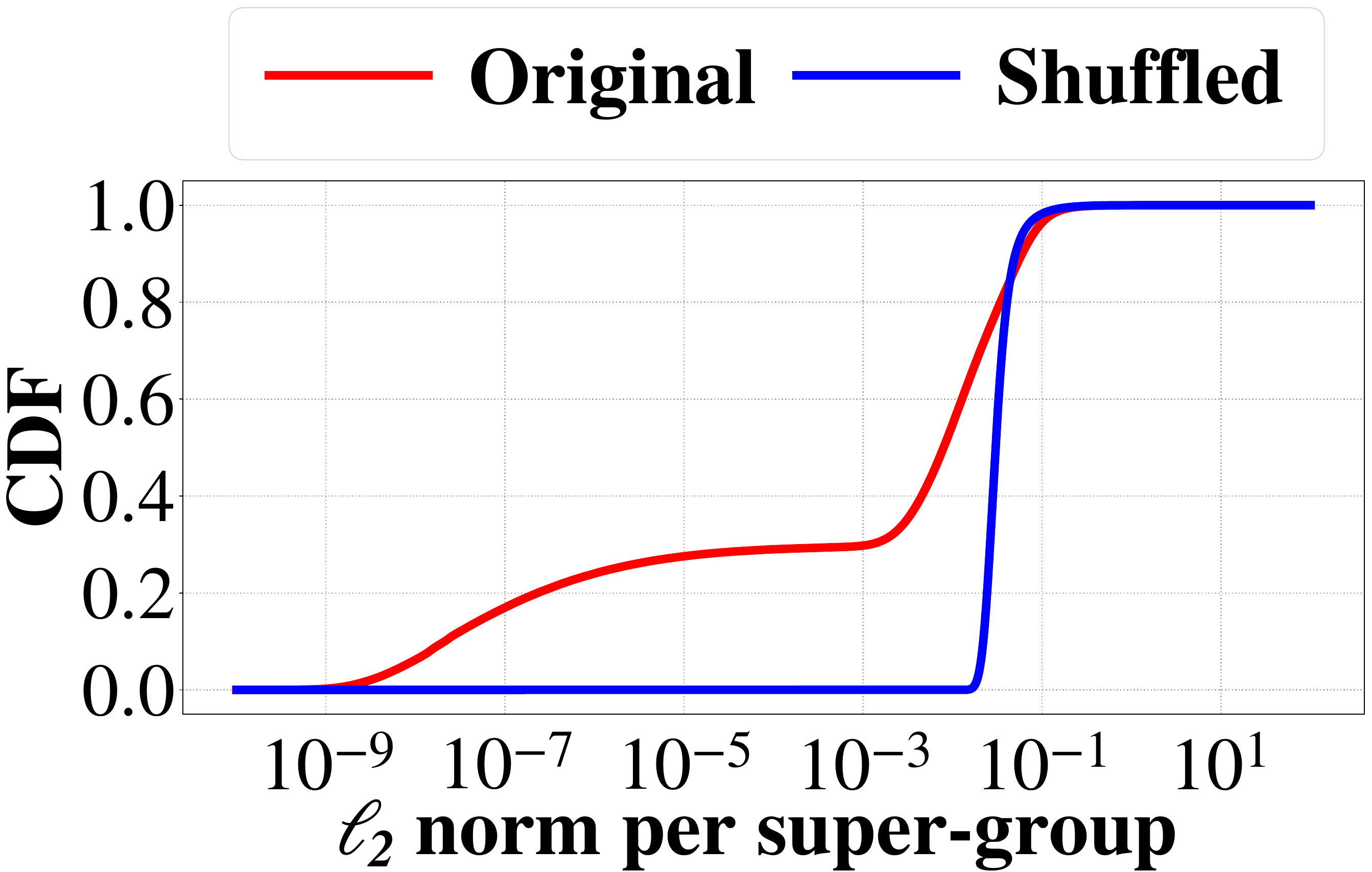}
		\end{center}
		}
		\label{subfig:demo-original-CDF-per-supergroup-gemma}
		\end{minipage}
	}

    \vspace{-0.1cm}
    \caption{$\ell_2$ norm distributions of the gradients and their random shuffle 
    for groups of size $16$ and super-groups of size $256$. The detailed experimental setups appear in Section~\ref{sec:eval}.}
    \label{fig:demo-per-group-vs-shuffled}
\end{figure}

We conduct an experiment to exemplify spatial locality and skewness. To do so, we compare the distribution of group and super-group norms in the original gradient to these distributions after randomly shuffling the entries. Intuitively, if there is no spatial locality, the distributions are likely to be similar.
We analyze the first gradient of fine-tuning LLaMA 1B for MMLU~\cite{hendrycks2020measuring} and Gemma 1B for Ultrachat~\cite{ding2023enhancing} (the complete experimental details appear in~\Cref{sec:eval}). We show the results for groups of size 16 in Figures~\ref{subfig:demo-original-CDF-per-group-vs-shuffled}-\ref{subfig:demo-original-CDF-per-group-gemma} and for a sequence of 16 groups (i.e., super-groups) in Figures~\ref{subfig:demo-original-CDF-per-supergroup-vs-shuffled}-\ref{subfig:demo-original-CDF-per-supergroup-gemma}. 
This spatial locality yields a significant fraction (e.g., about 20\% in LLaMA and 30\% in Gemma) of super-groups with a norm that is orders of magnitude smaller than the median, highlighting the opportunity for variable bitwidth allocation.



\subsection{Non-uniform quantization}
Given an input $X\in\mathbb R^d$, the common choice of $Q$ is to place $|Q|$ uniformly spaced quantization values in the range $[\min X, \max X]$ (e.g., QSGD~\cite{alistarh2017qsgd} and Uniform-THC~\cite{li2024thc}).
To optimize the accuracy for $|Q|$ quantization values, \mbox{we can place them non-uniformly. }

For example, consider $X=(-1,1/2,1)$; if we want $|Q|=3$, uniform SQ would use $Q=\set{-1,0,1}$ resulting in a  Mean Squared Error (MSE) of $\sum_{x\in X} \mathbb E[(\widehat {x} - x)^2] =1/4$. In contrast, picking the non-uniform $Q=\set{-1,1/2,1}$ results in an MSE of $0$. Generally, the MSE of non-uniform SQ can be asymptotically lower, e.g., for $X=(-1,1/2,\ldots,1/2,1)$, i.e., where $1/2$ repeats $d-2$ times. Here, the MSE of uniform SQ is $\Omega(d)$ \mbox{while the non-uniform SQ is accurate.}

\vspace{-0.1cm}
\subsection{Negative correlation}\label{subsec:background-negative}
\vspace{-0.05cm}
Intuitively, we can take the idea of unbiased quantization a step forward by explicitly `encouraging' errors to cancel out~\cite{suresh2022correlated}.
This is achieved by using shared randomness between different workers, which is facilitated by them sharing a peudo-random number generator seed. Using shared randomness is a common practice in quantization works (e.g.,~\cite{ben2025better,ben2024optimal,ben2020send,ben2026quantizing,ben2026note}).

For example, consider the simple case where two workers hold numbers $x_1,x_2\in[0,1]$ that need to be quantized into $1$ bit each, with the goal of estimating $x_1+x_2$. A standard approach is to have independent randomness where workers generate uniform random variables $u_1,u_2\sim \mathcal U[0,1]$ and quantize 
\begin{equation*}
    \widehat {x_1} = \begin{cases}
    1 & \mbox{if $u_1\le x_1$}\\
    0 & \mbox{otherwise}
    \end{cases}\quad , \quad
    \widehat {x_2} = \begin{cases}
    1 & \mbox{if $u_2\le x_2$}\\
    0 & \mbox{otherwise}
    \end{cases}\ .
\end{equation*}
To leverage negative correlation, we use the shared randomness to set $u_2=1-u_1$; i.e., both workers generate the same $u_1\sim \mathcal U[0,1]$ but use it in a different way to increase the chances that they round in the opposite direction. For example, if $x_1=x_2=1/2$, the independent randomness variance is $\Var[\widehat x_1 + \widehat x_2]=1/2$ while the negative correlation's variance is $0$. More generally, the variance (for any input $x_1,x_2$) of the negative correlation approach is at most $1/4$, i.e., improving the worst-case variance by a factor of 2.


\section{The \sys Framework}\label{sec:framework}

\begin{figure*}[t]
  \centering

  \hspace{-0.5cm}
  \subfigure[Each worker computes its super-group metadata.]{
    \begin{minipage}[t]{0.48\linewidth}
      \vspace{-0.00in}
      \begin{center}
        \includegraphics[trim=60mm 80mm 50mm 30mm,clip,width=\textwidth]{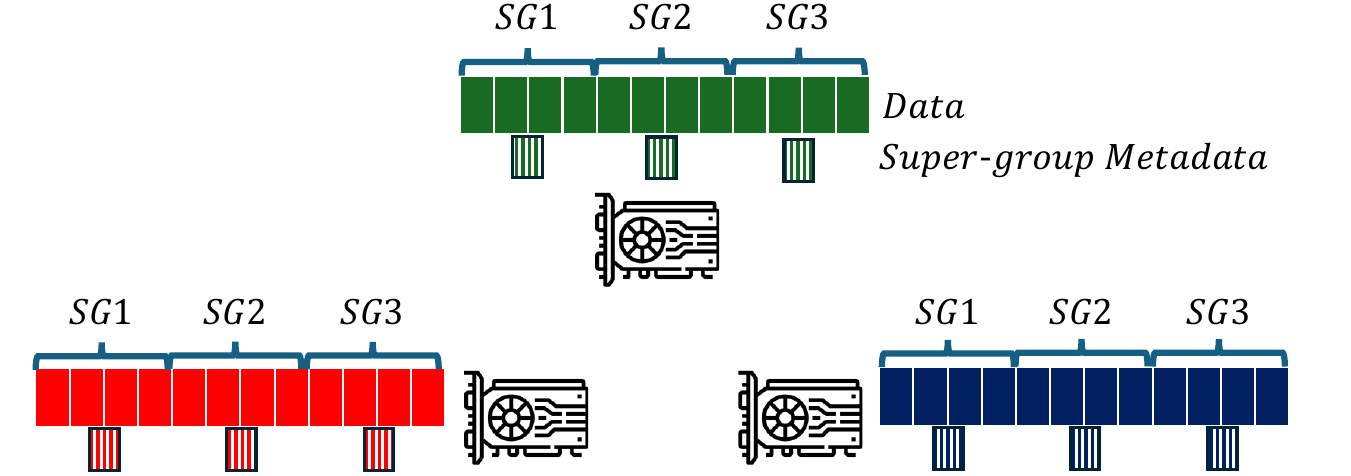}
      \end{center}
    \end{minipage}
    \label{subfig:overview-matrix-a}
  }
  \hspace{-0.1cm}
  \subfigure[Lightweight all-reduce syncs super-group metadata.]{
    \begin{minipage}[t]{0.48\linewidth}
      \vspace{-0.00in}
      \begin{center}
        \includegraphics[trim=60mm 80mm 50mm 30mm,clip,width=\textwidth]{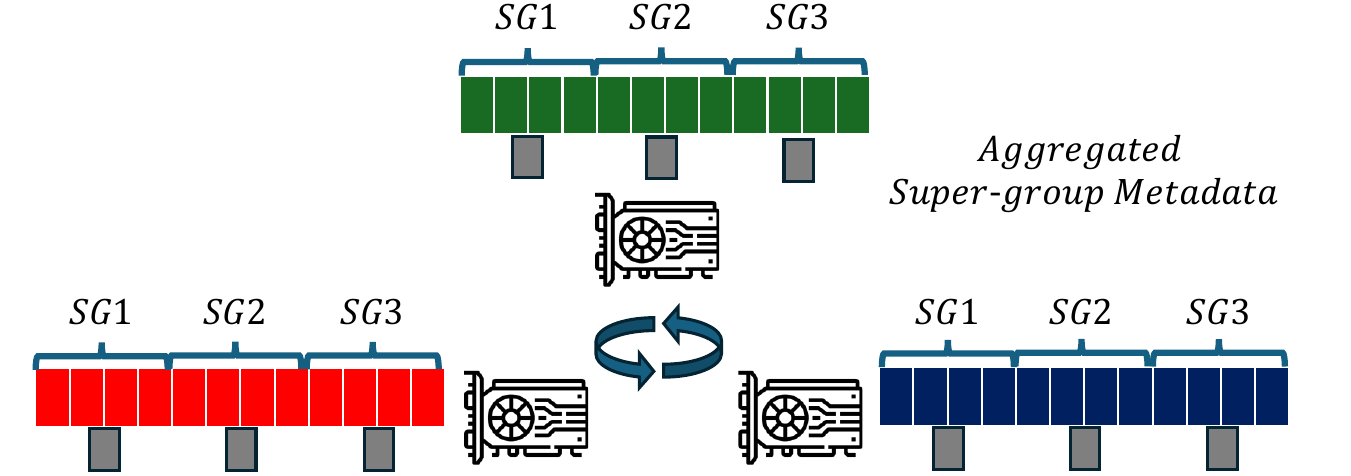}
      \end{center}
    \end{minipage}
    \label{subfig:overview-matrix-b}
  }

  \vspace{2mm}

  \hspace{-0.5cm}
  \subfigure[Data normalization and reordering based on synced metadata.]{
    \begin{minipage}[t]{0.48\linewidth}
      \vspace{-0.00in}
      \begin{center}
        \includegraphics[trim=60mm 80mm 50mm 30mm,clip,width=\textwidth]{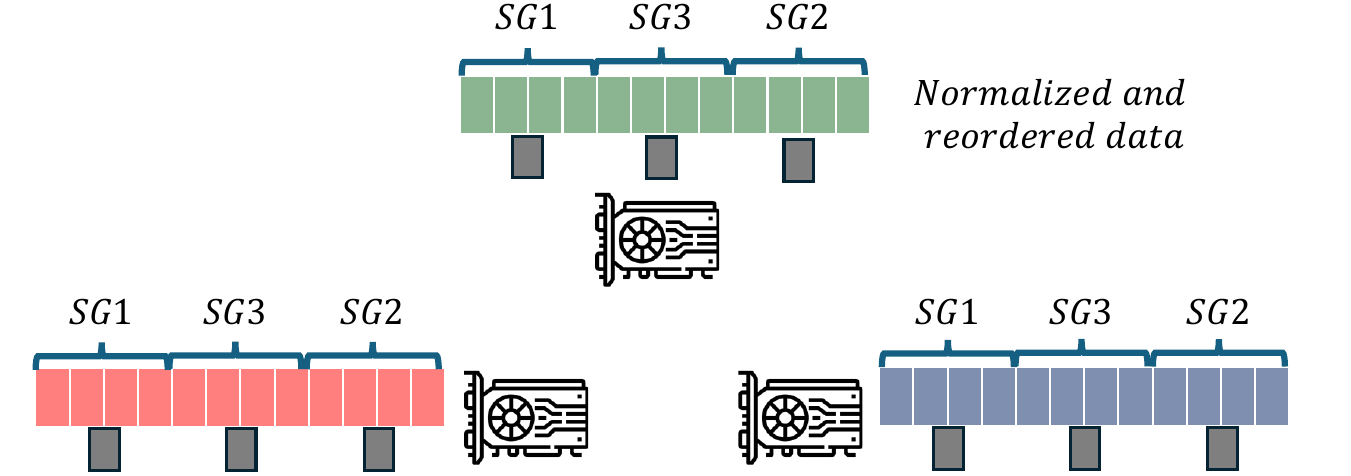}
      \end{center}
    \end{minipage}
    \label{subfig:overview-matrix-c}
  }
  \hspace{-0.1cm}
  \subfigure[A worker receives quantized data and applies the fused kernel.]{
    \begin{minipage}[t]{0.48\linewidth}
      \vspace{-0.00in}
      \begin{center}
        \includegraphics[trim=60mm 93mm 50mm 20mm,clip,width=\textwidth]{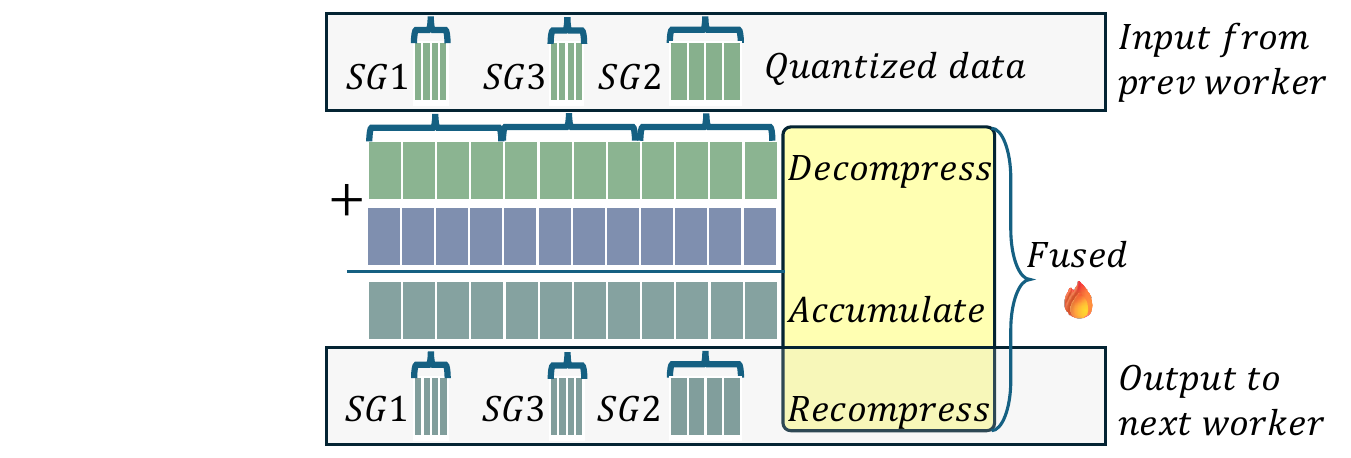}
      \end{center}
    \end{minipage}
    \label{subfig:overview-matrix-d}
  }

  \vspace{2mm}

  \hspace{-0.5cm}
  \subfigure[At the end of the main all-reduce, entries are synced.]{
    \begin{minipage}[t]{0.48\linewidth}
      \vspace{-0.00in}
      \begin{center}
        \includegraphics[trim=60mm 80mm 50mm 30mm,clip,width=\textwidth]{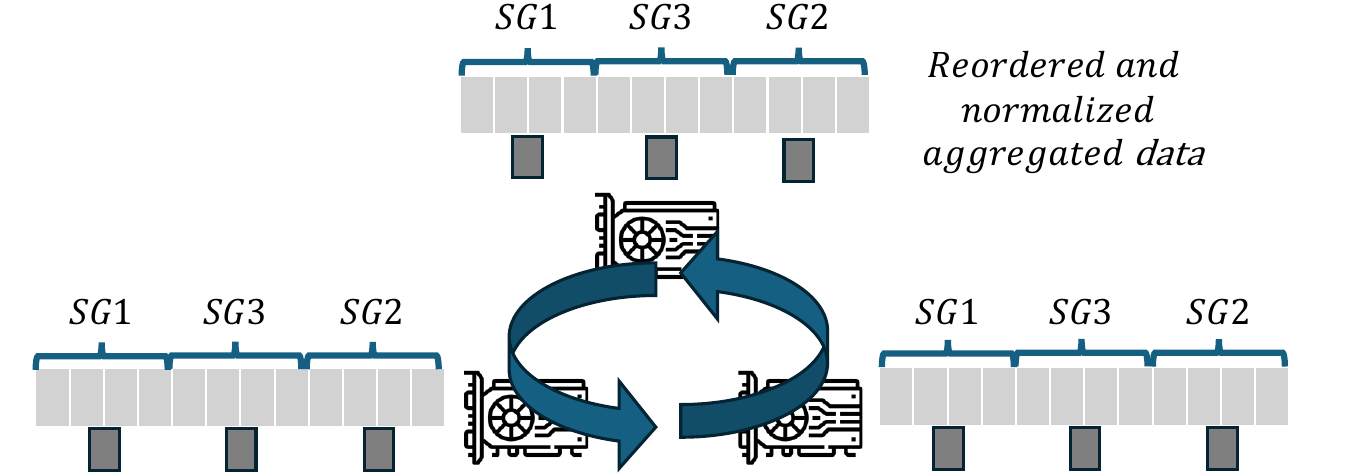}
      \end{center}
    \end{minipage}
    \label{subfig:overview-matrix-e}
  }
  \hspace{-0.1cm}
  \subfigure[Unnormalize and order entries to obtain the synced gradient.]{
    \begin{minipage}[t]{0.48\linewidth}
      \vspace{-0.00in}
      \begin{center}
        \includegraphics[trim=60mm 80mm 50mm 30mm,clip,width=\textwidth]{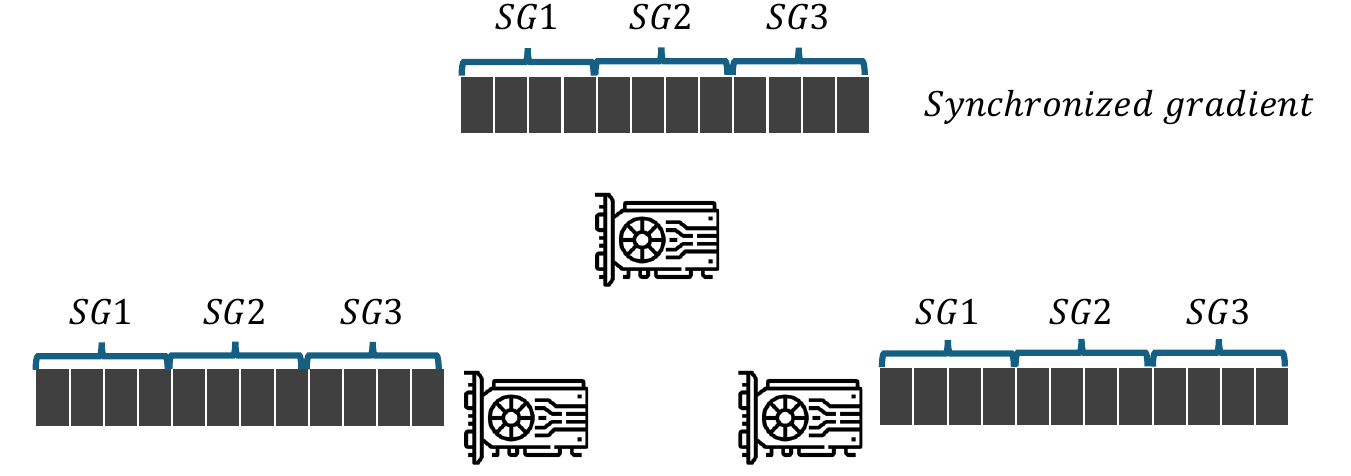}
      \end{center}
    \end{minipage}
    \label{subfig:overview-matrix-f}
  }
  \vspace{-0.2cm}

  \caption{The \sys workflow: (a) workers first compute the metadata (mean and $\ell_2$ norm) for each of their super-groups; (b) a lightweight all-reduce call aggregated the metadata such that all workers know that global super-group means and sum of $\ell_2$ norms; (c) based on the aggregated metadata, each worker normalizes each super-group by subtracting its global mean and reorders the super-groups based on their bit width which is based on the $\ell_2$ norms. Notice that in this example, SG3 has lower bit width than SG2 and thus their places are swapped; (d) illustrates how the blue worker operates during the main all reduce. It invokes the fused kernel to first decompress the received compressed partial sums data from the green worker, accumulates its local data, and recompresses the result before sending it to the red worker; (e) after the main all-reduce terminates, all workers have the same aggregated sums; (f) each worker adds back the global mean of each super-group and orders the data back to obtain the synced gradient. }
  \label{fig:overview-matrix}
  \vspace{-0.2cm}
\end{figure*}

Figure~\ref{fig:overview-matrix} illustrates an overview of the framework.

In the first stage, each worker partitions the gradient into super-groups of $S$ entries, and computes the metadata per super-group (\cref{subfig:overview-matrix-a}). Next, this local metadata is aggregated by an initial all-reduce call (\cref{subfig:overview-matrix-b}) that is lightweight since it contains only the super-group means and sum of $\ell_2$ norms, so the volume typically less than $1\%$ of the original gradient.
We then perform per-super-group normalization and allocate variable bitwidths for different super-groups. We reorder the gradient such that super-groups with the same bitwidth appear consecutively (\cref{subfig:overview-matrix-c}), and perform the main all-reduce (\cref{subfig:overview-matrix-d,subfig:overview-matrix-e}). 
Finally, we post-process the aggregated data by reordering the super-groups to their original position and adding their mean value back to {obtain the synced gradient (\cref{subfig:overview-matrix-f}).}
We also propose and adopt several techniques to reduce the compression error and implement our algorithm using fused kernels to minimize the computational overhead.

\subsection{Obtaining super-group statistics}\label{subsec:design-auxiliary-allreduce}

%
We formalize the first two stages (\cref{subfig:overview-matrix-a} and \cref{subfig:overview-matrix-b}).
Assume a setup with $n$ workers. Let $X_{i,j}$ be the $j$'th super-group of worker $i$.
For each $i,j$, the $i$'th worker first computes the mean given by $\mu_{i,j} = \sum_{x\in X_{i,j}} x / |X_{i,j}|$ and squared $\ell_2$ norm given by $F_{i,j}=\sum_{x\in X_{i,j}} x^2$. \sys then uses an initial all-reduce stage to aggregate these values. At the end of this stage, for each super-group $j$, all workers have the global mean $\mu_j$ and sum-of-squared-norms $F_j$ for that super-group. Formally,
{\setlength{\abovedisplayskip}{4pt}\setlength{\belowdisplayskip}{4pt}
\[
\mu_j = \frac{1}{n}\cdot \sum_{i=1}^n\mu_{i,j} \quad , \quad F_j=\sum_{i=1}^n F_{i,j} \,.
\]}

As illustrated in Figure~\ref{subfig:overview-matrix-c}, once these values are obtained, each worker normalizes their data by subtracting $\mu_j$ from each entry in super-group $j$, making it zero-mean, and then uses $F_j$ values to determine the bitwidth and reorder the data as we explain next.

\subsection{Determining super-group bitwidths}\label{subsec:design-super-group-bitwidth}
Due to the skewed distribution of the gradients (Figure~\ref{fig:demo-per-group-vs-shuffled}), allocating more bits to super-groups with larger norms could substantially decrease the quantization error. We use variable bitwidth allocation to minimize the quantization error and respect any given bandwidth constraint. \revise{In particular,
%
%
\sysname accepts a parameter $\bar{b}$ that denotes the average per-coordinate bit-budget. That is, for any throughput optimal primitive (e.g., Ring, Butterfly), the total number of bits to be sent by the network is $\frac{2(n-1)}{n}\bar{b}$ per coordinate per worker, where $n$ is the number of workers. This includes both the quantized data and any additional metadata (e.g., scales). (Note $\bar{b}$ must be large enough for the metadata and 1 bit per entry.)}

To allow efficient bit packing, we limit the possible quantized bitwidths to powers of 2, namely, 1,2,4,8, and 16. This also has the benefit of simplifying finding a performant variable quantization. As the choice of 16 bits corresponds to uncompressed values, this generalizes existing approaches where some entries (such as top-$k$/outliers) are encoded accurately while others are quantized to \mbox{a smaller number of bits~\cite{benaccelerating}. }

\begin{figure}
    \hspace{-0.2cm}
        \subfigure[LLaMA 1B MMLU]{
		\begin{minipage}[t]{0.48\linewidth}{
		\vspace{-0.00in}
		\begin{center}
		\includegraphics[width=\textwidth, ]{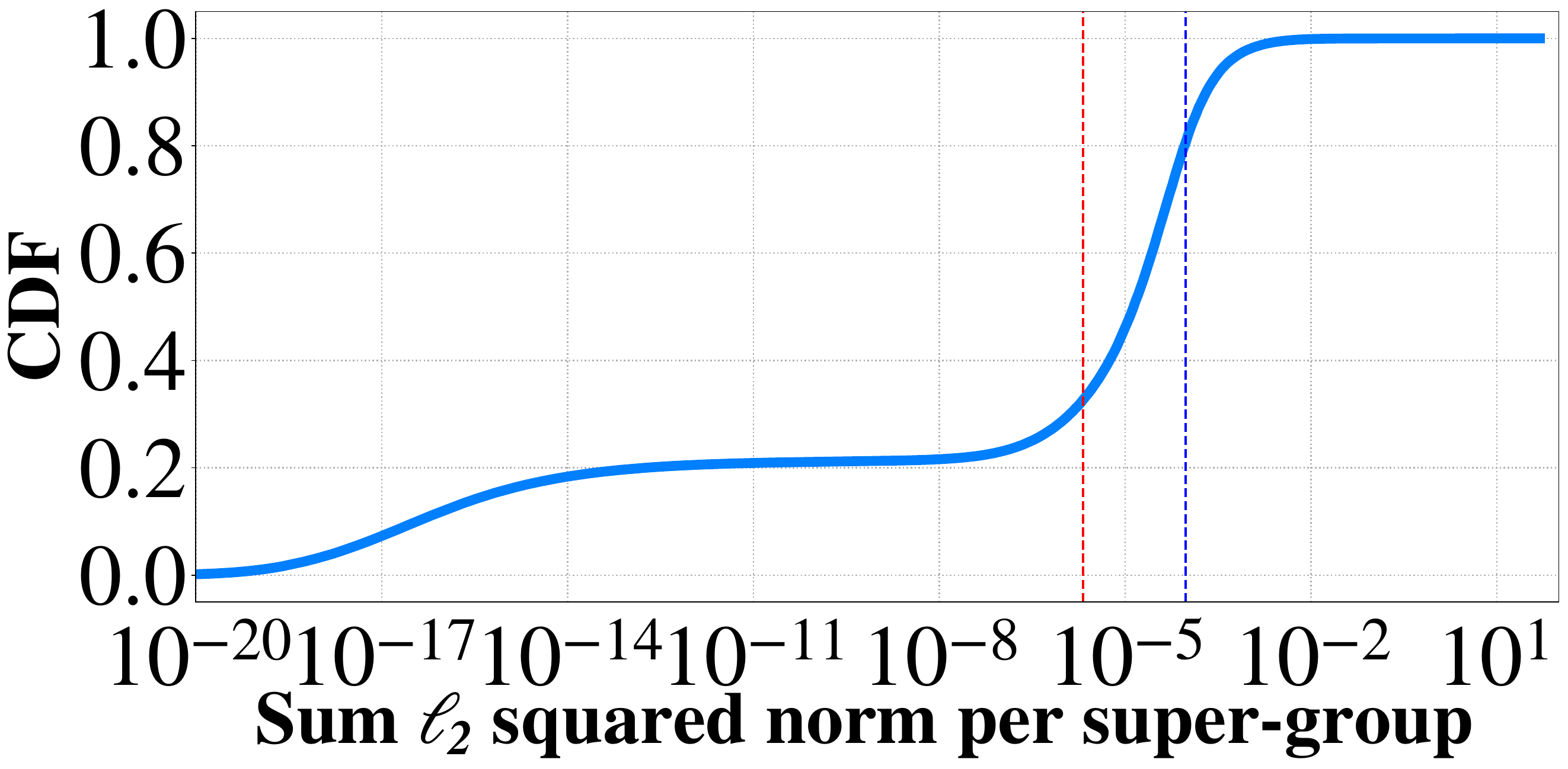}
		\end{center}
        \vspace{-0.1cm}
		}
		\label{subfig:design-multiple-bitrate_llama}
		\end{minipage}
	}
    \subfigure[Gemma 1B Chat]{
		\begin{minipage}[t]{0.48\linewidth}{
		\vspace{-0.00in}
		\begin{center}
		\includegraphics[width=\textwidth, ]{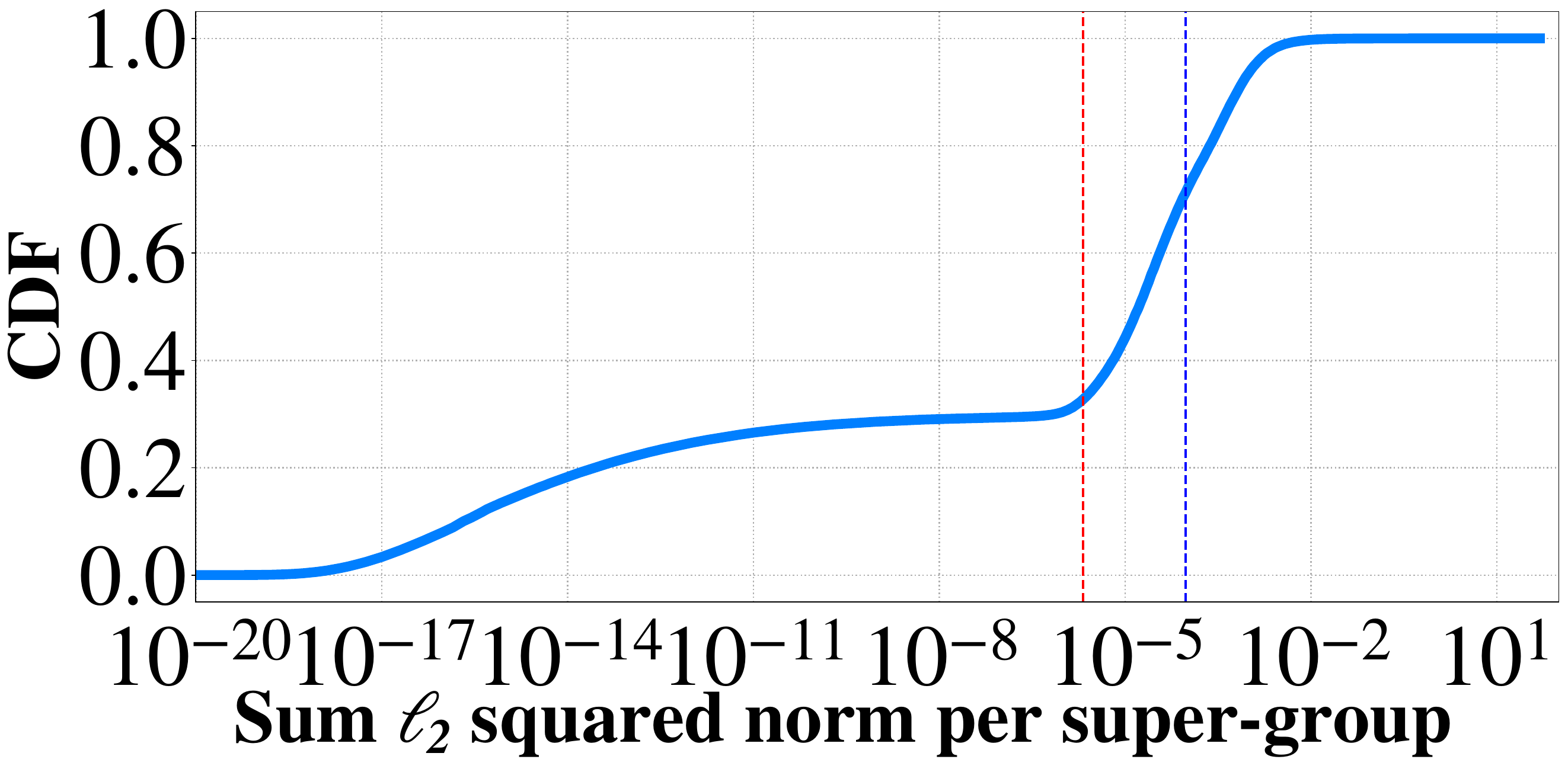}
		\end{center}
            \vspace{-0.1cm}
		}
		\label{subfig:design-multiple-bitrate_gemma}
		\end{minipage}
	}
    \vspace{-0.cm}

    \caption{The CDF distribution of $F_j$, summed $\ell_2$ squared norm per super-group across workers. The vertical dashed lines are thresholds for our variable bitwidth allocation algorithms, where super-groups with larger $\ell_2$ norms are assigned more bits in one of $2, 4$, or $8$ bits. }
    \vspace{0.1cm}
    \label{fig:design-multiple-bitrates}
\end{figure}

We now describe our fast heuristic approach to bandwidth partitioning to the different super-groups. For a given set of allowed bitwidths (e.g., $W=\set{1,2,4,8,16}$), we use \emph{thresholds} $T_{a,b}$ (dashed lines in Figure~\ref{fig:design-multiple-bitrates}), where $a,b$ are consecutive in $W$, to denote the boundary of the $F_j$ values with the same allocation. Intuitively, the MSE of quantizing a set is proportional to its squared norm, and thus the $F_j$ values serve as proxy for the expected error of the $j$'th super-group.
For ease of presentation, in this section we use the above $W$, although the technique is general.

Let $T_{0,1} = 0$ and $T_{16,32} = \infty$ for convenience. Then, all entries within super-groups with $F_j\in [T_{a,b}, T_{b,c})$ are quantized to $b$ bits. 
Next, we derive the relations between the thresholds. Suppose we start from a given set of thresholds $\set{T_{a,b}}$ and want to increase the bandwidth in a way that best reduces the MSE. This can be achieved by lowering a selected threshold $\set{T_{a,b}}$, increasing the quantized bit width of some super-group from $a$ to $b$ bits per entry.

The intuition behind our approach is based on a simple worst-case analysis. 
Consider an example where $Q$ includes two quantization values $0,1$ and an entry $x \in [0, 1]$. The worst-case variance for $\widehat x$ is when $x=1 / 2$, which yields $\Var[\widehat x] = 1 / 4$. Now, suppose we increase the quantization bit width by $1$ bit per entry (i.e., we double the size of $Q$). In that case, we can place a quantization value between every two consecutive values in $Q$, including between $0$ and $1$. If this extra quantization value is at $1/2$, the worst-case becomes when $x=1/4$, yielding $\Var[\widehat x] = 1/16$, i.e., a $4\times$ reduction. This reduction generalizes to any two consecutive quantization values $x_\downarrow, x_\uparrow\in Q$, i.e., for each additional bit, we can reduce the worst-case MSE of the entire vector by $4\times$.

Suppose we decrease ${T_{a,b}}$ just enough that a single super-group $j$ will have its entries encoded by $b$ bits instead of $a$. 
The above intuition suggests that each added bit decreases the MSE roughly $4\times$. If the super-group previously had an MSE proportional to $T_{a,b}\cdot4^{-a}$, then its MSE lowers by roughly proportionally to $T_{a,b}\cdot(4^{-a}-4^{-b})$ while increasing the bandwidth by $b-a$ bits per entry in this super-group.
We therefore estimate the \emph{per-bit benefit} of this action by $\frac{T_{a,b}\cdot(4^{b-a}-1)}{4^b\cdot(b-a)}$.
For example, if $a=1, b=2$, we lower the MSE by about $T_{1,2}\cdot 3/16$ with a per-bit benefit of $T_{1,2}\cdot 3/16$. Similarly, if $a=2, b=4$, we lower the MSE about $T_{2,4}\cdot 15/256$ and the per-bit benefit is $T_{2,4}\cdot 15/512$ while the per-bit benefit of lowering $T_{4,8}$ is $255/4^9$.
To optimize the thresholds selection, we require that the per-bit benefit of increasing all thresholds is roughly the same (e.g., $T_{1,2}\cdot 3/16 = T_{2,4}\cdot 15/256$), yielding:
$$
T_{1,2} = 5/32 \cdot T_{2,4},\quad T_{2,4} = 17/512\cdot T_{4,8} ,\quad T_{4,8} = 257/2^{17}\cdot T_{8,16}.
$$

Notice that this gives us $|W|-1$ constraints, leaving a single degree of freedom (e.g., by selecting $T_{1,2}$, the rest are determined). Accordingly, we search for the value of $T_{1,2}$ and determine the other thresholds by the above formula, such that the desired bandwidth constraint $\bar{b}$ is met.

Since we are interested in minimizing the computational overhead of finding the above threshold, for the practical case where at most three permissible bitwidths are used (e.g., we use $W=\set{2,4,8}$ in our implementation), we develop a fast binary-search-based solution to determine the thresholds in \Cref{app:variable-bitwidth-approx}.


\subsection{\sys's quantization}\label{subsec:quant-algorithm}
We now describe \sysname's quantization algorithm. The algorithm is used to compress the data of the first chunk along the aggregation topology, as well as for decompressing and recompressing partial sums, as illustrated in~\Cref{subfig:overview-matrix-d}.

\smallskip
\subp{Non-uniform quantization.} Motivated by~\Cref{subsec:background-grouped-quantization}, \sysname uses grouped quantization where a given group $G$ is quantized using $b$ bits. Each group is also associated with metadata, namely, a scaling parameter $\mathit{sf}$, which is used as explained below.
With $b$ bits per entry, we can represent each entry using a sign bit and a representation in $\set{0,\ldots,2^{b-1}-1}$. We hereafter assume that all quantization values $q\in Q$ are non-negative as we encode the sign bit separately.
We choose the values of $Q$ non-uniformly, similarly to~\cite{mee}. Specifically, denoting 
$
f(\epsilon,r) = \frac{(1+2\epsilon^2)^{r}-1}{(1+2\epsilon^2)^{2^{b-1}-1}-1}, 
$
for a parameter $\epsilon>0$, we use
$$
    Q = \set{ f(\epsilon,r)\mid r\in\set{0,\ldots,2^{b-1}-1}}.
$$

The result is some $Q\subset[0,1]$, where $\epsilon$ affects how non-uniform the quantization values are. Intuitively, when $\epsilon\approx 0$, $Q$ is roughly uniformly partitioned in $[0,1]$; a larger $\epsilon$ yields more quantization values close to zero and fewer large ones.

Note that as the entries in $G$ can be arbitrary BF16 values, we need to normalize them to $[0,1]$ before we can stochastically quantize to $Q$; this is achieved by dividing each entry $x$ by $\max |G| \triangleq \max\set{|x|\mid x\in G}$ and encoding the sign of $x$ separately.

\subp{Hierarchical quantization.} Next, the sender represents the group using a scaling factor $\mathit{sf}$ and a per-entry tuple $(r,\varsigma)$, where $r$ is the representation and $\varsigma$ is the sign bit.
In turn, the receiver estimates the entry as $\varsigma\cdot f(\epsilon,r)\cdot \mathit{sf}$. 

The natural choice is to set the scaling factor of the group $G$ to $\mathit{sf_G}=\max\set{|x|\mid x\in  G}$. However, this would require transmitting $\mathit{sf_G}$ in high precision (e.g., 16 bits), incurring a significant bandwidth overhead when the group size $s$ is small.

Instead, \sys optimizes the accuracy-bandwidth tradeoff by quantizing the scaling factors within a super-group, a method known as \textit{hierarchical quantization}~\cite{GGUF}. 
Namely, let $\mathcal G$ be a super-group, and let $\mathit{\mathit{sf_{\mathcal G}}}=\max |\mathcal G|
$ equal the largest absolute value of an entry in $\mathcal G$, i.e., $\max\set{|x|\mid x\in \mathcal G}$.
We encode $\mathit{\mathit{sf_{\mathcal G}}}$ in half-precision for the entire super-group, and quantize each individual group's scaling factor using uniform stochastic quantization such that $\mathbb E[\mathit{sf_G}] = \max |G|$. 
For example, we can represent $\mathit{sf_G}$ using a UINT8 representation $r_G\in\set{0,\ldots,255}$, decoded \mbox{as $\mathit{sf_G} = r_G \cdot \mathit{sf_{\mathcal G}} / 255$.}

An essential property of our hierarchical quantization is that the estimates of individual entries remain unbiased. Consider a specific entry $x\in G$. It is first normalized to $x'=x/\max |G|$, which is then stochastically quantized to $\widehat{x'}\in Q$. As explained above, the scale of the group itself is then stochastically quantized to $$\mathit{\mathit{sf_G}}\in\set{r_G \cdot \mathit{sf_{\mathcal G}} / 255 \mid r_G\in\set{0,\ldots,255}}.$$

The estimated value of $x$ is therefore:
$
\widehat x = \widehat{x'}\cdot \mathit{sf_G},
$
which, due to the independence of the randomness used in the two quantization steps (of $x'$ and of $\mathit{sf_G}$), satisfies:
$$
\mathbb E[\widehat x] \! = \! \mathbb E[\widehat{x'}\cdot \mathit{sf_G}]\! = \! \mathbb E[\widehat{x'}]\cdot \mathbb E[\mathit{sf_G}] \! = \! (x/\max |G|) \cdot \max |G| = x.
$$
That is, even though quantizing $\mathit{sf_G}$ simultaneously scales all entries in $G$ within the same scaling factor, the individual entries' unbiasedness is retained.



\subp{Correlated rounding.} We now explain how to leverage negative correlation (Section~\ref{subsec:background-negative}) in the stochastic quantization of the entries.
Intuitively, we want to increase the likelihood that if a given worker quantizes a specific partial sum upwards, another will quantize down, so that the overall result is closer to the true sum.

Formally, let $p_i$ be the probability of rounding up a given recompressed partial sum at worker $i$. 
As exemplified in~\Cref{subsec:background-negative}, stochastic rounding is commonly implemented by drawing a uniform random variable $u_i \sim \mathcal U[0,1]$ and rounding up if $u_i < p_i$  and rounding down otherwise. 
Using Suresh et al.~\cite{suresh2022correlated} correlated sampling method, instead of drawing $u_i$ independently at random for different workers, we set 
$$u_i = \frac{\pi_i + \gamma_i}{n}.$$
Here, $\pi = \set{\pi_i}$ is a random permutation of ${0, \ldots, n-1}$ and $\gamma_i \sim \mathcal U[0,1]$. Importantly, $\pi$ is implicitly agreed upon by all workers as it is independently generated using the same pseudo-random number generator and is not communicated.

Each $u_i$ is still uniformly distributed, but now correlated across workers. That is, in every interval $[0, \frac{1}{n}), \cdots, [\frac{n-1}{n}, 1)$ exactly one worker’s $u_i$ falls inside. Intuitively, if $u_i$ from one worker falls within $[0, \frac{1}{n})$ so that worker $i$ rounds up with high probability, there will be another worker $i'$ such that $U_{i'} \in [\frac{n-1}{n}, 1)$ and worker $i'$ rounds down with high probability, canceling the error.

\subsection{Main all-reduce}\label{subsec:design-main-allreduce}

\revise{\sysname's main all-reduce phase follows standard communication patterns such as ring~\cite{ring} and butterfly~\cite{thakur2005optimization}, but with compressed transfers. 
We next explain \sysname's two all-reduce components.
}
\remove{In particular, an all-reduce is composed of two sequential phases: the reduce-scatter phase and the all-gather phase.}

\subp{Reduce-scatter.} The $n$ workers split their gradients into $n$ \emph{chunks}, where $C_{i,j}$ is the $i$'th chunk of the $j$'th worker. For each index $i\in\set{0,\ldots,n-1}$, chunks $\set{C_{i,j}\mid j\in\set{0,\ldots,n-1}}$ are aggregated, in parallel with chunk sets of other $i$'s.
For each such $i$, the reduce-scatter topology is an in-arborescence, i.e., a tree where all edges point towards a single sink. For example, on a ring all-reduce, the aggregation topology for a single chunk is simply a path, while for butterfly we visualize the topology in \Cref{fig:butterfly-demo} in \Cref{sec:adaptation-butterfly}.\footnote{Note that as the different $i$'s are aggregated in parallel; e.g., in ring all-reduce, each node acts both as a sender and a receiver, and the total communication pattern forms a cycle.} 

The aggregation works as follows: Leaf nodes (which receive no external messages for a specific chunk index $i$) transmit their \revise{compressed} local chunk to the next node in the topology. Internal nodes serve as intermediaries that \revise{decompress and aggregate} partial sums \revise{(including their local chunk), recompress, and send to the next node}, while the sink node for \revise{the chunk terminates its reduce-scatter phase by decompressing and adding its local chunk.}

\subp{All-gather.} Upon the completion of the reduce-scatter phase, the all-gather phase commences, during which the sinks broadcast the aggregated sums to all other workers.

\subp{Fused kernels.} We now describe how we optimize the all-reduce (illustrated in~\Cref{subfig:overview-matrix-d}). 
\sys employs four distinct types of fused kernels determined by the accumulation state and node type. 
The first kernel is used to compress the chunk entries at the leaf nodes.
Internal nodes then use a decompress-accumulate kernel when they have received partial sums from all but the last of their parent nodes. 
When receiving the partial sum of the last parent, they apply a decompress-accumulate-recompress kernel to also get the sum ready for the next transmission.

The aggregated compressed chunk sums then continue to the all-gather phase, where they are broadcast. Whenever a node receives a compressed sum, it invokes the decompress kernel (ending in~\Cref{subfig:overview-matrix-e}).
The operation concludes with a final reconstruction step (\Cref{subfig:overview-matrix-f}) where entries are restored to their original order and unnormalized by adding back the mean value, which was subtracted during \mbox{the initial all-reduce step, to recover the final output.}

\subsection{Communication and runtime overhead}

\revise{We summarize \sysname's communication and runtime overheads.}

\subp{Communication overhead.}
\revise{
\sysname incurs communication in two all-reduce stages.
First, in the lightweight all-reduce stage, let $C$ and $C_{\mathit{s}}$ be the group and super-group sizes.
For each super-group $\mathcal G$, \sys computes exactly two BF16 numbers via the reduction, $\mu_{\mathcal G}$ and $F_{\mathcal G}$.
Therefore, this stage communicates $32/C_{\mathit{s}}$ bits per coordinate.
Second, in the main all-reduce stage, as explained in~\Cref{subsec:design-super-group-bitwidth}, \sys deducts the bandwidth used by the lightweight stage from the overall communication budget.
Thus, the main stage uses $\bar b - 32/C_{\mathit{s}}$ bits per coordinate. 
For example, for $\bar b = 5$ and $C_{\mathit{s}} = 256$, the first stage uses $32/C_{\mathit{s}}=0.125$, leaving $\bar b - 0.125=4.875$ bits per coordinate for the second (main) stage for both the quantized representations and group and super-group scales. For groups of size $C = 16$, the group's scale takes $8 / C = 0.5$ bits per coordinate and the super-groups' scale takes additional $16 / C_{\mathit{s}} = 1/16$ bits per coordinate. This leaves $4.875-0.5-1/16=4.3125$ bits on average for each quantized coordinate's representation. This bit budget is divided among super-groups with, e.g., $2$, $4$, or $8$ bitwidths (as detailed in Section \ref{subsec:design-super-group-bitwidth}) such that the resulting average is $4.3125$.
The reordering stage incurs no communication, since it only permutes super-groups locally into a pre-allocated tensor in monotonically increasing bitwidth order.
}

\subp{Compute overhead.}
\revise{
All \sysname operations are linear in the gradient size and rely on sequential memory accesses to HBM.
The lightweight stage computes $\mu_{\mathcal G}$ and $F_{\mathcal G}$ for all super-groups using a single kernel launch with one memory access per entry.
The reordering stage performs a single read and write per entry, with only a small additional overhead for maintaining the super-group indices.
The main decompress--accumulate and decompress--accumulate--recompress operations are each implemented as a single fused kernel. Each value is reconstructed on the fly from its quantization bin and scale, accumulated, and requantized without materializing intermediate tensors. 
Correlated rounding adds only local arithmetic and does not require additional memory accesses.
}

\section{Implementation}\label{sec:impl}


We implement the DynamiQ prototype atop PyTorch DDP~\cite{li2020pytorch} with NCCL~\cite{nccl} P2P. \revise{Our code is available at Github~\cite{open-source}}. As mentioned, our prototype centers on four CUDA kernels: 
\begin{enumerate}
    \item \texttt{DynamiQ\_compress($t$)} compresses a gradient chunk $t$ at a leaf node in the aggregation topology. 
    \item \texttt{DynamiQ\_decompress($\mathit{ct}$)} decompresses a compressed gradient chunk \texttt{$ct$} at the all-gather stage of all-reduce.
    \item \texttt{DynamiQ\_decompress\_accumulate\_recompress($ct$, $t$)} in non-leaf nodes fuses decompressing \texttt{$ct$}, accumulating it with \texttt{$t$}, and recompressing the sum.
    \item \texttt{DynamiQ\_decompress\_accumulate($ct$, $t$)} in intermediate hops executes decompression of \texttt{$ct$} and accumulating the result with \texttt{$t$} (without recompression). 
\end{enumerate}

\subp{Efficient fused-kernel CUDA implementation.} GPUs are typically memory-bound for elementwise operations~\cite{memory-bound, gholami2024ai, memory-bound2}, which means that the compression overhead is mainly determined by global memory transactions. Our design, therefore, minimizes this overhead by using fused kernels. 

\texttt{DynamiQ\_decompress\_accumulate\_recompress}, for example, fuses these three operations. The intermediate results are stored in registers, avoiding global memory accesses. As Table~\ref{tab:memory-transaction} shows, this significantly reduces memory traffic, making DynamiQ’s involved logic computationally lightweight. 

The super-group reordering allows GPU kernels to receive a sequence of streams of uniform bitwidth entries, enabling efficient memory addressing and coalescing. 
To share the maximum per-group gradient value for scaling, we use the classic parallel maximum reduction algorithm in CUDA. 
We use powers of two for the group size and super-group size to allow more effective threaded memory access and execution.


\subp{The DDP communication hook and RDMA engine.}
\revise{
We build the communication hook from P2P primitives, using NCCL within a node and RDMA across nodes. This enables all-reduce topologies not natively supported by NCCL collectives, including butterfly.
Also, our RDMA engine overlaps compression with communication using blockwise pipelining. As soon as a gradient block completes \texttt{decompress\_accumulate\_recompress}, its RDMA transfer begins while subsequent blocks are processed (we use a block size of 8~MB in our experiemtns).
}
%
%
\remove{For example, during the all-gather stage~\cite{collective-nccl}, if at a given time a worker received the compressed partial sum for the $i$'th chunk and has done aggregating and compressing chunk $j$, it decompresses $i$ while forwarding $j$ to its next hop.}





\section{Evaluation}\label{sec:eval}



\begin{figure*}
    \centering
    \begin{minipage}[t]{0.56\linewidth}{
		\vspace{-0.00in}
		\begin{center}
		\includegraphics[width=\textwidth, ]{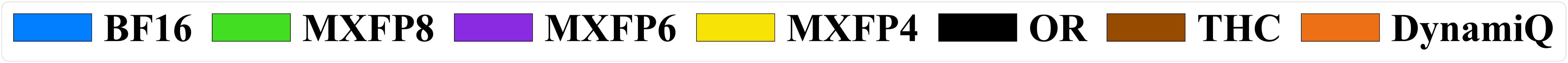}
		\end{center}
		}
        \end{minipage}

    \hspace{-0.2cm}
        \subfigure[BERT-large MaskedLM]{
		\begin{minipage}[t]{0.24\linewidth}{
		\vspace{-0.00in}
		\begin{center}
		\includegraphics[width=\textwidth, ]{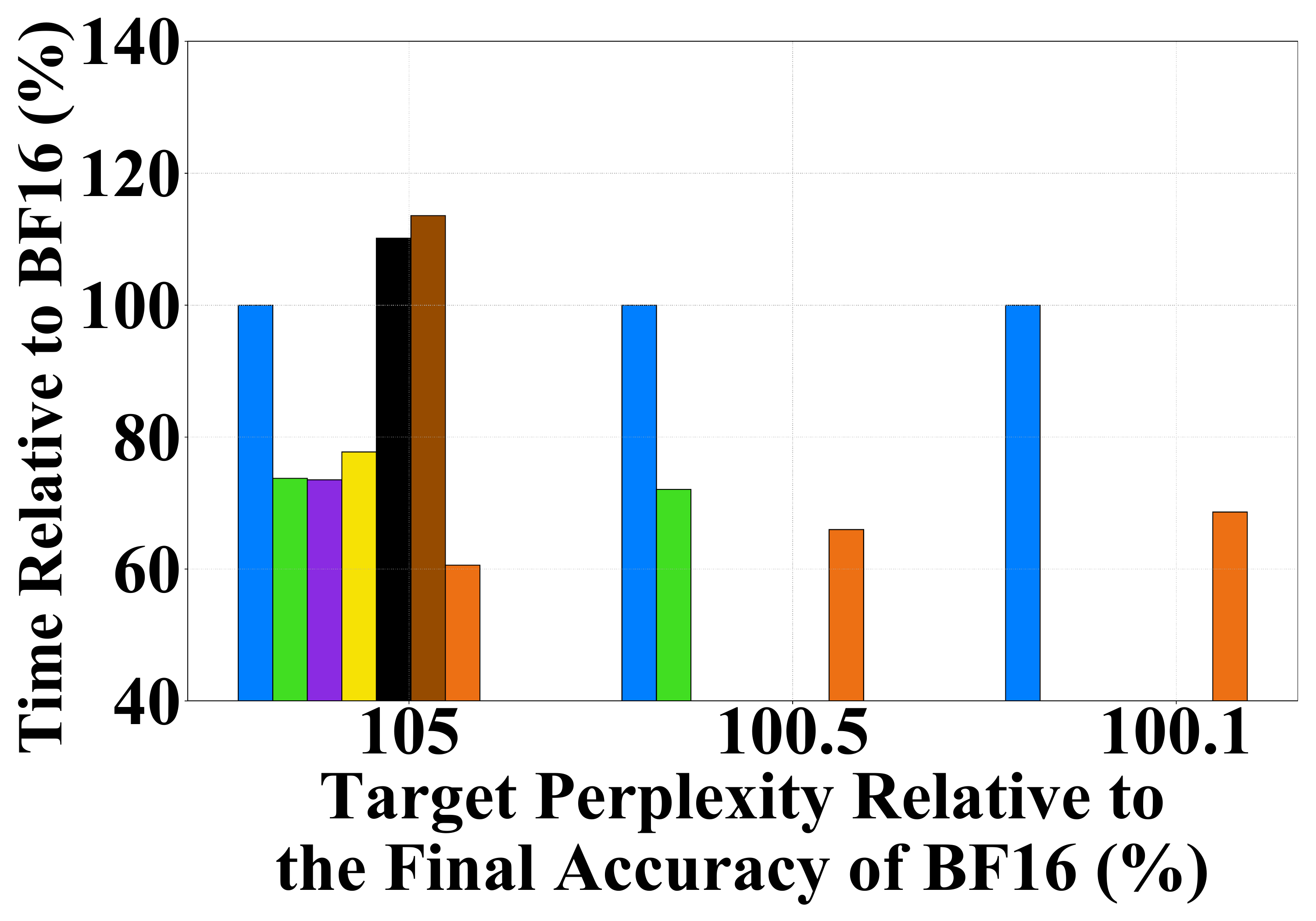}
		\end{center}
            \vspace{-0.1cm}
		}
		\label{subfig:demo-bert-maskedlm}
		\end{minipage}
	}
    \hspace{-0.2cm}
        \subfigure[Gemma 1B Chat]{
		\begin{minipage}[t]{0.24\linewidth}{
		\vspace{-0.00in}
		\begin{center}
		\includegraphics[width=\textwidth, ]{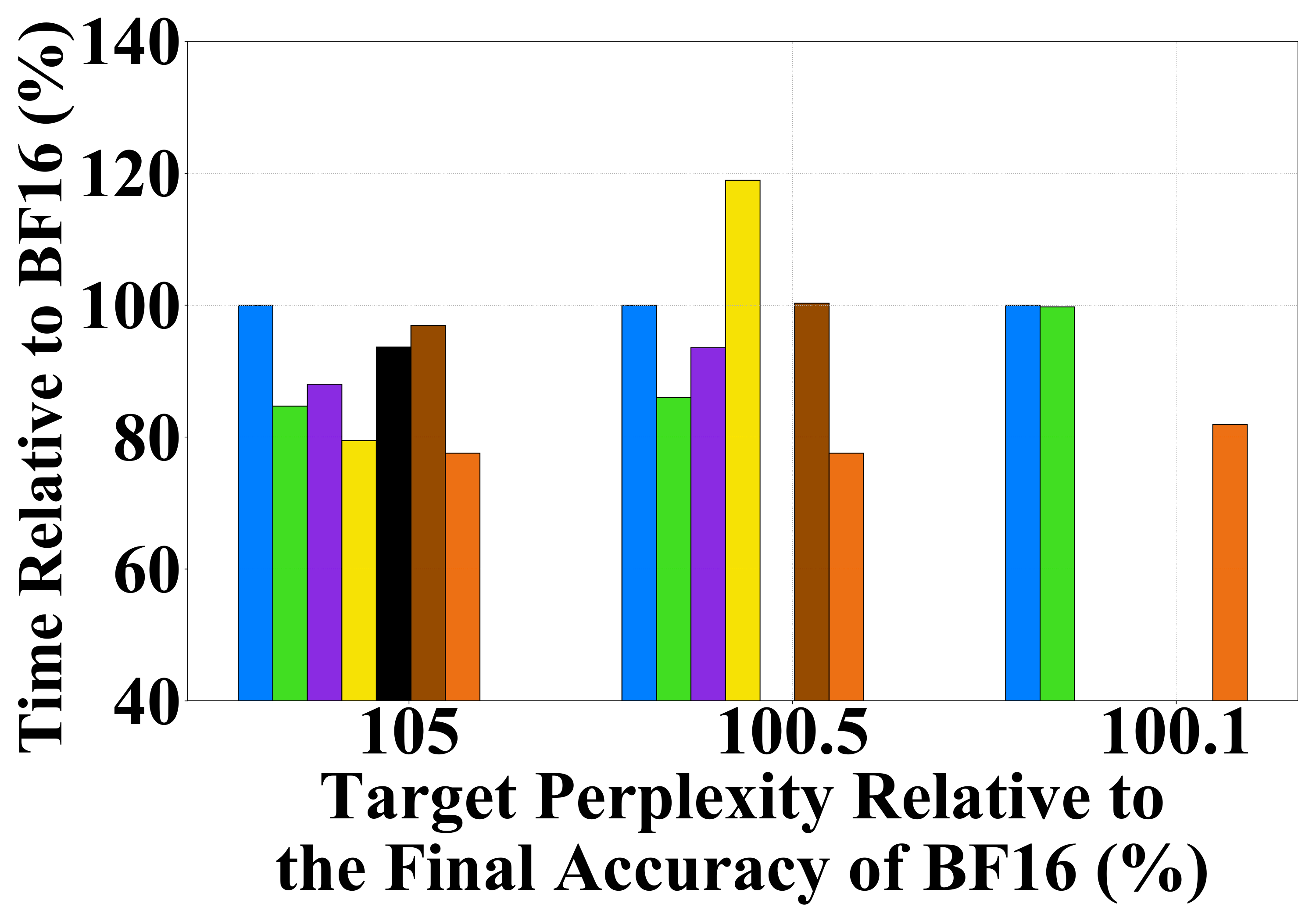}
		\end{center}
        \vspace{-0.1cm}
		}
		\label{subfig:demo-gemma-chat}
		\end{minipage}
	}
    \hspace{-0.2cm}
        \subfigure[LLaMA 1B Chat]{
		\begin{minipage}[t]{0.24\linewidth}{
		\vspace{-0.00in}
		\begin{center}
		\includegraphics[width=\textwidth, ]{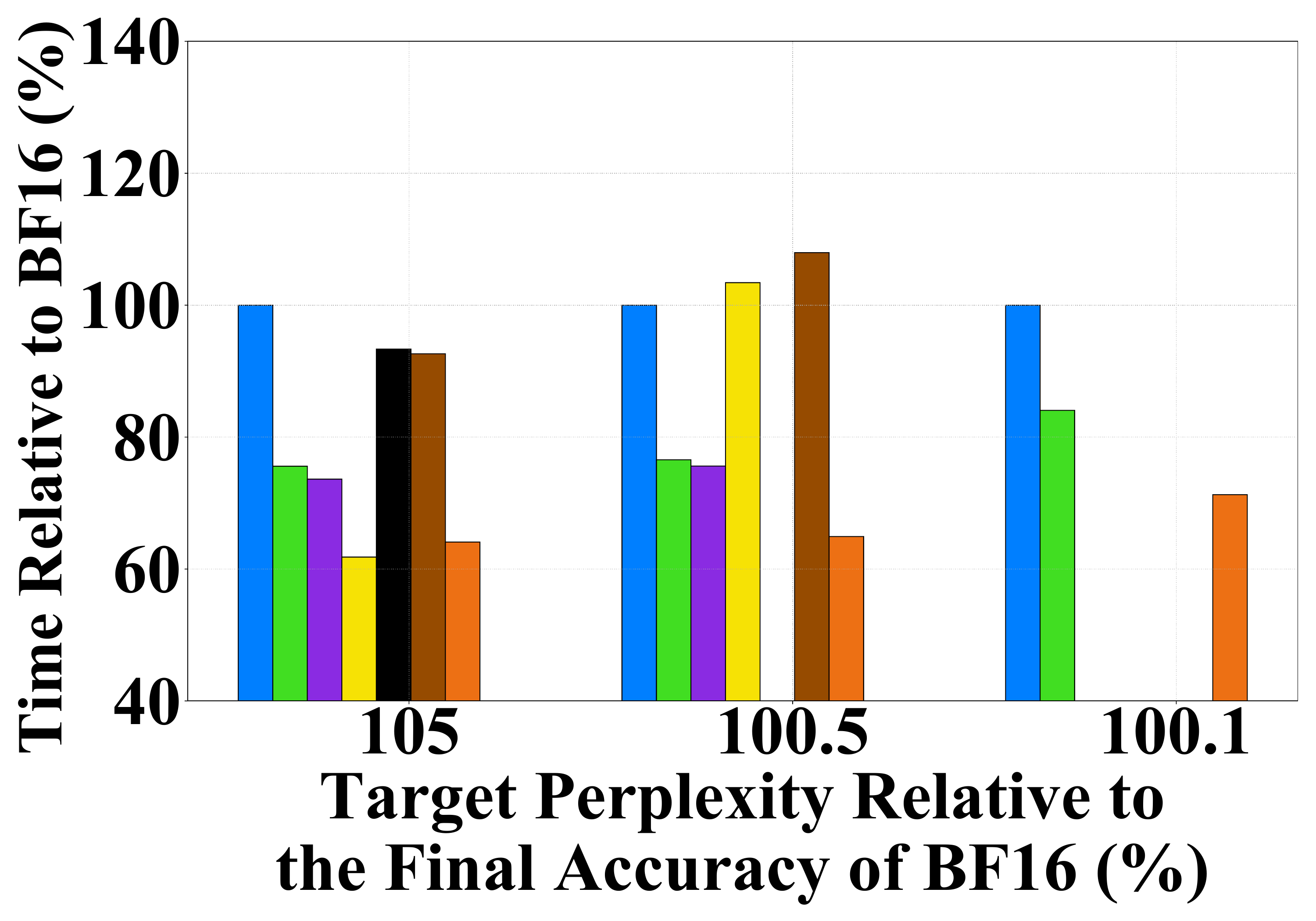}
		\end{center}
        \vspace{-0.1cm}
		}
		\label{subfig:demo-llama-chat}
		\end{minipage}
	}
    \hspace{-0.3cm}
        \subfigure[LLaMA 1B MMLU]{
		\begin{minipage}[t]{0.24\linewidth}{
		\vspace{-0.00in}
		\begin{center}
		\includegraphics[width=\textwidth, ]{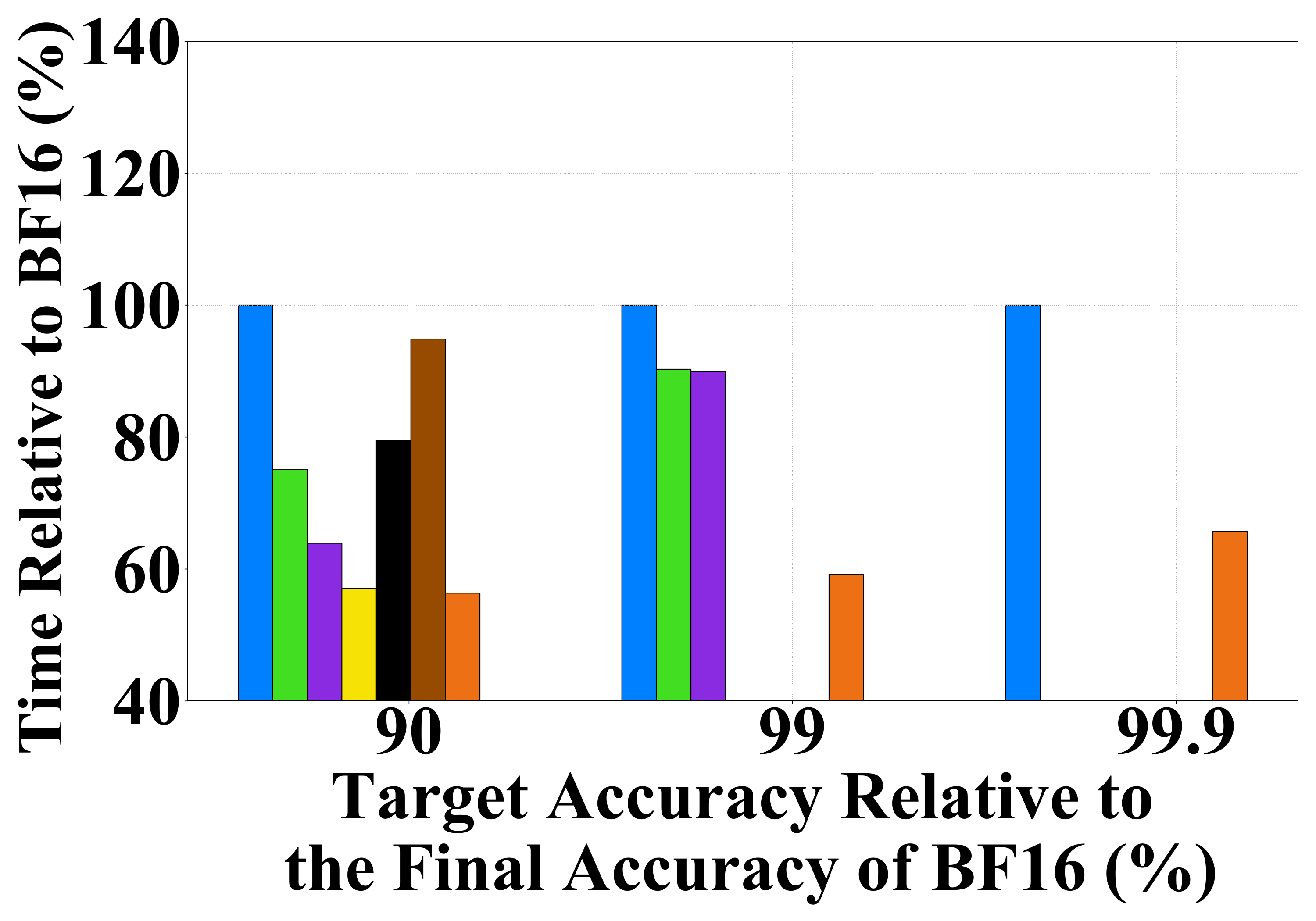}
		\end{center}
        \vspace{-0.1cm}
		}
		\label{subfig:demo-mmlu}
		\end{minipage}
	}

    \vspace{-0.2cm}
    \caption{Time-to-target perplexity and accuracy for training (fine-tuning) LLMs on 8-GPU/4-worker testbed using ring all-reduce. We measure the time required relative to BF16 (lower is better) to reach specific convergence targets defined by BF16's final metrics (perplexities of $3.107$, $2.996$, $3.095$ and accuracy of $73.04\%$). For example, for BERT-large, "105\%" means we measure the time it takes to reach the perplexity of $3.107\cdot1.05 \approx 3.22$, and for LLaMA 1B MMLU, "99\%" means we measure the time it takes to reach the accuracy of $73.04 \cdot 0.99 \approx 72.3\%$. Bars are omitted for methods that do not reach the specified target.
    }
    \label{fig:demo-time-to-accuracy}
    \vspace{-0.0cm}
\end{figure*}

In this section, we present an end-to-end testbed evaluation of \sysname across four LLM \revise{full-parameter supervised fine-tuning workloads (that produce dense gradients) and compare it with state-of-the-art gradient compression schemes. } 
We first use the ring all-reduce topology and evaluate performance both in isolation (\Cref{subsec:end-to-end-eval}) and over a shared network (\Cref{subsec:multi-tenancy}).
Then, to demonstrate the applicability of \sys to different topologies, we then consider butterfly all-reduce (\Cref{subsec:butterfly}).

\subp{Testbed.}
Our testbed consists of four servers running CentOS. Each server is equipped with two NVIDIA RTX A6000 GPUs, each with 48\,GB of GDDR6 memory and 768\,GB/s of memory bandwidth. The two GPUs are connected through NVLink (NV4). Each server also has two 100\,Gb/s NVIDIA Mellanox ConnectX-6 NICs, of which we use one in our experiments. Each server contains two 16-core AMD EPYC 7313 CPUs and 512\,GB of host memory. Each GPU is connected to the host through PCIe 4.0 x16, providing approximately 32\,GB/s of bandwidth per direction.


\subp{Workloads.} We evaluate \sysname on four diverse training (fine-tuning) tasks representative of distinct model families and paradigms: standard masked language modeling with BERT-large~\cite{devlin2018bert} on Wikitext-103~\cite{wikitext}; decoder-based instruction tuning with Gemma 1B~\cite{team2025gemma} on UltraChat~\cite{ding2023enhancing}; reasoning with LLaMA 1B~\cite{grattafiori2024llama3herdmodels} on both UltraChat and MMLU~\cite{hendrycks2020measuring}.

\subp{Parameter setup.}
Table~\ref{tab: workload-setups} summarizes the batch size (in terms of tokens and sequences) and learning rate configurations we use in our experiments. For Wikitext and UltraChat, following common practice, we truncate and pack tokens into fixed-length sequences (potentially merging consecutive samples).  For learning rate scheduling, we use the standard \texttt{torch.optim.lr\_scheduler.LinearLR} scheduler with parameters detailed in the table.


\subp{Baseline compression schemes.} We compare \sysname against standard uncompressed BFloat16~\cite{bfloat16} (BF16) format and 5 state-of-the-art gradient compression schemes covering both traditional and emerging standards: adaptive sparsification via OmniReduce (OR)~\cite{fei2021efficient}, optimized fixed-point quantization via THC~\cite{li2024thc}
and emerging microscaling formats (MXFP8, MXFP6 and MXFP4)~\cite{rouhani2023microscaling}.

THC and OmniReduce were originally designed for the parameter-server architecture~\cite{li2014scaling} and are thus not optimized for multi-hop all-reduce. To ensure a fair comparison, we adapt them as follows. For THC, directly summing quantized gradients via multi-hop all-reduce leads to catastrophic overflows given limited bits $b$. Consequently, following the original paper, we compress local gradients into $q=4$-bit integers and allocate $b=8$ bits per coordinate for aggregation. For OmniReduce, we allocate $b=8$ bits and employ its chunked Top-$k$ compression variant. Since local Top-$k$ chunk indices differ across workers, we first aggregate indices appearing in at least one worker and then transmit the union. The challenge lies in dynamically determining $k$ such that the union of local Top-$k$ chunks equals $K$; we address this with a heuristic algorithm described in Appendix~\ref{appendix:exp_setup}. 

For the microscaling formats, we use E4M3, E3M2, and E2M1 for MXFP8, MXFP6, and MXFP4, respectively. For all formats, the chunk size is 32 and the per-chunk shared scale is BF16. For summation arithmetic and overflow and underflow hadling, we follow the implementation of FP8-LM~\cite{peng2023fp8}. We also dynamically adjust the per-chunk scales prior to the main all-reduce stage across training rounds to minimize overflows and underflows as described in \Cref{appendix:exp_setup}. Finally, since MXFP4 and MXFP6 are not natively supported by our testbed GPUs, we report their best-case (i.e., lower-bound) TTA estimates. This is done by decoupling accuracy (which is software-based) and timing (we transmit the equivalent traffic without performing any compute) measurments.


\subp{\sysname's configuration.} Unless otherwise stated, we set the group size as $s=16$ coordinates and the super-group size as $S=256$ coordinates (i.e., $16$ consecutive groups). As explained in \Cref{subsec:quant-algorithm}, per-group scaling parameters are quantized into UINT8 and per-super-group scaling parameters are kept as BF16.

We use $W=\{2, 4, 8\}$ and the approximation algorithm of~\Cref{app:variable-bitwidth-approx} to determine the bit allocation. Unless otherwise noted, we fix the overall budget at 
$b=5$ bits per coordinate, which we demonstrate to achieve the best end-to-end performance in our ablation study (see \Cref{fig: ablation_nbits}). 

\subp{Metrics.} We evaluate the end-to-end performance using the well-established time-to-accuracy (TTA) metric and pay special attention to the final accuracy. Namely, while TTA tracks evaluation accuracy~\footnote{Here we use the general term "accuracy" to refer to the respective LLM evaluation metric, including perplexity for MaskedLM and CausalLM for UltraChat, and classification accuracy for multiple choice question answering.} as a function of wall-clock time, final accuracy is recorded once the model converges. We also benchmark throughput (in rounds per second) and quantization error to assess the speed and quality of our quantization compared to other schemes. For the latter, we employ the vector normalized mean squared error (vNMSE) metric~\cite{karimireddy2019error, li2024thc, vargaftik2021drive, vargaftik2022eden, benaccelerating,dorfman2023docofl}, defined as $\mathbb E\left[\norm{X-\hat X}^2\right] / \norm{X}^2$.

\begin{table}[htbp]
\resizebox{1.\linewidth}{!}{
    \begin{tabular}{|l|c|c|c|c|}
    \hline
    \multirow{2}{*}{Workload} & BERT-large & LLaMA 1B & Gemma 1B Chat & LLaMA 1B \\ 
    & MaskedLM & Chat & Chat & MMLU \\ \hline
        Tokens per batch &  2048    &  3000    & 3000    &  $\sim 1600$    \\ \hline
        Batch size & 1 & 1 & 1 & 4 \\ \hline
        Initial LR &  $5\times 10^{-5}$    &  $2\times 10^{-5}$    & $1.4\times 10^{-5}$     &  $6\times 10^{-6}$    \\ \hline
        Linear LR end factors & ${1}/{16}$ & ${1}/{8}$ & ${1}/{8}$ & ${1}/{8}$ \\ \hline
        Linear LR total iters (Epochs) & 15 & 2 & 2 & 2 \\ \hline
        Total iters (Epochs) & 21 & 3 & 3 & 3 \\ \hline
    \end{tabular}
}
\caption{Configurations of the average number of tokens and the learning rate schedules in our workloads.}
\label{tab: workload-setups}
\end{table}

\vspace{-0.3cm}
\subsection{Ring all-reduce}\label{subsec:end-to-end-eval}

We next evaluate \sysname's end-to-end performance with ring all-reduce, demonstrating that it offers substantial acceleration over BF16 and significantly better TTA curves than all other tested compression schemes. Importantly, \sysname outperforms MXFP8 across all workloads despite using a lower bit budget, which is attributed to its ability to maintain both low compression error and to incur small computational overhead despite its ``involved'' two-phase~workflow.



\begin{figure*}[]
    \centering
    \begin{minipage}[t]{0.75\linewidth}{
		\begin{center}
		\includegraphics[width=\textwidth, ]{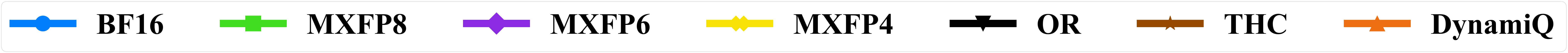}
		\end{center}
		}
        \end{minipage}

	\hspace{-0.2cm}
        \subfigure[BERT-large MaskedLM]{
		\begin{minipage}[t]{0.245\linewidth}{
		\vspace{-0.00in}
		\begin{center}
		\includegraphics[width=\textwidth, ]{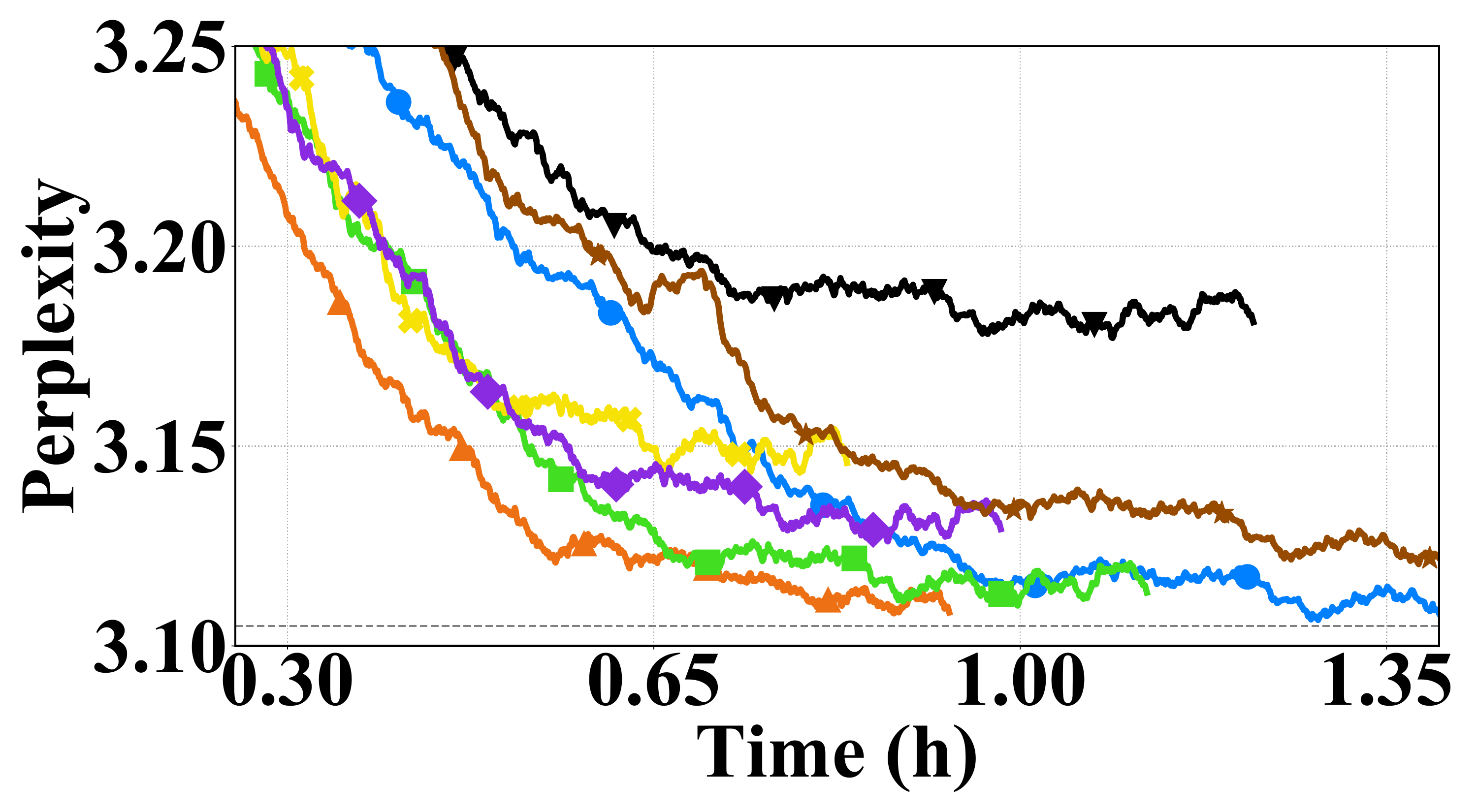}
		\end{center}
		}
		\label{subfig:e2e-bert-large}
		\end{minipage}
	}
        \hspace{-0.3cm}
        \subfigure[Gemma 1B Chat]{
		\begin{minipage}[t]{0.245\linewidth}{
		\vspace{-0.00in}
		\begin{center}
		\includegraphics[width=\textwidth, ]{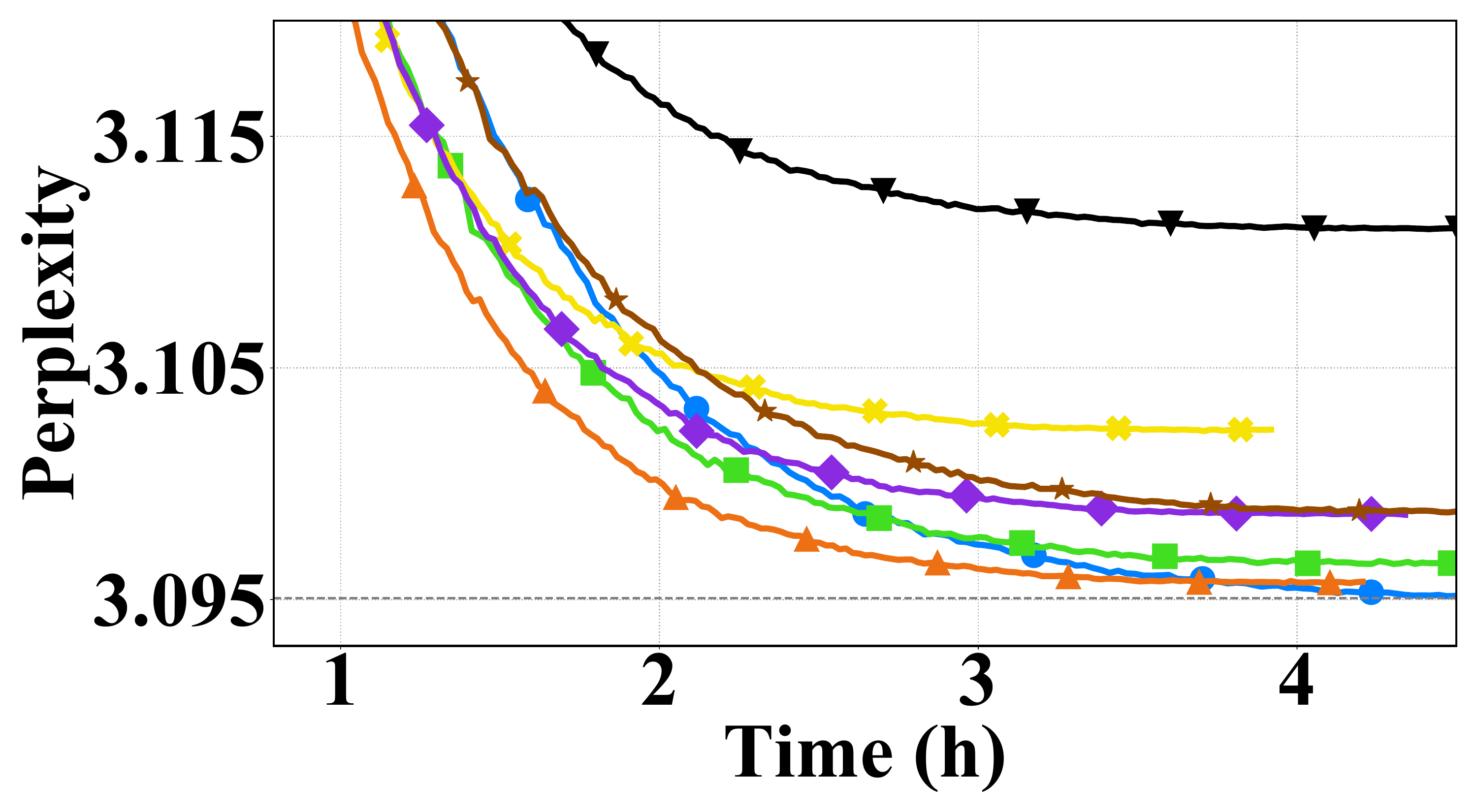}
		\end{center}
		}
		\label{subfig: e2e-gemma}
		\end{minipage}
	    }
    \hspace{-0.3cm}
	\subfigure[LLaMA 1B Chat]{
		\begin{minipage}[t]{0.245\linewidth}{
		\vspace{-0.00in}
		\begin{center}
		\includegraphics[width=\textwidth, ]{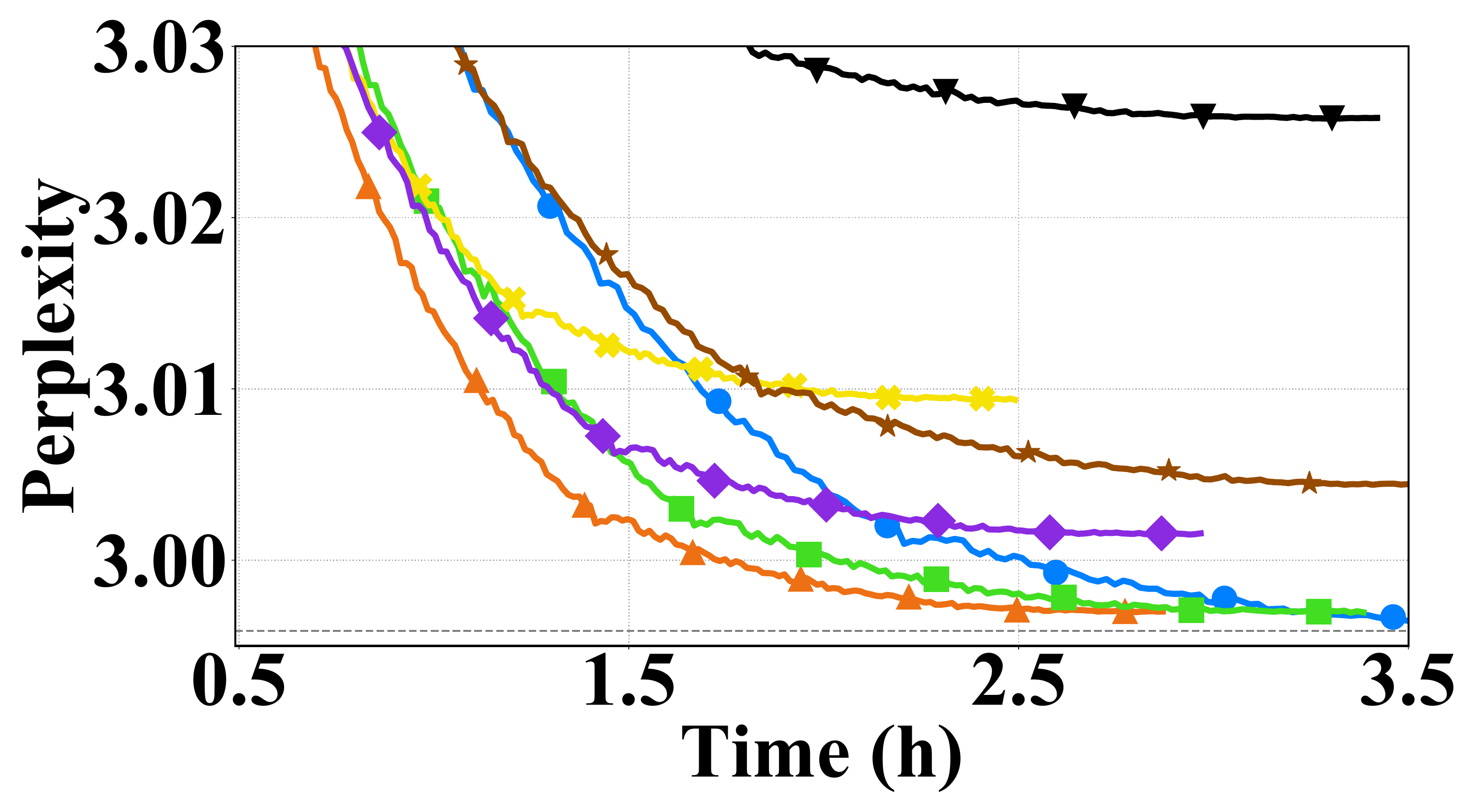}
		\end{center}
		}
		\label{subfig: e2e-LLaMA}
		\end{minipage}
	}
        \hspace{-0.3cm}
        \subfigure[LLaMA 1B MMLU]{
            \begin{minipage}[t]{0.245\linewidth}{
            \vspace{-0.00in}
            \begin{center}
            \includegraphics[width=\textwidth, ]{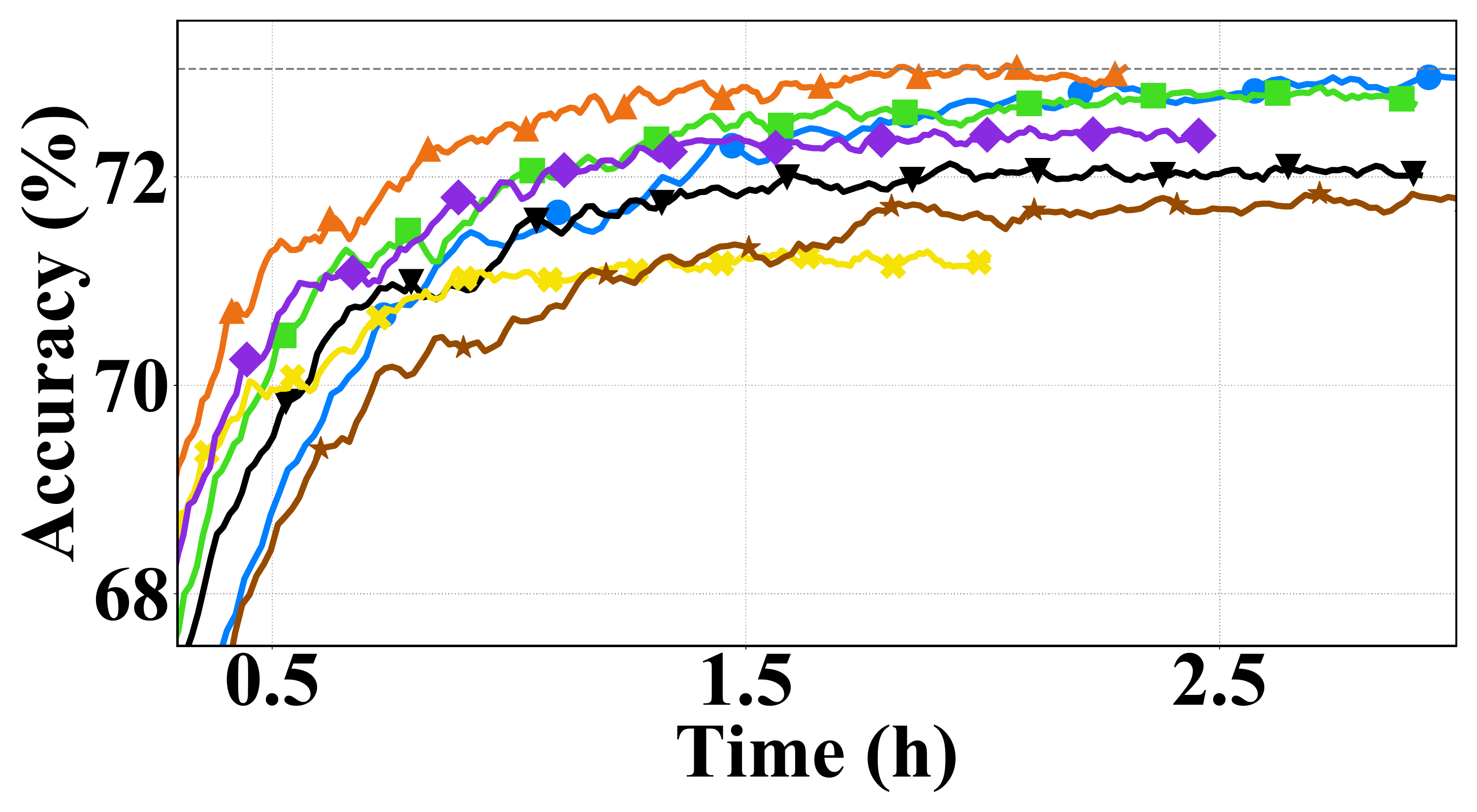}
            \end{center}
            }
            \label{subfig: e2e-mmlu}
            \end{minipage}
        }
        \vspace{-0.4cm}
    \caption{Zoomed-in Time to Accuracy (TTA) curves for LLM training (fine-tuning) on an 8-GPU/4-worker testbed using ring all-reduce. Horizontal dashed lines indicate the final BF16 accuracy. As mentioned, MXFP4 and MXFP6 curves represent a best-case scenario based on upper-bound throughput estimation. \sysname is the only method to consistently converge faster than BF16 while roughly matching its perplexity and accuracy, followed by MXFP8. Although alternatives like THC and OR also show faster-than-baseline initial convergence (see Appendix~\cref{fig: e2e-tta-full}), their performance ultimately stalls due to high compression error (see Appendix~\cref{fig: vnmse-e2e}).}
	\label{fig: e2e-tta}
\end{figure*}

\begin{figure*}[htbp]

    \centering
    \begin{minipage}[t]{1\linewidth}{
		\begin{center}
		\includegraphics[width=0.55\textwidth, ]{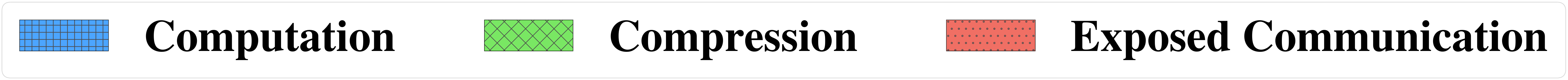}
		\end{center}
		}
       \end{minipage}

       \hspace{-0.2cm}
	\subfigure[BERT-large MaskedLM]{
		\begin{minipage}[t]{0.245\linewidth}{
		\vspace{-0.00in}
		\begin{center}
		\includegraphics[width=\textwidth, ]{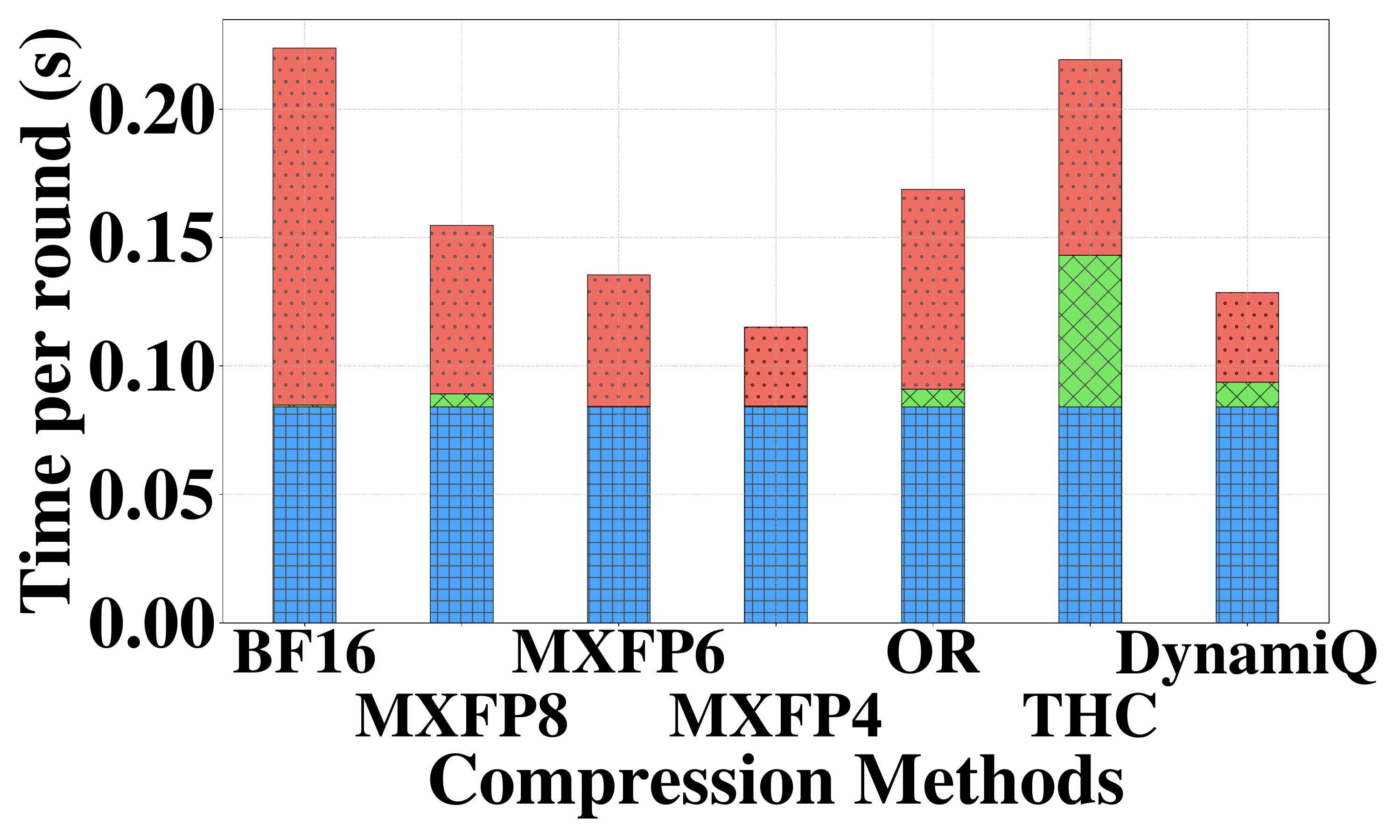}
		\end{center}
		}
		\label{subfig: breakdown-bert}
		\end{minipage}
	}
        \hspace{-0.1in}
        \subfigure[Gemma 1B Chat]{
		\begin{minipage}[t]{0.245\linewidth}{
		\vspace{-0.00in}
		\begin{center}
		\includegraphics[width=\textwidth, ]{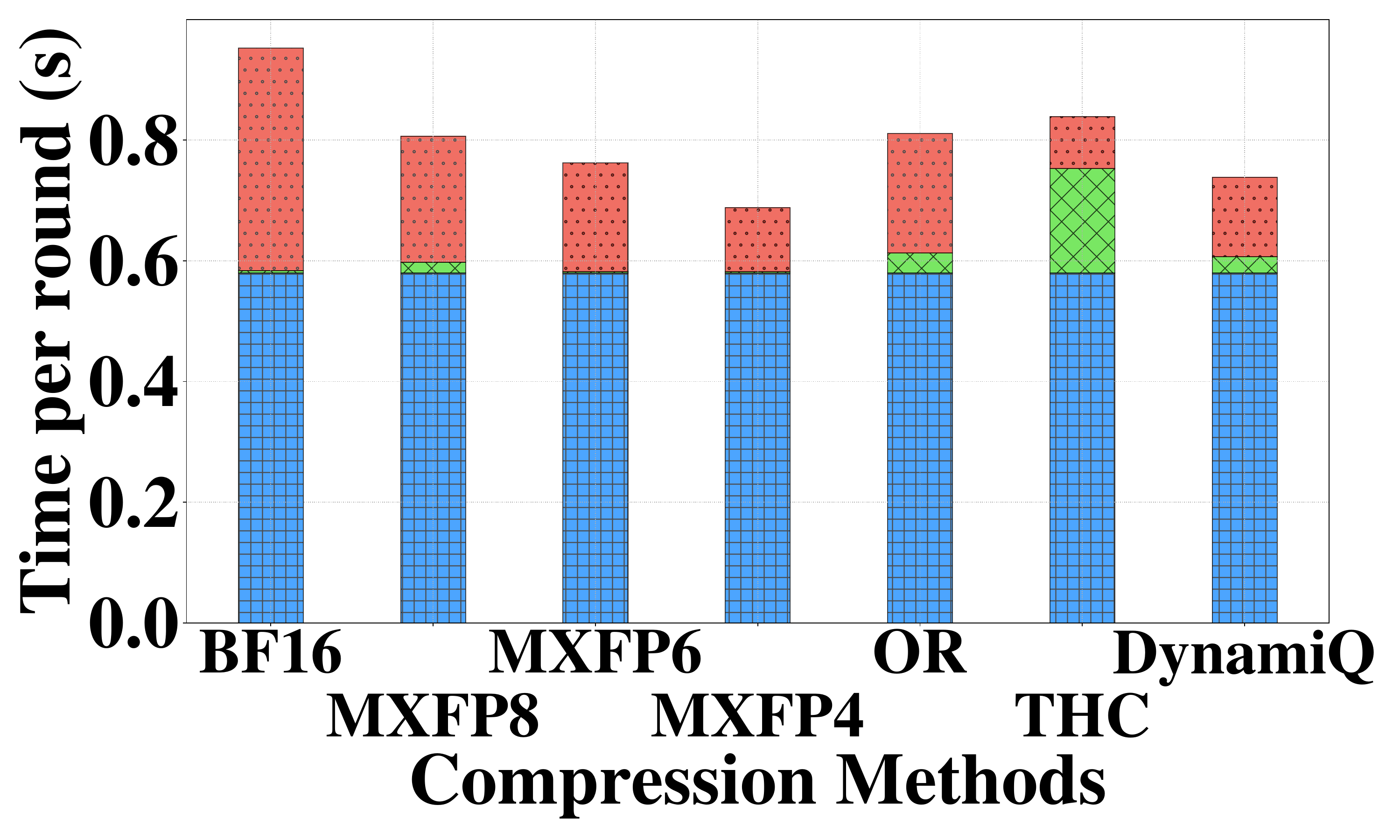}
		\end{center}
		}
		\label{subfig: breakdown-gemma}
		\end{minipage}
        }
        \hspace{-0.1in}
        \subfigure[LLaMA 1B Chat]{
		\begin{minipage}[t]{0.245\linewidth}{
		\vspace{-0.00in}
		\begin{center}
		\includegraphics[width=\textwidth, ]{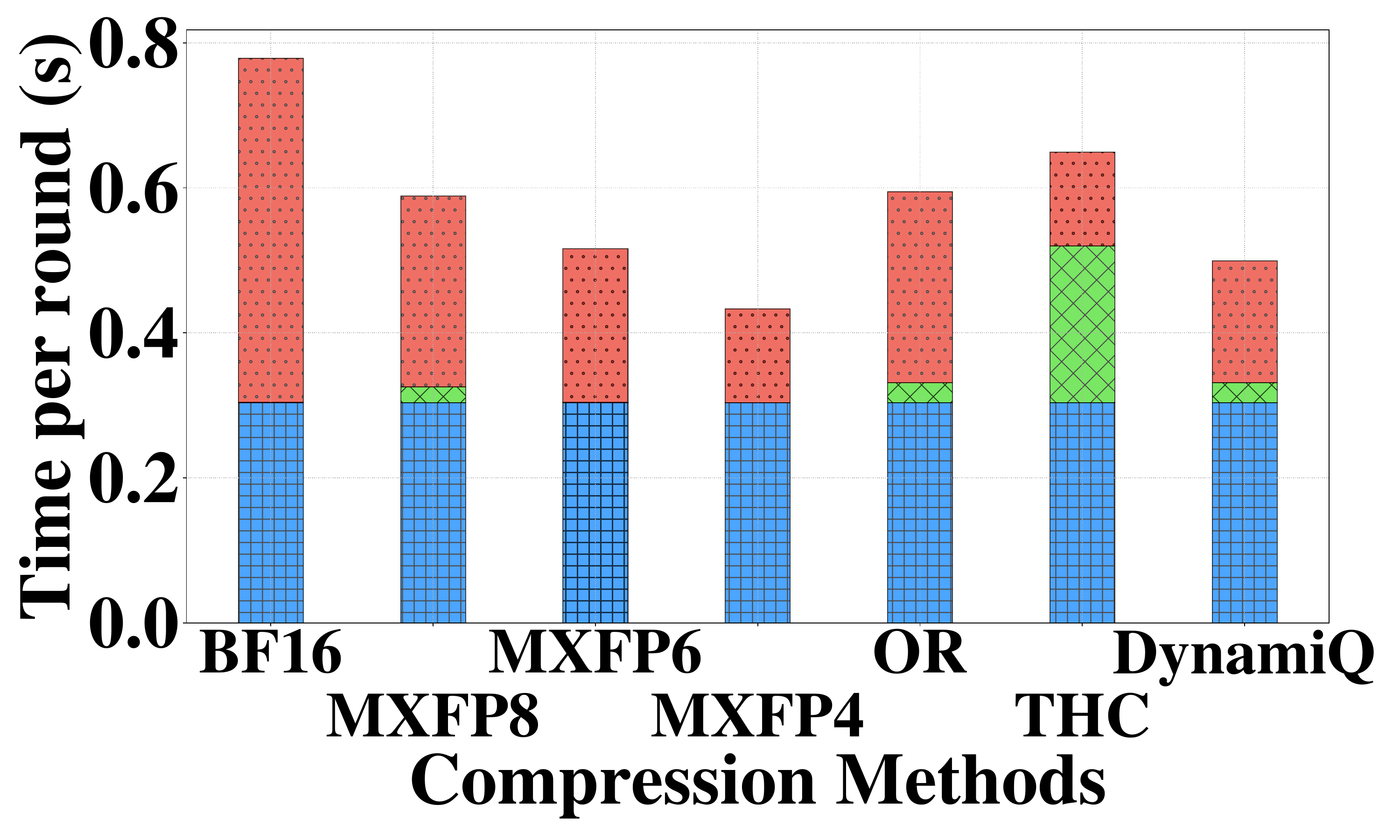}
		\end{center}
		}
		\label{subfig: breakdown-llama}
		\end{minipage}
        }
        \hspace{-0.1in}
        \subfigure[LLaMA 1B MMLU]{
		\begin{minipage}[t]{0.245\linewidth}{
		\vspace{-0.00in}
		\begin{center}
		\includegraphics[width=\textwidth, ]{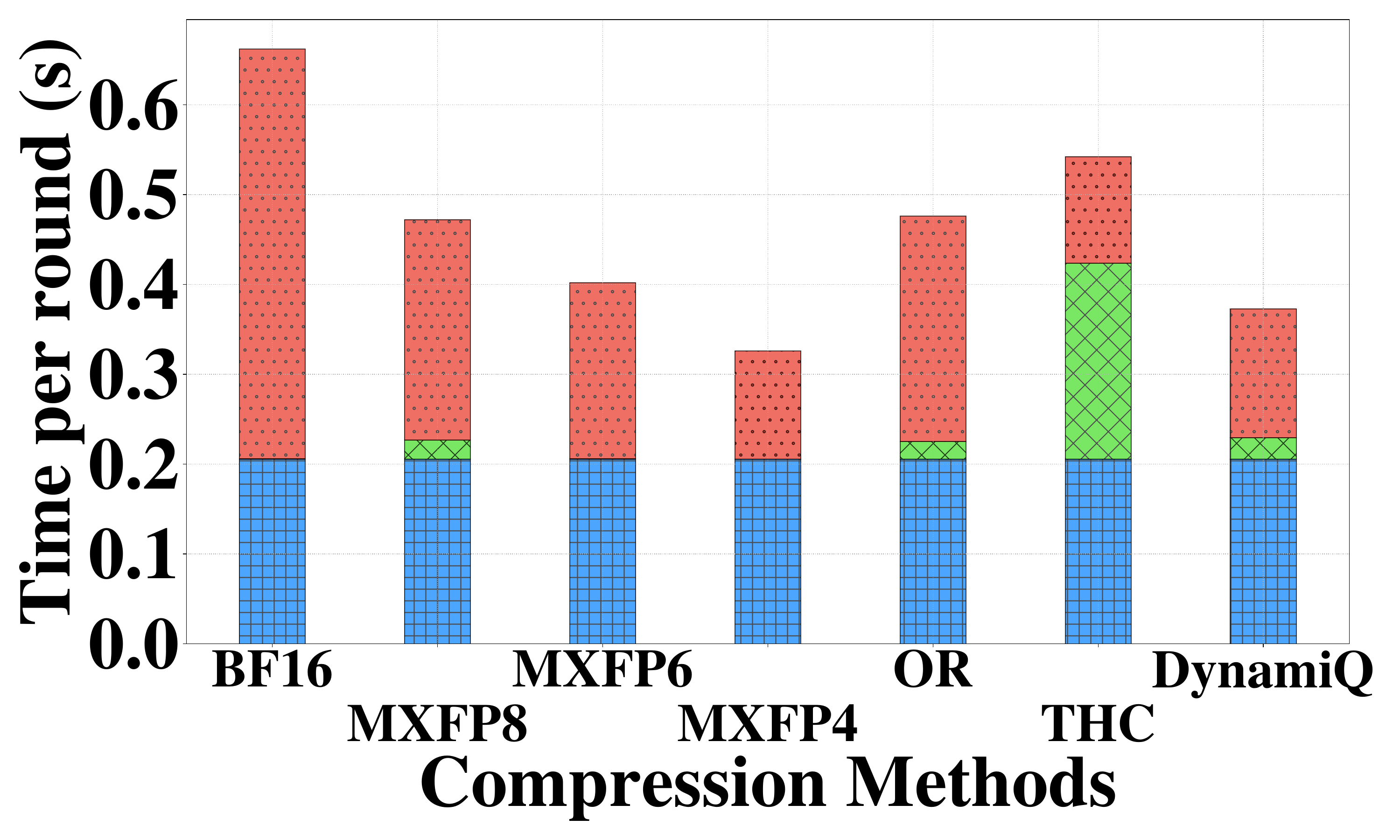}
		\end{center}
		}
		\label{subfig: breakdown-mmlu}
		\end{minipage}
	}
        \vspace{-0.3cm}
	\caption{Breakdown of training time into computation, exposed communication, and compression overhead. Compression refers to the extra time-per-round latency introduced with \remove{compression} \revise{all compression operations, including fused kernels and reordering costs}, that cannot overlap with computation. Exposed communication refers to the portion of communication latency not overlapped with computation or compression thus directly contributes to training duration.}
	\label{fig: time-breakdown}
\end{figure*}

\subp{Time to accuracy (TTA).} As highlighted in \Cref{fig:demo-time-to-accuracy} and further detailed in \Cref{fig: e2e-tta}, \sysname consistently produces better TTA curves across all workloads. For instance, on Gemma 1B, \sysname achieves target perplexity $18\%-28\%$ faster than MXFP8 and MXFP6 respectively, while THC, Omnireduce, and MXFP4 either converge slower than the BF16 baseline or do not reach target accuracy due to excessive compression error. Similarly, for LLaMA fine-tuning, \sysname reaches $72.38\%$ ($99\%$ of BF16's accuracy) accuracy approximately $34.5\%$ faster than MXFP8 and $40.8\%$ faster than BF16. Crucially, \sysname maintains final accuracy within $0.1\%$ of uncompressed BF16 across all scenarios, whereas other compression schemes exhibit degradation of up to $2.5\%$.

To shed more light on the performance of the different schemes, we next display their throughput and compression error measurements separately. Then, we show TTA curves for \sysname under different bit-budgets, clarifying our choice of five bits per-parameter.




\subp{Throughput.} The TTA gains of \sysname are partially attributable to its improved training throughput, as shown in Figure~\ref{fig: time-breakdown}. It is evident that the compression overhead of \sysname remains small, as gradient compression on GPUs is typically memory-bound rather than compute-bound~\cite{why-elementwise-memory-bound}. Accordingly, by leveraging fused kernels, \sysname ensures a coalesced memory access pattern where each gradient coordinate is accessed only once, maintaining parity with the memory transaction volume of MXFP8 (see Table~\ref{tab:memory-transaction}). In contrast, the randomized Hadamard transform used in THC~\cite{li2024thc} requires $O(\log d)$ additional global GPU memory accesses, creating a bottleneck that consumes up to $42\%$ of the training round time in the LLaMA 1B workload.


\begin{table}[]
    \centering
    \resizebox{0.8\linewidth}{!}{
    \begin{tabular}{|c|c|} \hline
        Compression scheme & Global Memory Transactions \\ \hline
        BF16 & $4 + 4 \cdot AR$ \\ \hline
        \sysname & $22 + 11.875 \cdot AR$ \\ \hline
        MXFP8 & $18 + 13 \cdot AR$ \\ \hline
        THC  & $74 + 2 \cdot AR$ \\ \hline
    \end{tabular}
    }
    \caption{Estimated extra DRAM memory transactions (bytes per coordinate) for different all-reduce compression schemes, excluding NIC-GPU data transfer. $AR=\frac{n-1}{n} \in [\frac{1}{2}, 1)$ denotes the per-worker data fraction transferred during reduce-scatter and all-gather phases.}
    \label{tab:memory-transaction}
\end{table}

\begin{table}[]
\resizebox{1.\linewidth}{!}{
    \begin{tabular}{|c|c|c|c|c|}
    \hline
    \multirow{2}{*}{Workload} & BERT-large & LLaMA 1B & Gemma 1B & LLaMA 1B \\ 
    & MaskedLM & Chat & Chat & MMLU \\ \hline
        \textbf{\sysname}  &  0.00217 & 0.00149 & 0.00122 & 0.00096 \\ \hline
        MXFP8 &  0.00591 & 0.00320 & 0.00308 & 0.00299 \\ \hline
        MXFP6 &  0.02332 & 0.01350 & 0.01458 & 0.01298 \\ \hline
        MXFP4 &  0.12080 & 0.11059 & 0.11583 & 0.09039 \\ \hline
        OR    &  0.15499 & 0.08044 & 0.04676 & 0.04530 \\ \hline
        THC   &  0.00897 & 0.11978 & 0.15168 & 0.19599 \\ \hline

    \end{tabular}
}
\caption{Comparing compression error (vNMSE) for LLM training (fine-tuning) on an 8-GPU/4-worker testbed using ring all-reduce. In this table, we compute the average vNMSE over the entire end-to-end training process. The full version of round-to-vNMSE curves is shown in~\Cref{fig: vnmse-e2e}. 
}
\label{tab: vNMSE-data}
\end{table}

\subp{Compression error.} As detailed in \Cref{tab: vNMSE-data}, \sysname demonstrates better fidelity to uncompressed gradients, achieving $2.5\text{--}3\times$ lower vNMSE than MXFP8 and orders-of-magnitude lower vNMSE than MXFP4, THC, and OmniReduce. These results further clarify the performance trade-offs observed earlier: while MXFP4 offers higher throughput (Figure~\ref{fig: time-breakdown}), its excessive error slows-down convergence and degrades final accuracy. Similarly, OmniReduce underperforms in these benchmarks because it relies on gradient sparsity and skewness (i.e., a large fraction of near-zero entries), an attribute largely absent in dense LLM gradients~\cite{zhao2024galore}. 


\begin{figure}[]
    \centering
    \begin{minipage}[t]{1\linewidth}{
		\vspace{-0.00in}
		\begin{center}
		\includegraphics[width=\textwidth, ]{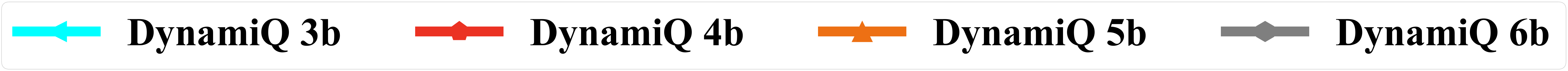}
		\end{center}
		}
        \end{minipage}

    \centering
    \hspace{-0.1in}
	\subfigure[The overall TTA curves]{
		\begin{minipage}[t]{0.45\linewidth}{
		\vspace{-0.00in}
		\begin{center}
		\includegraphics[width=\textwidth, ]{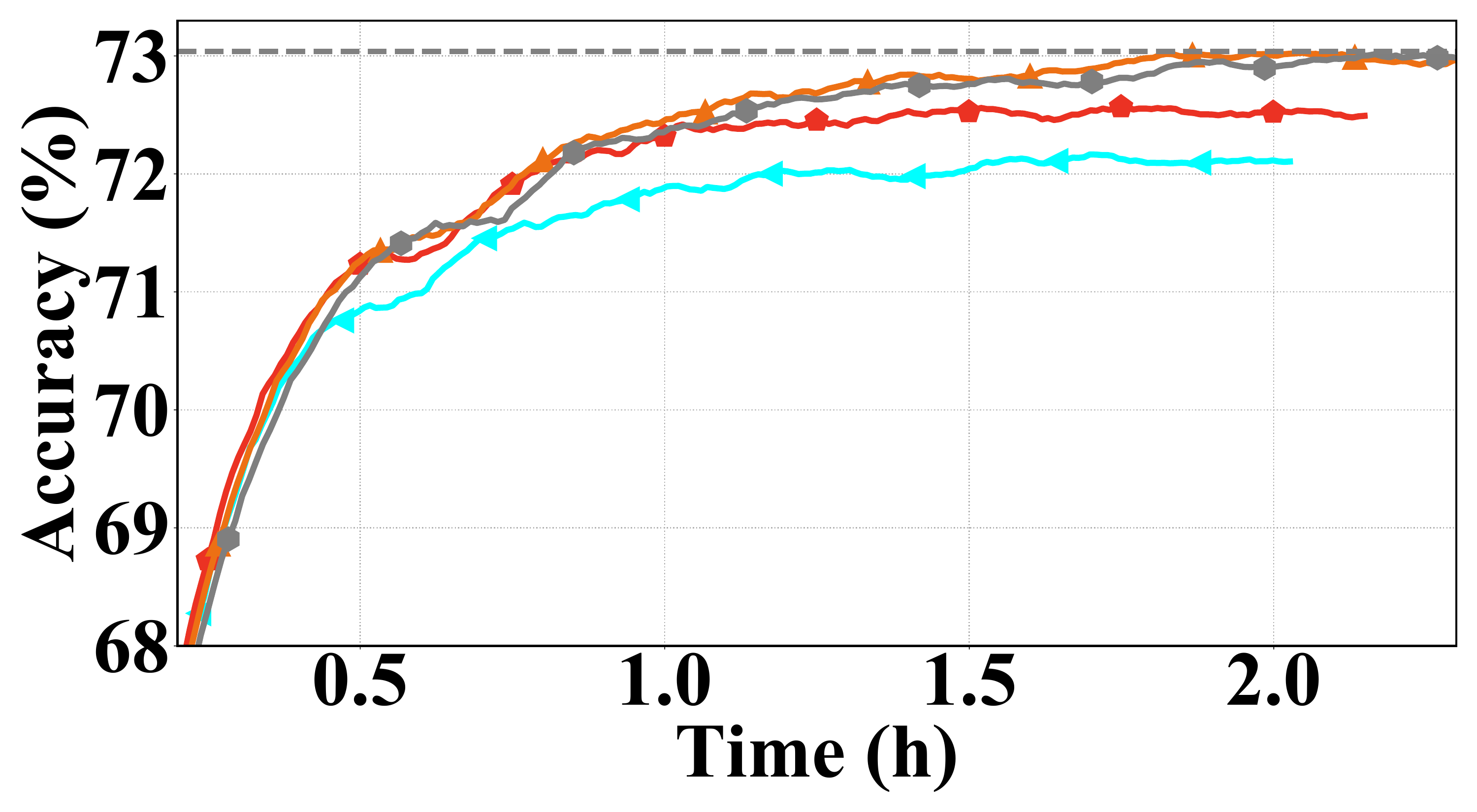}
		\end{center}
		}
		\label{subfig: ablation_nbits_mmlu}
		\end{minipage}
    }
    \subfigure[The zoomed-in TTA curves] {
        \begin{minipage}[t]{0.45\linewidth}{
		\vspace{-0.00in}
		\begin{center}
		\includegraphics[width=\textwidth, ]{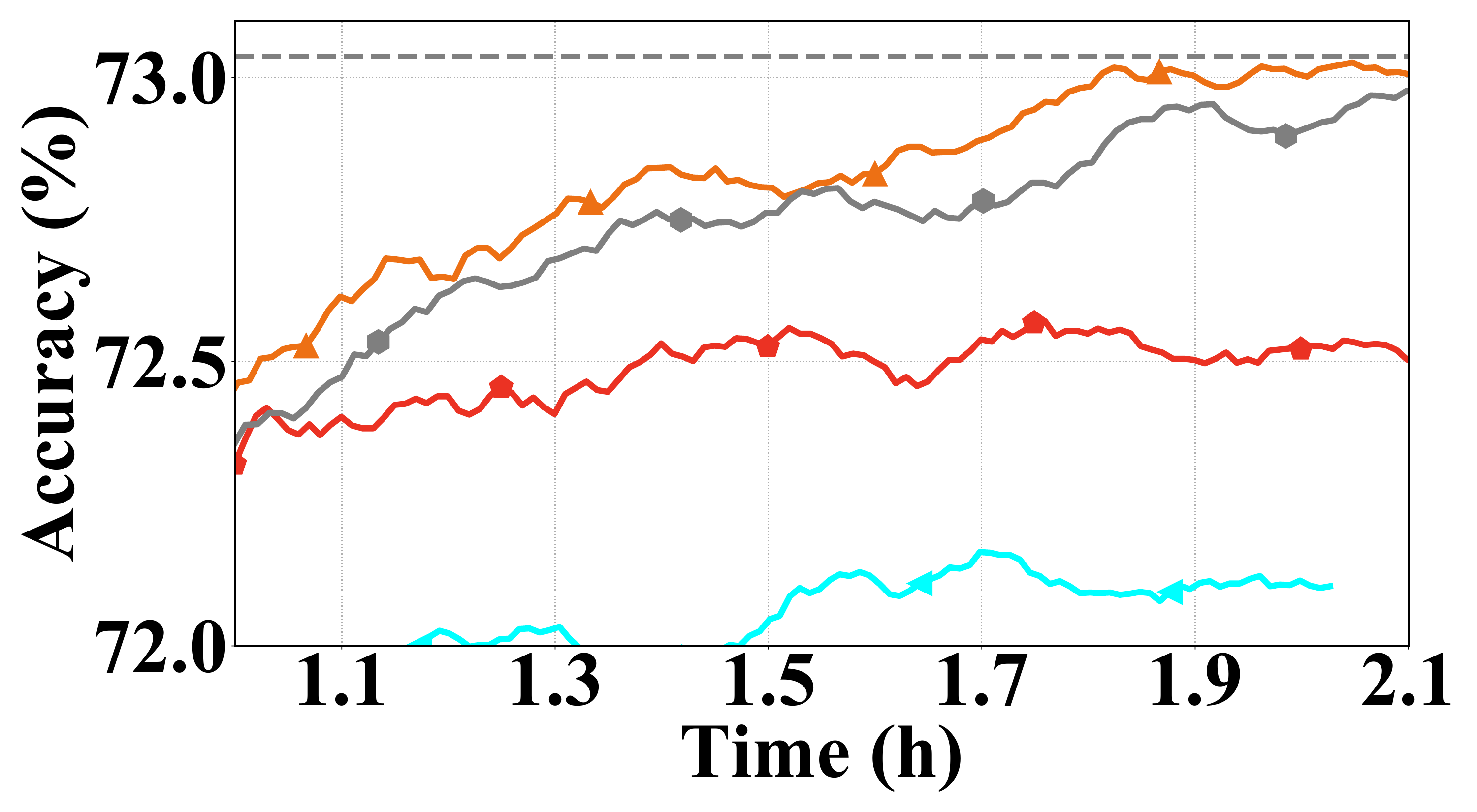}
		\end{center}
		}
		\label{subfig: ablation_nbits_mmlu_zoomed_in}
		\end{minipage}
    }
    \vspace{-0.2cm}
    \caption{DynamiQ’s bit-budget ablation. Displaying TTA curves for varying overall bit budget for training (fine-tuning) LLaMA 1B on the MMLU workload on an 8-GPU/4-worker testbed using ring all-reduce. The horizontal dashed lines represent the convergence accuracy of BF16.}
    \label{fig: ablation_nbits}
\end{figure}

\begin{table}[htbp]
    \centering
    \resizebox{0.9\linewidth}{!}{
    \begin{tabular}{|c||c|c||c|c|} \hline
        \multirow{2}{*}{Method} 
        & \multicolumn{2}{c||}{LLaMA 1B MMLU} 
        & \multicolumn{2}{c|}{Gemma 1B Chat} \\ \cline{2-5}
        & vNMSE & Thp. 
        & vNMSE & Thp. \\ \hline
        \sysname 3b & 0.01603 & 3.051 & 0.02334 & 1.440 \\ \hline
        \sysname 4b & 0.00589 & 2.842 & 0.00831 & 1.397 \\ \hline
        \sysname 5b & 0.00096 & 2.604 & 0.00122 & 1.353 \\ \hline
        \sysname 6b & 0.00059 & 2.390 & 0.00053 & 1.306 \\ \hline
        MXFP8 & 0.00299 & 2.123 & 0.00308 & 1.246 \\ \hline
    \end{tabular}
    }
    \caption{DynamiQ’s bit-budget ablation. Displaying Throughput (in rounds per second) and vNMSE for varying overall bit budget for training (fine-tuning) LLaMA 1B on the MMLU workload on an 8-GPU/4-worker testbed using ring all-reduce. Results for MXFP8 are displayed for comparison.}
    \label{tab:ablation-nbits}
\end{table}

\subp{\sysname's bit-budget ablation.} To justify our choice of using $b=5$ bit-per-coordinate for \sysname, we next evaluate the impact of $b$ on \sysname's TTA and convergence accuracy. Figure~\ref{fig: ablation_nbits} presents results for the LLaMA 1B MMLU workload, demonstrating that $b=5$ indeed achieves the best trade-off in this scenario. Reducing the bit budget below this threshold increases compression error and degrades final accuracy, while increasing $b$ yields no accuracy gains and, as detailed in Table~\ref{tab:ablation-nbits}, merely reduces throughput due to higher communication volumes.

\begin{figure}[]

    \centering
    \begin{minipage}[t]{0.9\linewidth}{
		\vspace{-0.00in}
		\begin{center}
		\includegraphics[width=\textwidth, ]{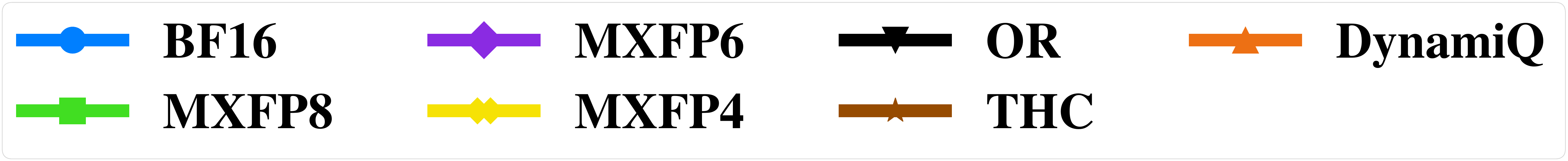}
		\end{center}
		}
        \end{minipage}

        \hspace{-0.2cm}
        \subfigure[Gemma 1B Chat]{
		\begin{minipage}[t]{0.49\linewidth}{
		\vspace{-0.00in}
		\begin{center}
		\includegraphics[width=\textwidth, ]{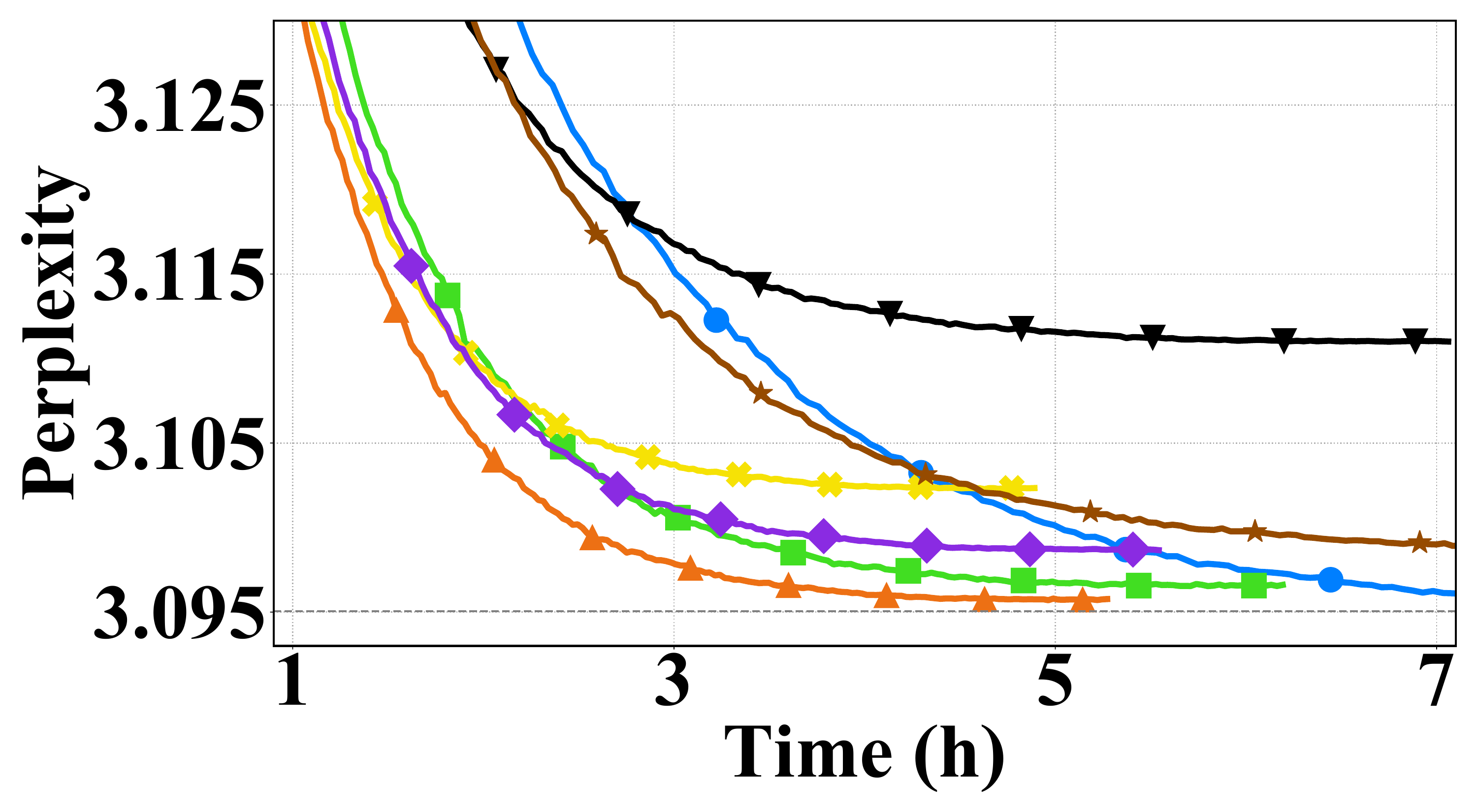}
		\end{center}
		}
		\label{subfig: e2e-gemma-flowcollision}
		\end{minipage}
	    }
        \hspace{-0.3cm}
        \subfigure[LLaMA 1B MMLU]{
            \begin{minipage}[t]{0.49\linewidth}{
            \vspace{-0.00in}
            \begin{center}
            \includegraphics[width=\textwidth, ]{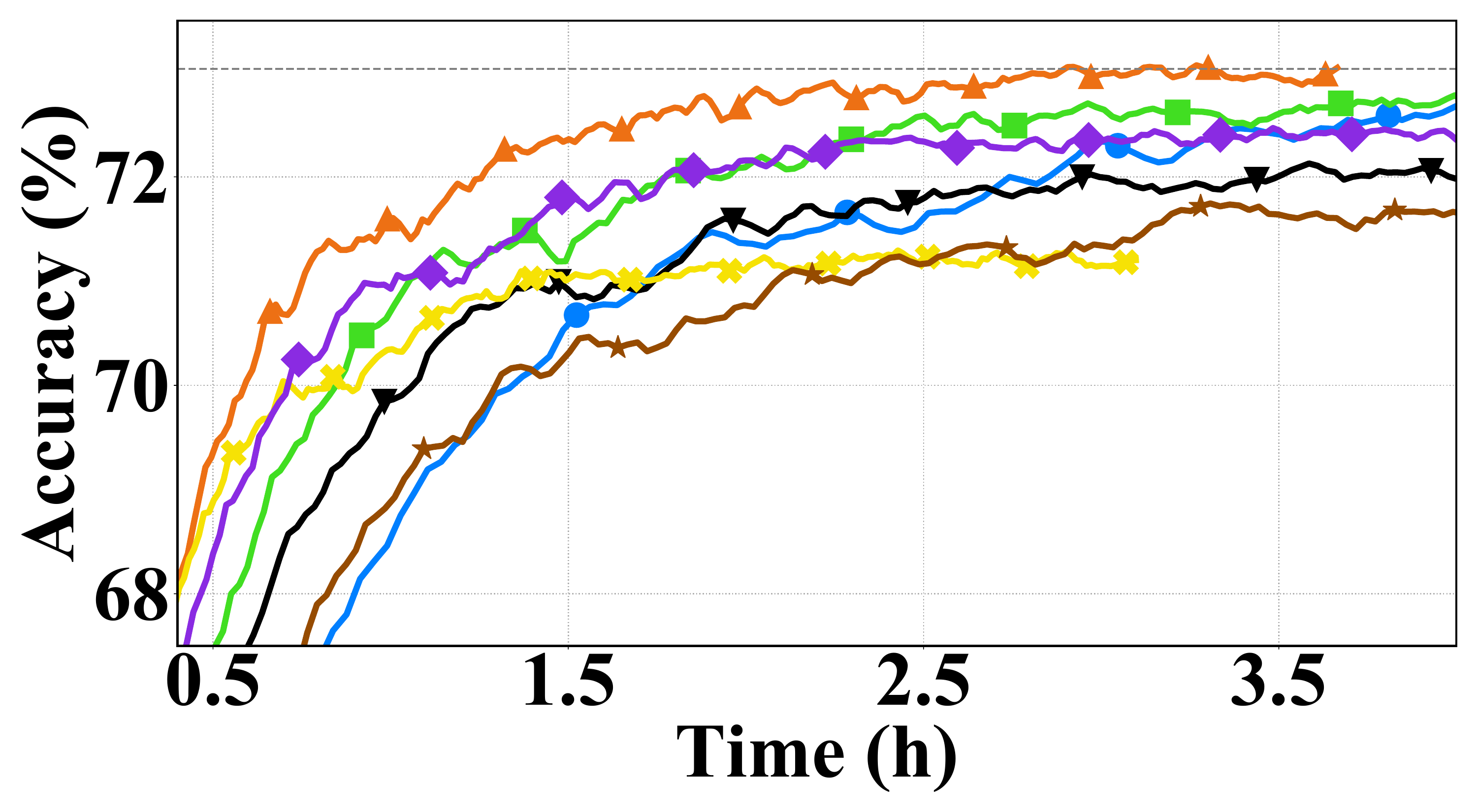}
            \end{center}
            }
            \label{subfig: e2e-mmlu-flowcollision}
            \end{minipage}
        }
        \vspace{-0.4cm}
    \caption{Zoomed-in TTA curves over a shared network. The full curves and additional results appear in \Cref{appendix:exp_results}.}
	\label{fig: e2e-tta-flowcollision}
\end{figure}

\subsection{Ring all-reduce over a shared network}\label{subsec:multi-tenancy}

In many setups, the training job may not run in isolation and has to share the network with other jobs or tenants.
For example, multi-tenancy in cloud providers is a common practice for maximizing GPU utilization~\cite{cao2024crux, hwang2021elastic}. 
%
Accordingly, in this experiment, we launch three additional DDP processes that continuously perform ring all-reduce operations, competing with the training job for bandwidth. 

Figure~\ref{fig: e2e-tta-flowcollision} shows, as expected, that the compression methods', and particularly \sysname's, TTA advantage over the BF16 baseline increases under bandwidth contention. 
For example, on Gemma 1B + Chat, \sysname's advantage over MXFP8 increases from $16\%$ in isolation to $21.5\%$ over a shared network. Likewise, in the LLaMA 1B + MMLU workload, the advantage increases from $34.5\%$ to $40.2\%$.
%
Interestingly, the exposed communication time is shorter than 4$\times$ the time in isolation, as the different jobs converge to transmitting only on partially overlapping timeframes.

Finally, in the interest of space, full TTA curves and additional results are deferred to in \Cref{appendix:exp_results}.

\begin{table}[]
    \centering
    \resizebox{0.6\linewidth}{!}{
    \begin{tabular}{|c||c|c|} \hline
        Method & Acc (\%) & vNMSE \\ \hline\hline
        BF16 & $73.04$ & $0$ \\ \hline\hline
        \textbf{\sysname} & $73.04$& $0.00067$ \\ \hline 
        MXFP8 & $72.86$ & $0.00203$ \\ \hline
        MXFP6 & $72.46$ & $0.02008$ \\ \hline
        MXFP4 & $71.59$ & $0.17058$ \\ \hline
        
    \end{tabular}
    }
    \caption{Evaluation with butterfly all-reduce on the LLaMA 1B MMLU workload. We list the final accuracy the model converges, to relative to the BF16 baseline, and average quantization error (vNMSE).}
    \label{tab:eval-butterfly}
\end{table}

\begin{figure}[]
    \centering
    	\vspace{-0.4cm}
    \begin{minipage}[t]{0.7\linewidth}{
		\begin{center}
		\includegraphics[width=\textwidth, ]{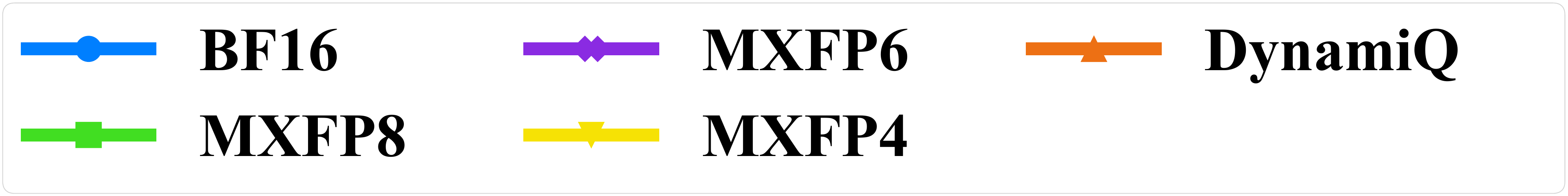}
		\end{center}
		}
        \end{minipage}

    \centering
    \hspace{-0.1in}
	\subfigure[LLaMA 1B MMLU]{
		\begin{minipage}[t]{0.6\linewidth}{
		\vspace{-0.00in}
		\begin{center}
		\includegraphics[width=\textwidth, ]{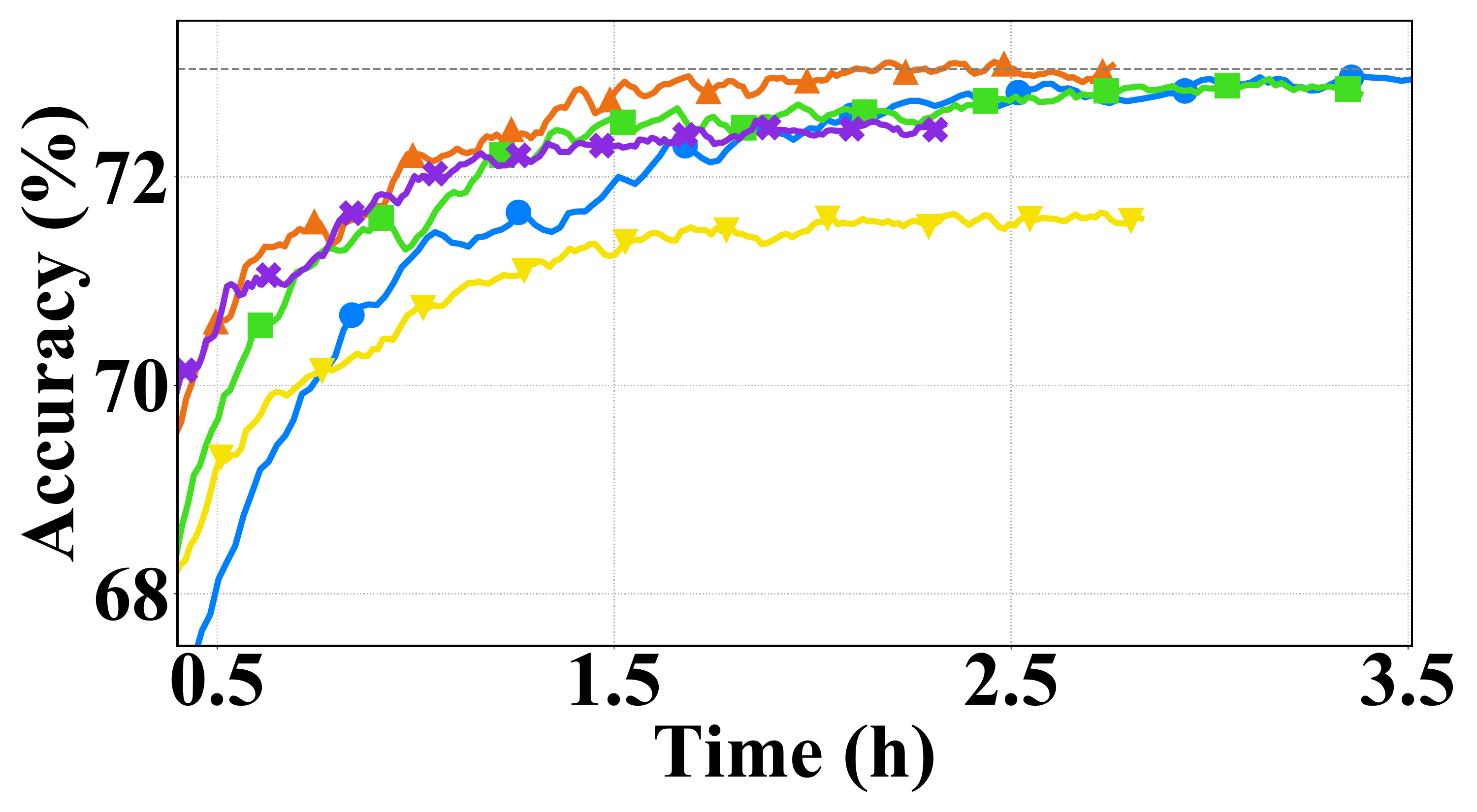}
		\end{center}
		}
		\label{subfig: butterfly_mmlu}
		\end{minipage}
    }
    \vspace{-0.3cm}
    \caption{The zoomed-in TTA of \sysname with butterfly all-reduce compared with the baselines. The full version is shown in Figure~\ref{fig: overall-tta-butterfly} in the Appendix.}
    \vspace{-0.1cm}
    \label{fig: butterfly_mmlu}
\end{figure}

\subsection{Butterfly all-reduce}\label{subsec:butterfly}
We proceed with an experiment with butterfly all-reduce~\cite{thakur2005optimization}, which reduces the number of hops to logarithmic in the number of workers, thereby reducing latency. Interestingly, it also reduces quantization error, as fewer re-quantizations are needed and the summed partial sums on the aggregation path tend to have a closer order of magnitude~\cite{patarasuk2009bandwidth}.


Figure~\ref{fig: butterfly_mmlu} depicts that, on the LLaMA 1B MMLU benchmark, \sysname achieves better TTA and, in particular, higher final accuracy than the MXFP4, MXFP6 and MXFP8 baselines. Specifically, \sysname attains an accuracy of $72.38\%$ -- corresponding to $99\%$ of BF16’s final accuracy—$12.0\%$ faster than MXFP8; this advantage further increases to $37.8\%$ when targeting $99.5\%$ of BF16’s final accuracy. In addition, the microscaling baselines exhibit measurable degradation in final accuracy (\Cref{tab:eval-butterfly}), whereas \sysname achieves a final accuracy comparable to BF16. 
This is explained by \sysname's lower vNMSE (quantization error), as~\Cref{tab:eval-butterfly} shows.

Lastly, we argue that this trend is expected to continue as the number of workers increases and provide theoretical intuition to support this in~\Cref{sec:adaptation-butterfly}.


\begin{figure}
    \hspace{0.8in}
    \begin{minipage}[t]{0.9\linewidth}{
		\vspace{-0.00in}
		\begin{center}
		\includegraphics[width=\textwidth, ]{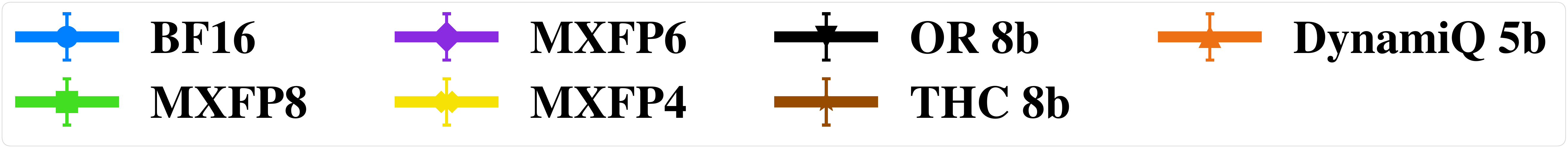}
		\end{center}
		}
        \end{minipage}
        \hspace{-0.1in}
        \subfigure[LLaMA 1B MMLU vNMSE]{
		\begin{minipage}[t]{0.485\linewidth}{
		\vspace{-0.00in}
		\begin{center}
		\includegraphics[width=\textwidth, ]{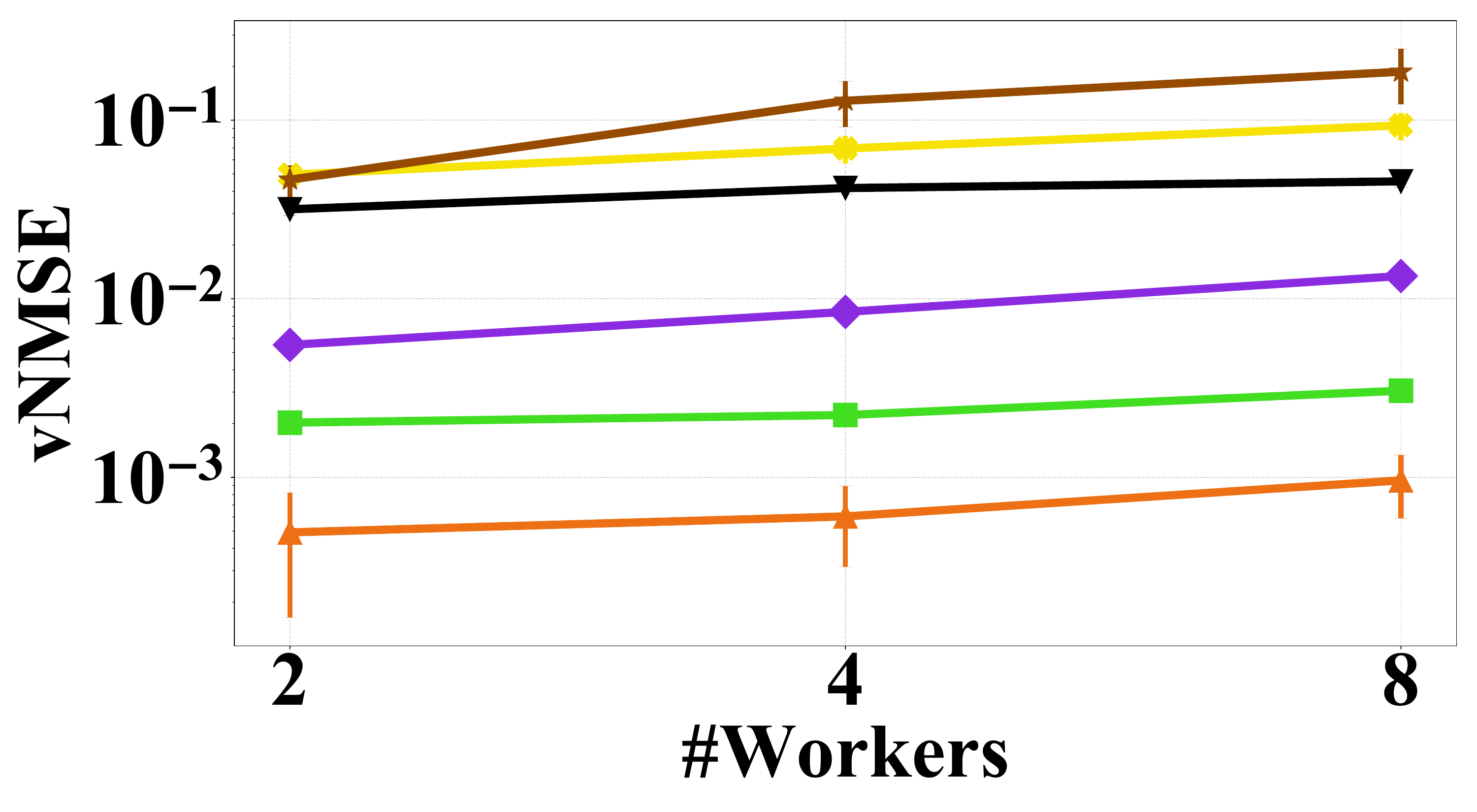}
		\end{center}
		}
		\label{subfig: scalability-mmlu_error}
		\end{minipage}
	}
        \hspace{-0.1in}
	\subfigure[LLaMA 1B MMLU $\Delta$Accuracy]{
		\begin{minipage}[t]{0.485\linewidth}{
		\vspace{-0.00in}
		\begin{center}
		\includegraphics[width=\textwidth, ]{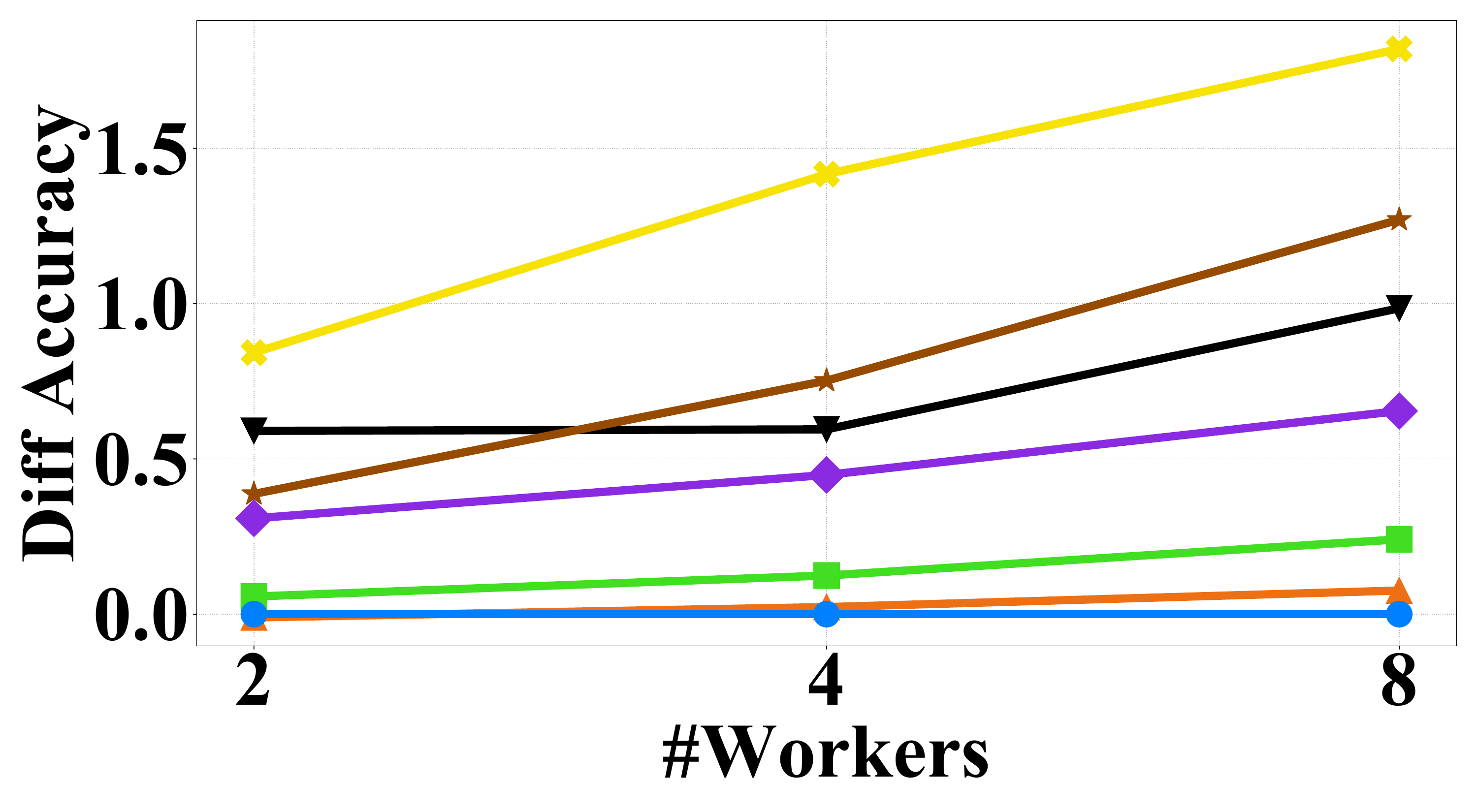}
		\end{center}
		}
		\label{subfig: scalability-mmlu-acc}
		\end{minipage}
	}

        \vspace{-0.2cm}
	
	\caption{Scalability evaluation on the LLaMA + MMLU task with $2$ to $8$ workers, measuring the vNMSE and the MMLU's accuracy with respect to the BF16 baseline. }
	\label{fig: scalability-small}
\end{figure}

\begin{figure}[]
    \centering
    \begin{minipage}[t]{0.9\linewidth}{
		\vspace{-0.00in}
		\begin{center}
		\includegraphics[width=\textwidth, ]{exp_figures/scalability_error_legend.pdf}
		\end{center}
		}
        \end{minipage}

	\hspace{-0.08in}
        \subfigure[TinyBERT GLUE vNMSE]{
		\begin{minipage}[t]{0.49\linewidth}{
		\vspace{-0.00in}
		\begin{center}
		\includegraphics[width=\textwidth, ]{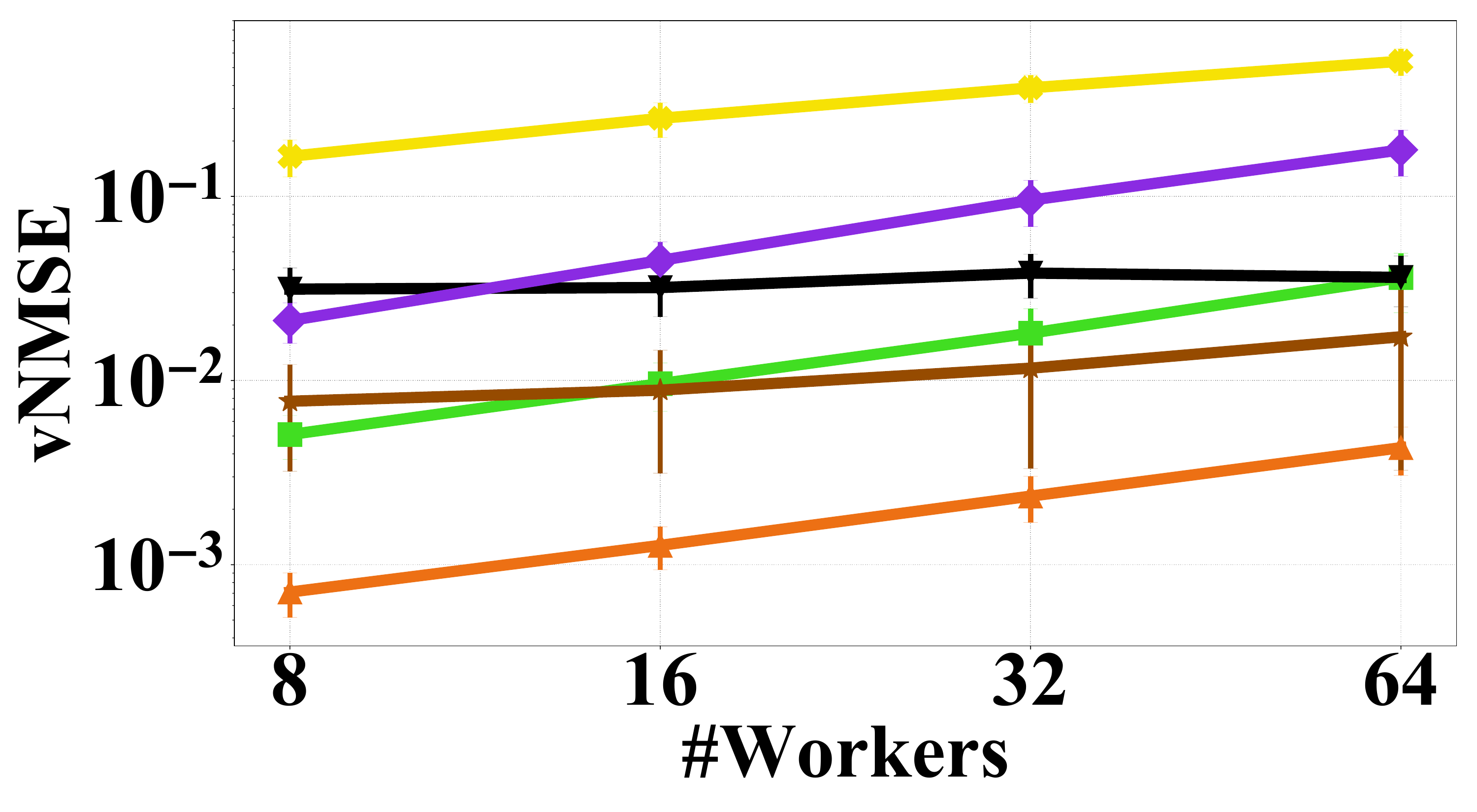}
		\end{center}
		}
		\label{subfig:scalability-vnmse-fixed}
		\end{minipage}
	}
        \hspace{-0.1in}
        \subfigure[TinyBERT GLUE $\Delta$CE loss]{
		\begin{minipage}[t]{0.485\linewidth}{
		\vspace{-0.00in}
		\begin{center}
		\includegraphics[width=\textwidth, ]{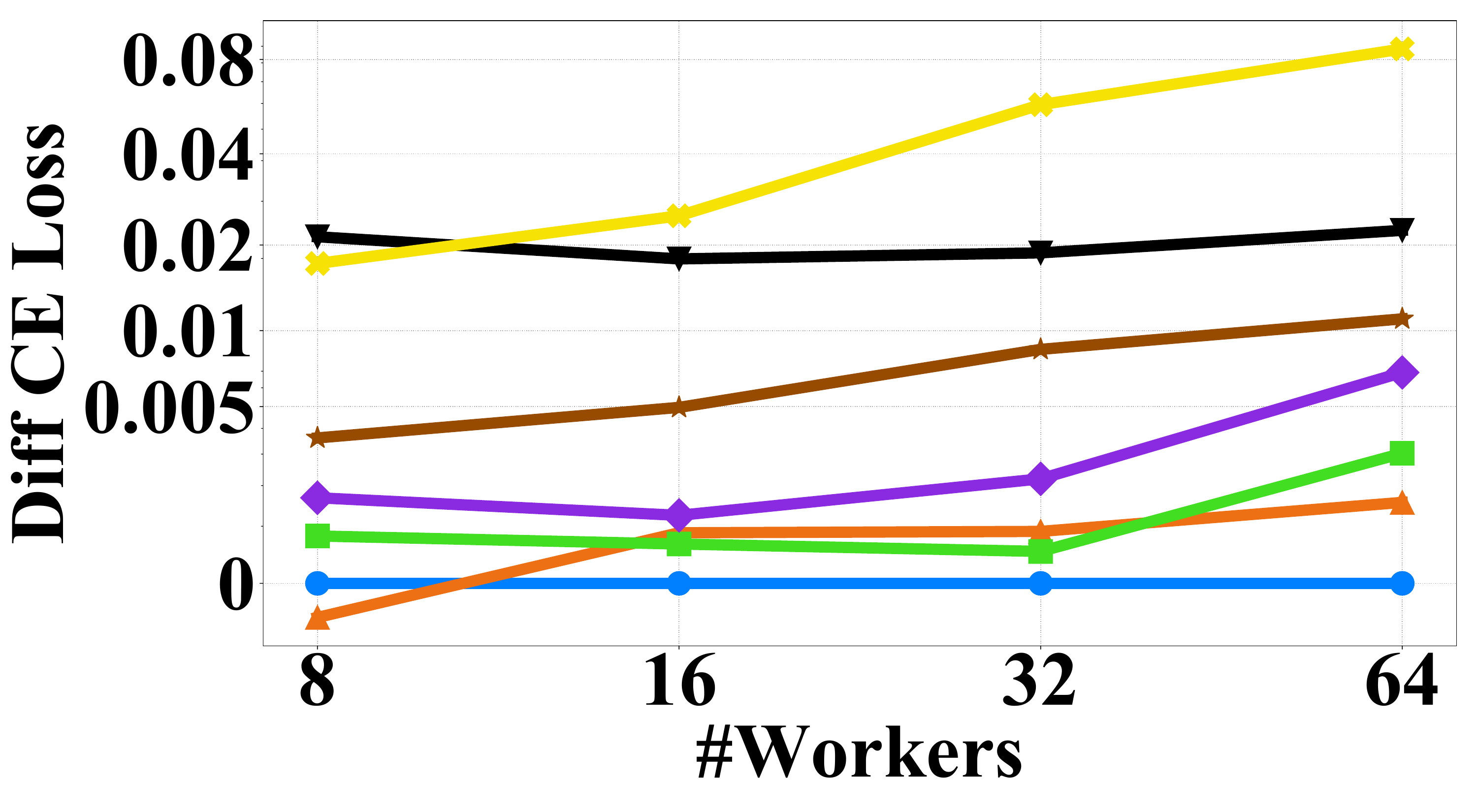}
		\end{center}
		}
		\label{subfig: scalability-ce-fixed}
		\end{minipage}
        }
    \vspace{-0.4cm}
    \caption{Scalability evaluation on the TinyBERT + GLUE task with $8$ to $64$ workers, measuring vNMSE and the cross entropy loss with respect to the BF16 baseline.}
    \label{fig: scalability-large}
\end{figure}

\section{Simulation Studies}\label{sec:simulation}
We next test the scalability of \sysname and perform a parametric study to shed more light on its individual design components.


\subsection{Scalability analysis}\label{subsec:scalability}

\subp{Setup and Methodology.}
We evaluate scalability by varying the worker count $n$ from 2 to 64 across two workloads: LLaMA 1B MMLU (2--8 workers) and the smaller TinyBERT~\cite{jiao2019tinybert} on GLUE~\cite{wang2018glue} (8--64 workers). In all experiments, we utilize ring all-reduce and benchmark performance against the BF16 baseline, measuring quantization error (vNMSE) and final accuracy (LLaMA 1B) and cross-entropy (CE) loss (TinyBERT). For THC, we adopt the authors' recommendation to allocate 12 bits for $n > 8$ to prevent gradient overflow during aggregation.

\subp{LLaMA 1B.}
As the worker count increases, both the vNMSE and the accuracy degradation naturally increase across all methods. However, as shown in Figure~\ref{fig: scalability-small}, \sysname shows better scaling properties compared to the baselines nearing the accuracy of BF16 even with 8 workers. 

\subp{TinyBERT.}
Extending the analysis to larger clusters with TinyBERT, Figure~\ref{fig: scalability-large} confirms that \sysname consistently achieves the lowest vNMSE among all compression schemes up to 64 workers. Consequently, it yields the final accuracy closest to the BF16 baseline (Figure~\ref{subfig: scalability-ce-fixed}). We note that inherent training variance in small models leads to minor fluctuations. For example, \sysname slightly outperforms BF16 at $n=8$ or MXFP8, showing marginally lower CE loss at $n=16, 32$. However, the overall trend confirms that \sysname's more scalable and \mbox{stable than other compression methods.}

Finally, we observe that, as illustrated in Figure~\ref{subfig:scalability-vnmse-fixed}, THC and OR display slower vNMSE growth as $n$ scales. For THC, this results from increasing the allocation from 8 to 12 bits for $n > 8$ to prevent overflows (satisfying $b \ge \lceil \log (15n+ 1) \rceil$). However, as a strategy, this remains effective only up to $n=64$. For OmniReduce ($b=8$), the error profile at this scale is determined by its sparsification policy, which consistently discards the bottom $50\%$ of gradients.

\begin{table}[]
    \centering
    \resizebox{1\linewidth}{!}{
    \begin{tabular}{|c|c|c|} \hline
        Method & LLaMA 1B Chat & LLaMA 1B MMLU \\ \hline
        Uniform quantization & 0.1278 & 0.1207 \\ \hline
        Non-uniform quantization & 0.0707  & 0.0664 \\ \hline
        + Variable bitwidth allocation & 0.0198 & 0.0130  \\ \hline
        + Hierarchical quantization & 0.0138 & 0.0092  \\ \hline
        + Correlated rounding & 0.0091  & 0.0059  \\ \hline
    \end{tabular}
    }
    \caption{vNMSE comparisons among different variants of \sysname for the LLaMA 1B Chat and LLaMA 1B MMLU workloads. For hierarchical quantization, we use a group size of $16$ instead of $32$.}
    \vspace{-0.2cm}
    \label{tab:ablation-merged}
\end{table}

\subsection{\revise{Large-scale simulation}}
\label{subsec:eval-extreme-large-scale}

\begin{table}[t]
    \centering
    \resizebox{0.9\linewidth}{!}{%
    \begin{tabular}{|l||c|c|c|c|} \hline
       Number of bits $\bar{b}$ & 5 & 6 & 8.5 \\ \hline\hline
       DynamiQ-ring
       & $4.76$
       & $0.751$
       & $0.0105$ \\ \hline
       MXFP8-ring
       &  -- & -- & $6.11$ \\ \hline\hline
       DynamiQ-butterfly
       & $0.0336$
       & $3.23 \times 10^{-3}$
       & $2.75 \times 10^{-5}$ \\ \hline
       MXFP8-butterfly
       &  -- & -- & $0.0353$ \\ \hline
    \end{tabular}%
    }
    \caption{\revise{Compression error (vNMSE) in a simulation with synthetic data and a DP dimension of $n=8{,}192$, and a total of $32768$ super-groups of $256$ coordinates per worker.}}
    \label{tab:large-scale}
\end{table}

\revise{
To evaluate \sysname at a large scale, we simulate compressed all-reduce with a DP dimension of $n=8{,}192$. 
To model the spatial locality and scale skew observed in~\Cref{subsec:background-grouped-quantization,fig:demo-per-group-vs-shuffled}, we assign each super-group a magnitude parameter, which determines the variance of all coordinates in that super-group.
}

\subp{Synthetic data.}
\revise{We model the magnitude $\mathit{M_{\mathcal G}}$ of a super-group $\mathcal G$ as a mixture of two LogNormal distributions; this captures well the empirical distribution of~\Cref{subfig:demo-original-CDF-per-supergroup-vs-shuffled}, as described in~\Cref{app:synthetic_data}. 
For each $\mathcal G$ (of size 256), we sample each of its coordinates, for each worker, i.i.d. from
$\mathcal{N}(0,\frac{\mathit{M_{\mathcal G}}^2}{256})$. 
Consequently, coordinates within a super-group share a common scale, preserving spatial locality, while scales across \mbox{super-groups span several orders of magnitude.}
}

\subp{Setup.}
\revise{
We evaluate communication budgets of $\bar{b}\in\{5,6,8.5\}$ bits per coordinate under both ring and butterfly all-reduce. We compare \sysname against \texttt{MXFP8-e4m3} (we omit other baselines since they overflow in this setting, yielding a prohibitively large vNMSE). Because \texttt{MXFP8-e4m3} has a fixed communication cost of $\bar{b}=8.5$, we report it only at that budget. For \sysname, we allow bitwidths from $W=\{2,4,8,16\}$.
}

\subp{Results.}
\revise{
As shown in~\Cref{tab:large-scale}, \sysname achieves substantially lower vNMSE than MXFP8. With ring all-reduce and $\bar{b}=8.5$, \sysname achieves a vNMSE of $0.0105$, which is $\approx 582$ times lower than MXFP8. On the other hand, \sysname's vNMSE with $5$ bits roughly matches with MXFP8 that consumes $\bar{b}=8.5$ bits. Similarly, with butterfly all-reduce and $\bar{b}=8.5$, \sysname achieves a vNMSE of $2.75\times10^{-5}$, which is $1283$ times lower than MXFP8. We note that at this large DP dimension, MXFP8-ring and \sysname-ring with $\bar{b} \in \{5,6\}$ incur unacceptably large vNMSE. Thus, for \sysname-ring with this DP dimension, we recommend using $\bar{b}=8.5$, where for \sysname-butterfly, $\bar{b}=6$ suffices.
}

\revise{
We partially attribute these gains to two features of \sysname. First, its bit-allocation scheme assigns up to 16 bits to rare, high-norm super-groups without imposing this cost on all coordinates. Second, its decompress--accumulate--recompress procedure updates the group scales as the partial sums evolve, allowing the quantization range to track the data well throughout the all-reduce. In contrast, \texttt{MXFP8-e4m3} reserves four of its eight per-value bits for the exponent to represent a wide dynamic range uniformly.
}

\subsection{Parametric study}\label{subsec:ablation}

We next isolate the impact of \sysname's optimization components, namely, variable bitwidth allocation, non-uniform quantization, hierarchical quantization, and correlated rounding, on compression error. The group size is set to 32 and reduced to 16 when hierarchical quantization (with INT8 scaling parameters) is used. 

Table~\ref{tab:ablation-merged} demonstrates that the cumulative application of these techniques reduces vNMSE by a factor of $14\times$ for LLaMA 1B Chat and $22\times$ for MMLU. Variable bitwidth allocation serves as the primary driver, improving quantization accuracy by $3.5\text{--}5.1\times$. Complementary techniques provide significant additive gains: non-uniform quantization reduces vNMSE by $\sim 45\%$ (see Appendix Figure~\ref{fig:comparison-benchmark-non-uniform}), hierarchical quantization by $\sim 30\%$, and correlated rounding by $\sim 35\%$. As discussed in Section~\ref{subsec:end-to-end-eval}, this order-of-magnitude reduction in error is essential for maintaining model accuracy comparable to uncompressed baselines. Crucially, as was shown in \Cref{fig: time-breakdown}, these enhancements introduce only a small computational overhead.

\vspace{-0.cm}
\section{Related Work}\label{sec:related}

\subp{Gradient Compression and the Shift to all-reduce.}
Gradient compression is a well-established strategy for accelerating distributed data-parallel (DDP) training by mitigating communication bottlenecks~\cite{terngrad, li2014scaling, sapio2021scaling, desensi2024swing}. While many such methods have been proposed~\cite{alistarh2017qsgd, bai2021gradient, bernstein2018signsgd, fei2021efficient, kim2019parallax, li2024accelerating, li2024thc, m2021efficient, topk, powersgd, wang2018atomo, wang2023hi, wang2023cupcake, chen2024justintime}, these were designed for the parameter server architecture~\cite{li2014scaling, jiang2020unified}. Indeed, the recent paradigm shift toward multi-hop all-reduce for scaling LLM training~\cite{ring, desensi2024swing, grattafiori2024llama3herdmodels, topology-at-scale, hammingmesh2022} reveals significant limitations in these approaches. For instance, sparsity-based methods like OmniReduce~\cite{fei2021efficient} struggle to merge local TopK chunks efficiently across decentralized topologies. Similarly, quantization schemes such as \revise{QSGD~\cite{alistarh2017qsgd},} THC~\cite{li2024thc}, and Terngrad~\cite{terngrad} are prone to gradient overflow during the aggregation of partial sums, a fundamental issue in multi-hop topologies that worsens with system size~\cite{han24hotnets}. While microscaling-based methods (\eg, MXFP4)~\cite{opencompute, peng2023fp8} alleviate this, they do not eliminate overflow/underflow entirely. By contrast, \sysname is explicitly architected for multi-hop all-reduce, employing hop-wise decompression/recompression to strictly prevent overflow and utilizing variable \mbox{bitwidth allocation to ensure robustness.}

\subp{Compression Error and Scalability.}
Although LLMs exhibit some tolerance for compression noise, excessive error significantly degrades convergence stability and final accuracy~\cite{lee2024fp8, lee2023training}. Existing schemes often prioritize inference hardware compatibility or sparsity over minimizing the error (vNMSE). For example, microscaling techniques~\cite{rouhani2023microscaling} optimize for GPU throughput but lack advanced error-reduction mechanisms, while OmniReduce relies on gradient sparsity that is largely absent in dense LLM updates. Furthermore, maintaining bounded error as the worker count $n$ increases presents a significant challenge; errors accumulate hop-by-hop, typically necessitating a bitwidth growth that is proportional to the logarithm of the aggregation path length (as in THC) to prevent overflow. We empirically observe that for \sysname this growth is slower, but leave further investigation for future work.

\subp{Hardware-Aware Implementation.}
Gradient compression on GPUs is predominantly memory-bound rather than compute-bound; performance is dictated by HBM bandwidth rather than floating-point throughput~\cite{memory-bound, gholami2024ai, memory-bound2, why-elementwise-memory-bound}. Consequently, efficient implementations must minimize HBM transactions, ideally ensuring sequential, single-pass access via kernel fusion~\cite{kernel-fusion}. Methods that fail to respect this constraint incur substantial overhead. A notable example is THC, where the Hadamard transform~\cite{hedayat1978hadamard} requires $O(\log d)$ passes over memory, creating a bottleneck. \sysname avoids such overhead by leveraging fused kernels that keep intermediate results in registers or shared memory, maintaining a memory access pattern comparable to standard uncompressed updates.

\subp{Mixed-precision training.}
An emerging technique to accelerate training is using lower precision arithmetics~\cite{castro2025quartet, panferovquest, 10.5555/3692070.3694598, nvfp4, efficientqat, wang2025optimizing, pmlr-v258-tseng25a, Hack}. This is motivated by new hardware capabilities that deliver higher throughput for low-precision operations~\cite{blackwell}.
Current best practices keep certain fields (e.g., outlier values or accumulators) in higher precision while using low precision elsewhere~\cite{wang2025optimizing}. Recently, researchers proposed running the entire training process in low precision~\cite{castro2025quartet, pmlr-v258-tseng25a}, often leading to a degradation of accuracy that can be acceptable in certain scenarios.

\subp{Sharded models.}
  When models are too large to fit on a GPU, practitioners shard them across multiple workers, each holding a portion~\cite{rajbhandari2020zero,zhao2023pytorch}. In such cases, one may not need an all-reduce operation but rather only the reduce-scatter phase, since gradients and weights are split across GPUs. \sysname can seamlessly integrate with this approach by decompressing at the end of the reduce-scatter phase.

\section{Discussion and Limitations}

\subp{Implications for large-scale pre-training.}
\revise{
Our testbed evaluation focuses on fine-tuning due to limited hardware resources, but it already covers DP dimensions up to $64$. In real pre-training clusters, this can correspond to much larger total GPU counts, since 3D parallelism keeps the DP dimension substantially smaller than the total number of GPUs; for example, even clusters with up to $100$K GPUs may still operate with comparable DP dimensions~\cite{chu2025scalingllama3,llama3herd,narayanan2021efficient}. Thus, our testbed results are directly relevant to realistic large-scale pre-training deployments.
For even larger deployments, such as future million-GPU clusters, scaling \sysname primarily requires either increasing the bit budget or using hierarchical aggregation. Our simulations show that \sysname remains accurate at DP dimension $8192$ with an $8.5$-bit budget, suggesting that modestly larger budgets can support much larger DP groups. Alternatively, hierarchical butterfly- or hypercube-style aggregation, which is commonly needed at such scales to bound tail latency, can also reduce the effective aggregation depth.
}

\subp{Preserving model quality and numerical stability.}
\revise{For large-scale pre-training tasks, practitioners are often conservative about low-precision compression schemes~\cite{fishman2025scalingfp8, fujii2024balancing}, because compression error can manifest as hidden instability that harms final model quality or convergence. For example, occasionally quantizing extreme values into an unsafe low-precision representation can lead to NaNs, causing model divergence, or otherwise substantially degrade accuracy. \sysname's super-group bitwidth allocation scheme (\Cref{subsec:design-super-group-bitwidth}) naturally supports preserving such outlier values in the training precision, e.g., BF16.}

\subp{Compatibility with FSDP and MoE~\cite{shazeer2017outrageously} gradients.}
\revise{\sysname is compatible with FSDP-style synchronization as FSDP decomposes synchronization into reduce-scatter and all-gather phases, to which \sysname's multi-hop communication primitive can be applied as described in \Cref{subsec:design-main-allreduce}. For MoE gradients, where experts can be active or inactive, \sysname's bit allocation can assign $0$ bits to inactive expert super-groups while allocating \mbox{more bits to active ones.}}

\subp{Implications for high-end training hardware.}
\revise{Our hardware evaluation uses A6000 GPUs with 768~GB/s GPU memory bandwidth and 100~Gb/s inter-node bandwidth. Modern clusters with, e.g., GB200-class GPUs provide faster scale-out networks, such as 400~Gb/s or 800~Gb/s links~\cite{nvidia_connectx7_400g_adapters,nvidia_connectx7_user_manual}, but also $16 \times$ higher FP16/BF16 FLOPS and $10.4 \times$ memory bandwidth~\cite{nvidia_gb200_nvl72}. Therefore, communication remains a bottleneck, especially under large data-parallel dimensions or shared-cluster contention.}

\section{Conclusion}

In this paper, we presented \sysname, a practical gradient compression framework optimized for multi-hop all-reduce that can adjust to different bandwidth constraints and presents an attractive tradeoff between communication overhead and accuracy. In contrast with existing gradient compression systems, which are designed for the parameter-server architecture and incur accuracy degradation when deployed to multi-hop all-reduce, \sysname preserves low compression error along the aggregation paths, resulting in accelerated training without compromising model accuracy.

We implement \sysname and evaluate its performance across diverse LLM training workloads using both ring and butterfly all-reduce. Our results show that \sysname consistently achieves significantly better time-to-accuracy compared to alternatives. Notably, \sysname reaches 99.9\% of BF16 baseline accuracy with only 5 bits per coordinate, outperforming the state-of-the-art MXFP8. It is the \emph{only} evaluated method to consistently maintain this fidelity while providing significant acceleration, a result driven by its fast, co-designed fused CUDA kernels.
%
%
\remove{We plan to open-source our implementation upon publication.}\revise{Our code is available at~\cite{open-source}.} 

\revise{
\section*{Acknowledgments} We thank the anonymous reviewers and our shepherd for their insightful feedbacks.
Michael Mitzenmacher was supported in part by grants NSF CNS-2107078 and NSF DMS-2023528. Ran Ben Basat was supported in part by the Google
Research Scholar Award.
}

\newpage

\bibliographystyle{ACM-Reference-Format}
\bibliography{sigcomm}

\appendix



\section{A Faster Solution for Variable Bitwidth Allocation}\label{app:variable-bitwidth-approx} 

We propose a faster solution that dynamically maintains and adjusts an approximate value of $T_{a, b}$ across training rounds, assuming that there are at most three possible bitwidths. We explain the algorithm for the setup used in \sysname's prototype, where the allowed bitwidth is $W=\{2, 4, 8\}$. 
To accelerate the calculations, we avoid sorting the array of $F_j$ values (which is needed by the algorithm described in \Cref{subsec:design-super-group-bitwidth}) and instead calculate how many bits $q_j$ each super-group $j$ is quantized by the following equation:

$$q_j = 2^{\text{clamp}\parentheses{[1, 3], \floor{\log_2(\frac{4}{\log_2 (512/17)} \log_2 F_j + u)}}}.$$ 

Recall that the thresholds need to satisfy $T_{2,4}=17/512 \cdot T_{4,8}$ and the bandwidth constraint, which is $S\cdot \sum_j q_j \le d \cdot \bar{b}$. Here, $\bar{b}$ represents the overall bitwidth budget minus the per-entry bandwidth used for transmitting the metadata, etc. 

We now explain why this equation respects these constraints and how we determine $u$. For a constant $u$, let $$z_j = \frac{4}{\log_2 (512/17)} \log_2 F_j + u.$$ Observe that $q_j = 2$ if $z_j < 4$, $q_j = 4$ if $z_j \in [4, 8)$ and $q_j = 8$ if $z_j > 8$. Since $T_{a,b}$ is defined as the threshold for which super-groups $j$ with $F_j\ge T_{a,b}$ are assigned with at least $a$ bits and super-groups with $F_j\le T_{a,b}$ are assigned with at most $b$ bits, we have $T_{2,4} = F_j \implies z_j = 4$ and $T_{4,8} = F_j \implies z_j=8$. This yields the following equations:

\begin{itemize}
    \item $4 = \frac{4}{\log_2 (512/17)} \log_2 T_{2,4} + u$
    \item $8 = \frac{4}{\log_2 (512/17)} \log_2 T_{4,8} + u$
    \item $S \cdot \sum_j q_j \le d \cdot \bar{b}$
\end{itemize}

Our goal is thus to adjust $u$ via a binary search to satisfy the bitwidth constraint. That is, we decrease $u$ if the calculated $\sum_j q_j > d \cdot \bar{b}$ in the current round, and vice versa. With $u$ established, we can determine $q_j$. 

\begin{figure}
    \centering
    \begin{minipage}[t]{0.9\linewidth}{
		\vspace{-0.00in}
		\begin{center}
		\includegraphics[width=\textwidth, ]{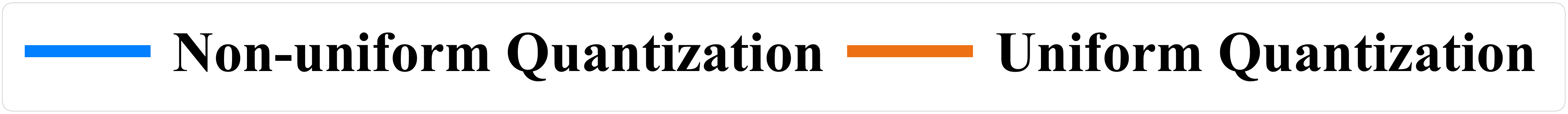}
		\end{center}
		}
    \end{minipage}

    \hspace{-0.3cm}
        \subfigure[LLaMA MMLU 8b]{
		\begin{minipage}[t]{0.33\linewidth}{
		\vspace{-0.00in}
		\begin{center}
		\includegraphics[width=\textwidth, ]{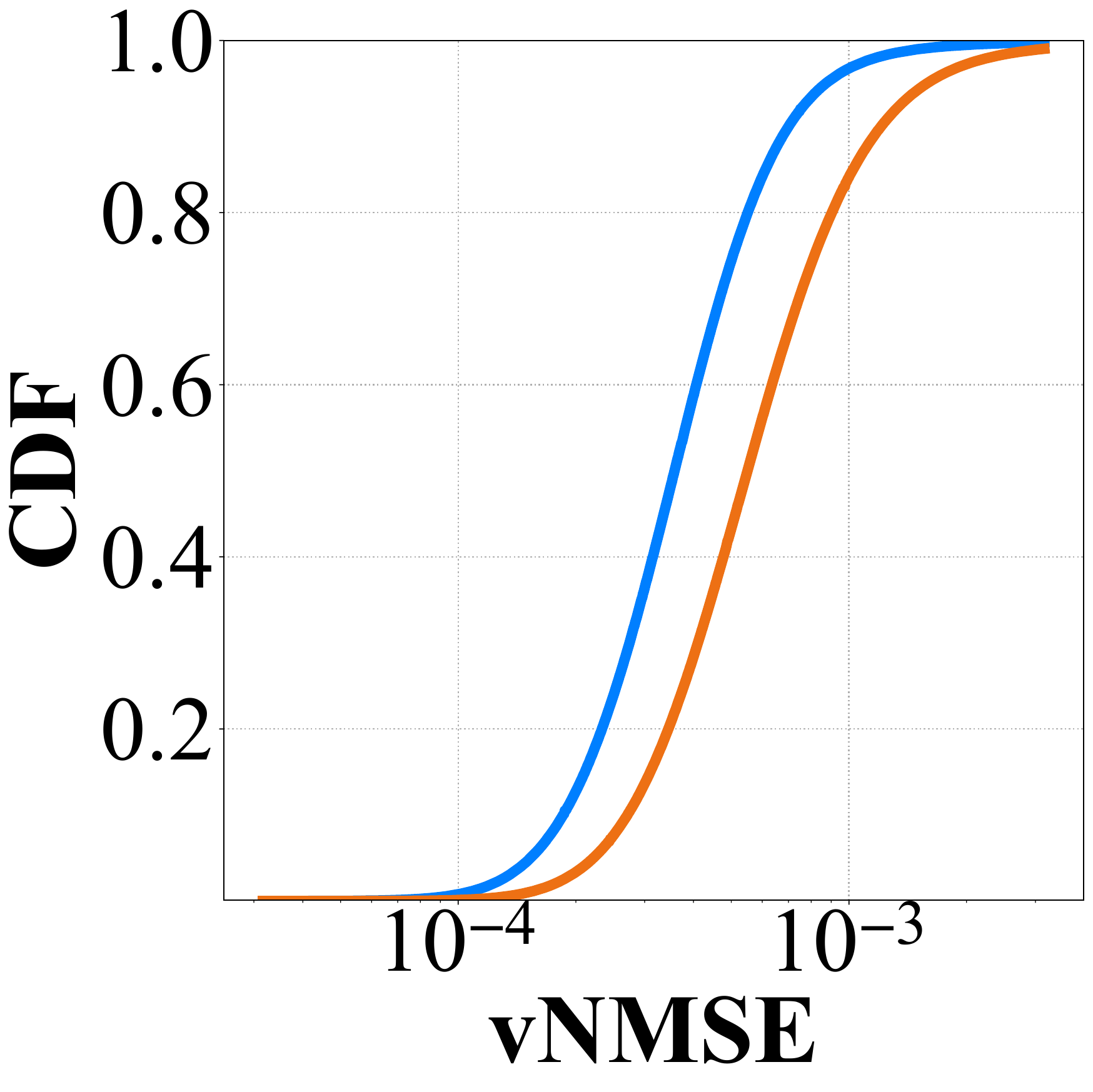} 
		\end{center}
            \vspace{-0.1cm}
		}
		\label{subfig:design-non-linear_8bit}
		\end{minipage}
	}
    \hspace{-0.3cm}
        \subfigure[LLaMA MMLU 4b]{
		\begin{minipage}[t]{0.33\linewidth}{
		\vspace{-0.00in}
		\begin{center}
		\includegraphics[width=\textwidth]{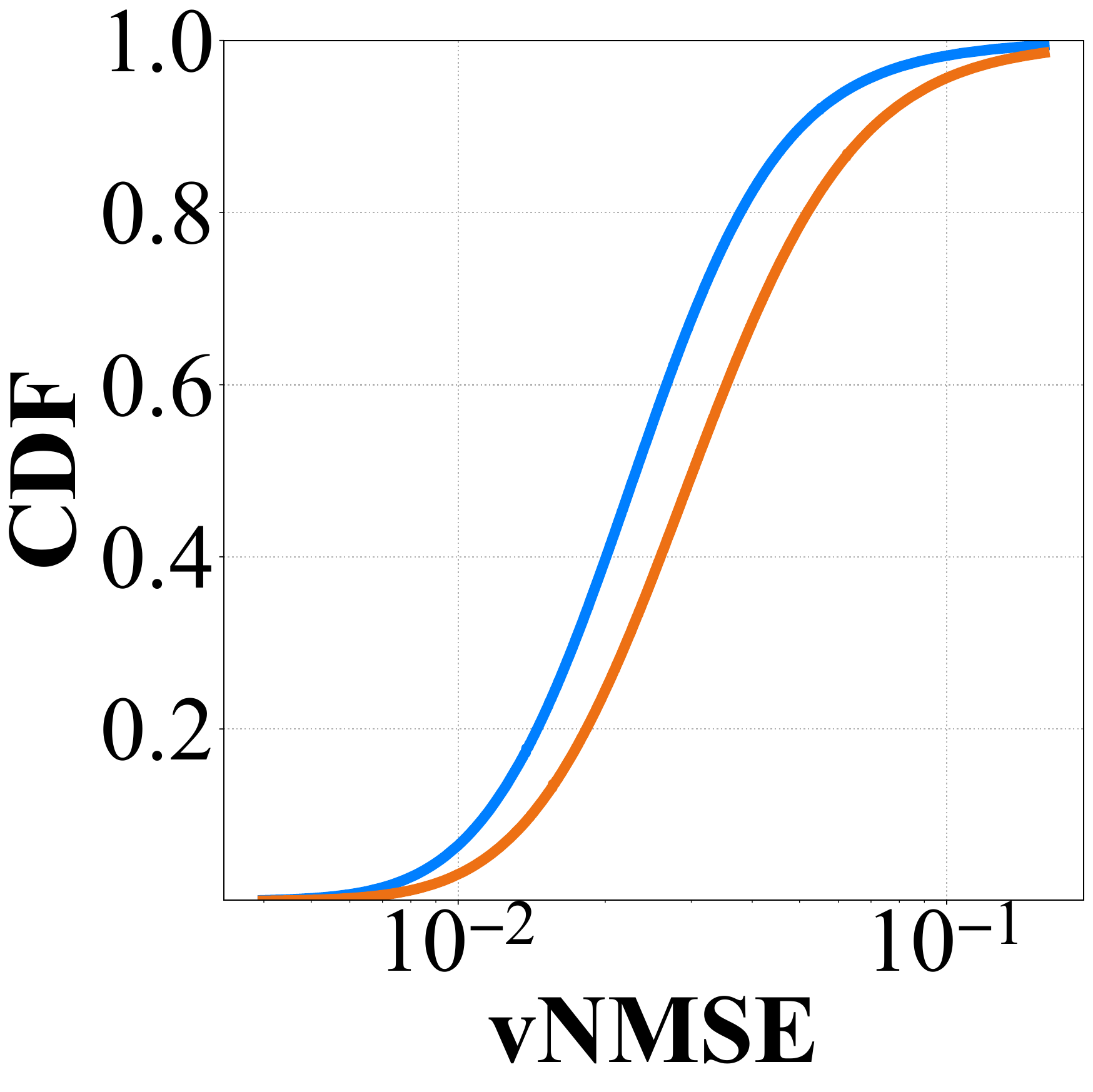}
		\end{center}
            \vspace{-0.1cm}
		}
		\label{subfig:design-non-linear-4bit}
		\end{minipage}
	}
    \hspace{-0.3cm}
        \subfigure[LLaMA MMLU 2b]{
		\begin{minipage}[t]{0.33\linewidth}{
		\vspace{-0.00in}
		\begin{center}
		\includegraphics[width=\textwidth]{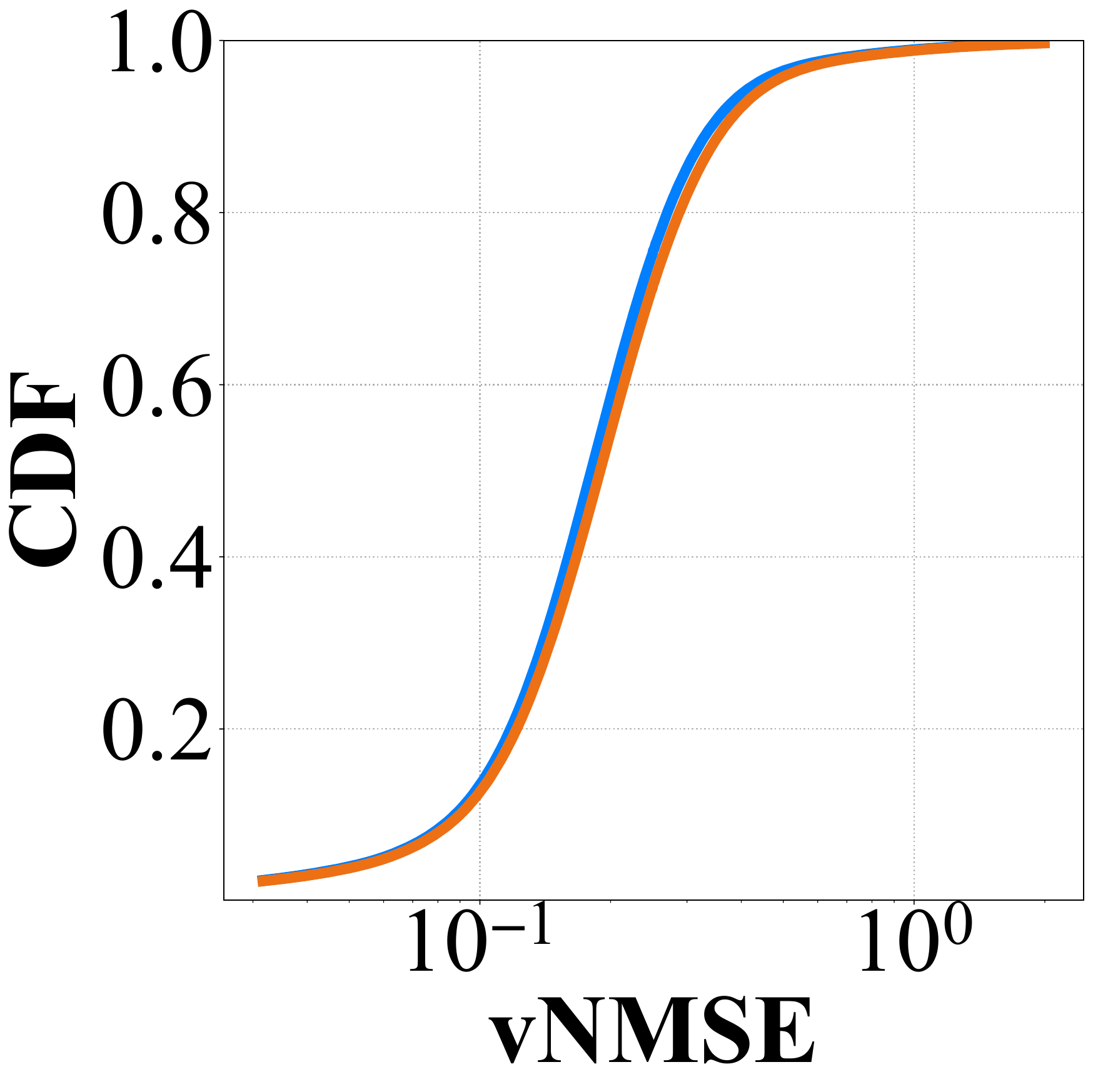}
		\end{center}
            \vspace{-0.1cm}
		}
		\label{subfig:design-non-linear-2bit}
		\end{minipage}
	}

    \hspace{-0.3cm}
        \subfigure[Gemma Chat 8b]{
		\begin{minipage}[t]{0.33\linewidth}{
		\vspace{-0.00in}
		\begin{center}
		\includegraphics[width=\textwidth, ]{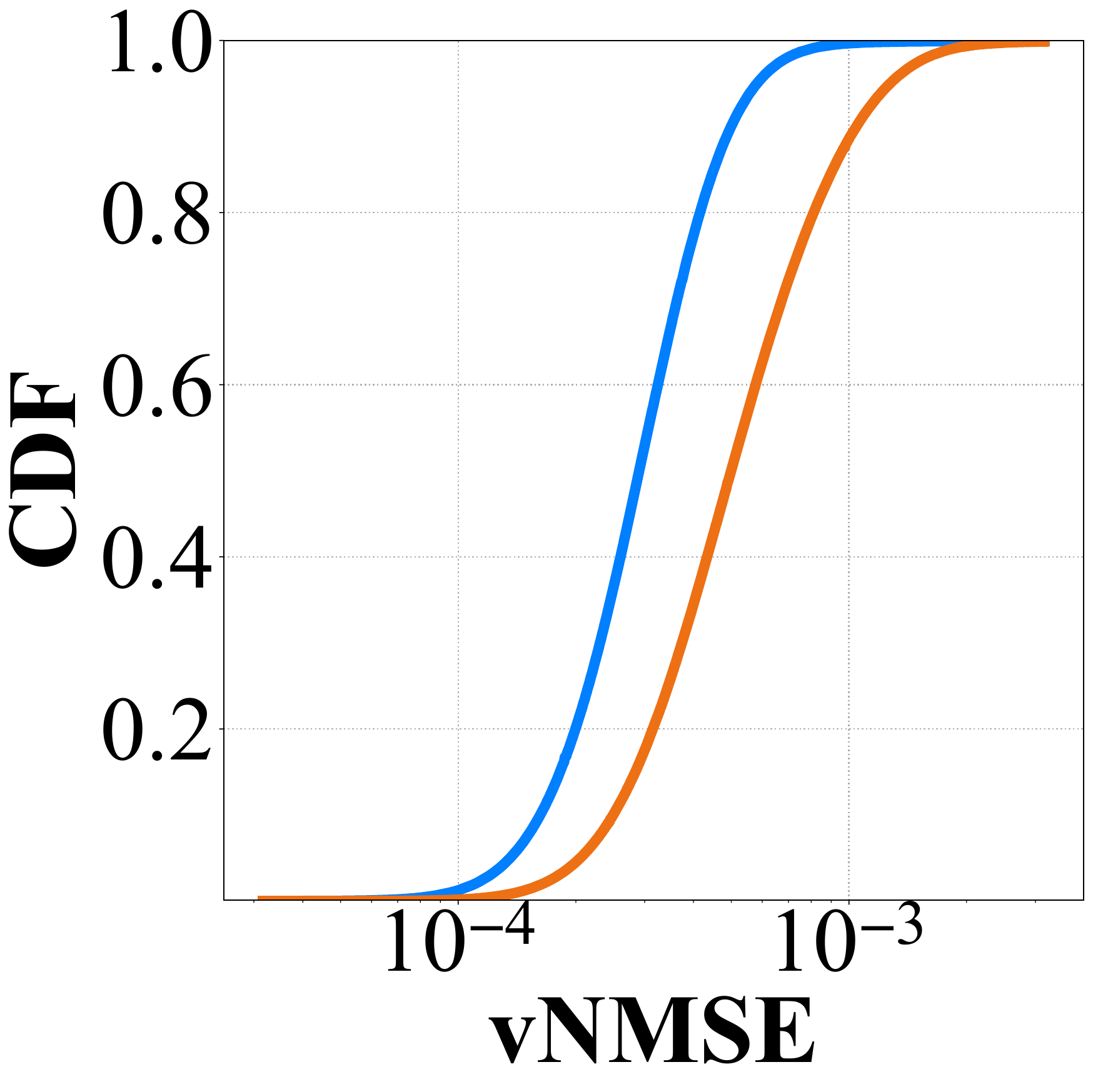} 
		\end{center}
            \vspace{-0.1cm}
		}
		\label{subfig:design-non-linear_8bit_gemma}
		\end{minipage}
	}
    \hspace{-0.3cm}
        \subfigure[Gemma Chat 4b]{
		\begin{minipage}[t]{0.33\linewidth}{
		\vspace{-0.00in}
		\begin{center}
		\includegraphics[width=\textwidth]{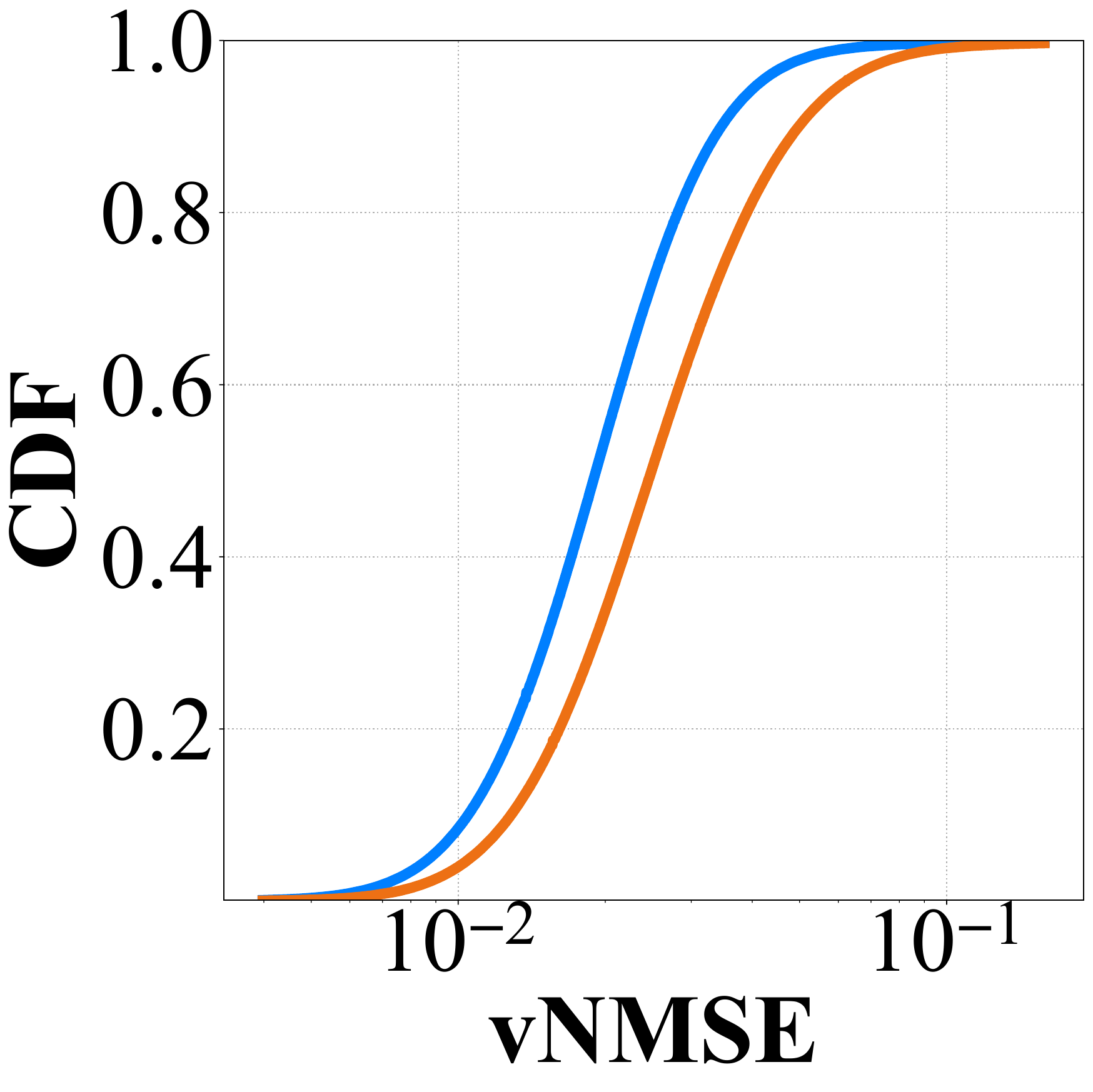}
		\end{center}
            \vspace{-0.1cm}
		}
		\label{subfig:design-non-linear-4bit_gemma}
		\end{minipage}
	}
    \hspace{-0.3cm}
        \subfigure[Gemma Chat 2b]{
		\begin{minipage}[t]{0.33\linewidth}{
		\vspace{-0.00in}
		\begin{center}
		\includegraphics[width=\textwidth]{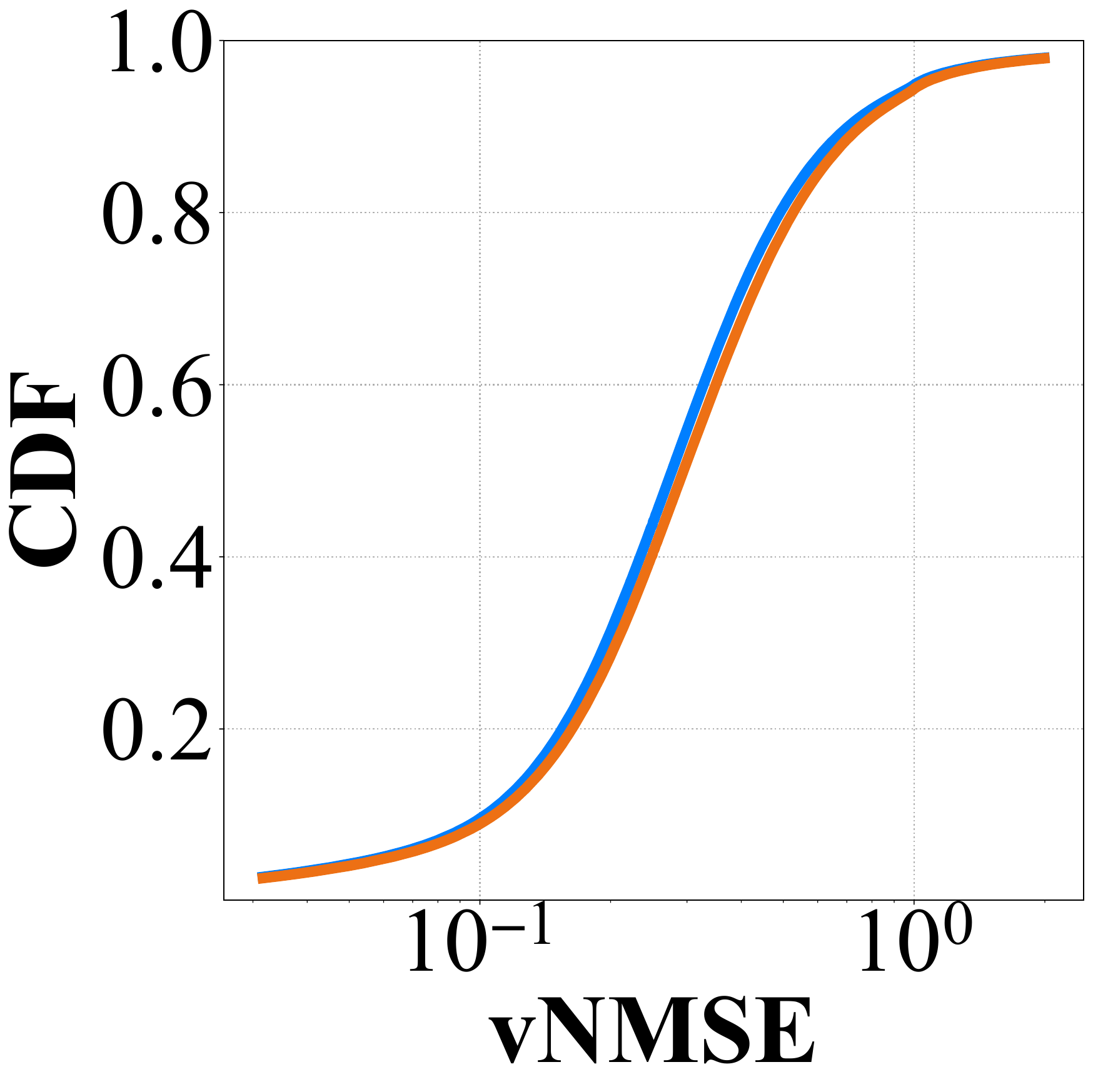}
		\end{center}
            \vspace{-0.1cm}
		}
		\label{subfig:design-non-linear-2bit_gemma}
		\end{minipage}
	}
    \vspace{-0.41cm}
    \caption{Compression error comparisons measured in vNMSE~\cite{vargaftik2021drive} per super-group between non-uniform quantization and uniform quantization. Each super-group uses 2, 4, or 8 bits per coordinate, and we plot the CDFs for each bit width separately.}
    \label{fig:comparison-benchmark-non-uniform}
\end{figure}

\section{Analysis with Different All-reduce Topologies}\label{sec:adaptation-butterfly}

\begin{figure}
    \centering
    \includegraphics[width=0.85\linewidth]{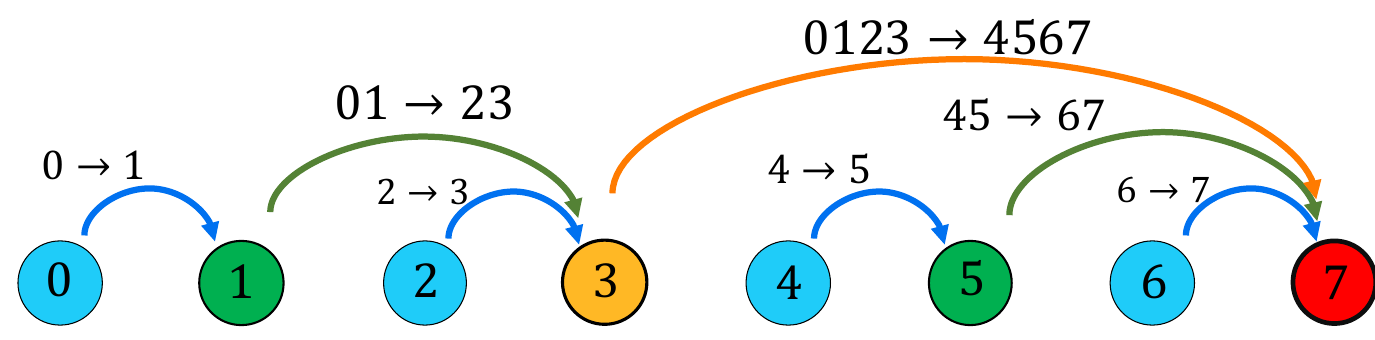}
    \caption{Butterfly all-reduce topology for a specific gradient chunk. Each color represents concurrent transmissions.}
    \label{fig:butterfly-demo}
\end{figure}

\begin{figure*}[]
    \centering
    \begin{minipage}[t]{0.7\linewidth}{
		\vspace{-0.00in}
		\begin{center}
		\includegraphics[width=\textwidth, ]{new_legends/bert_perp_legend_testing_tta_normal.pdf}
		\end{center}
		}
        \end{minipage}

	\hspace{-0.5cm}
        \subfigure[BERT-large MaskedLM]{
		\begin{minipage}[t]{0.25\linewidth}{
		\vspace{-0.00in}
		\begin{center}
		\includegraphics[width=\textwidth, ]{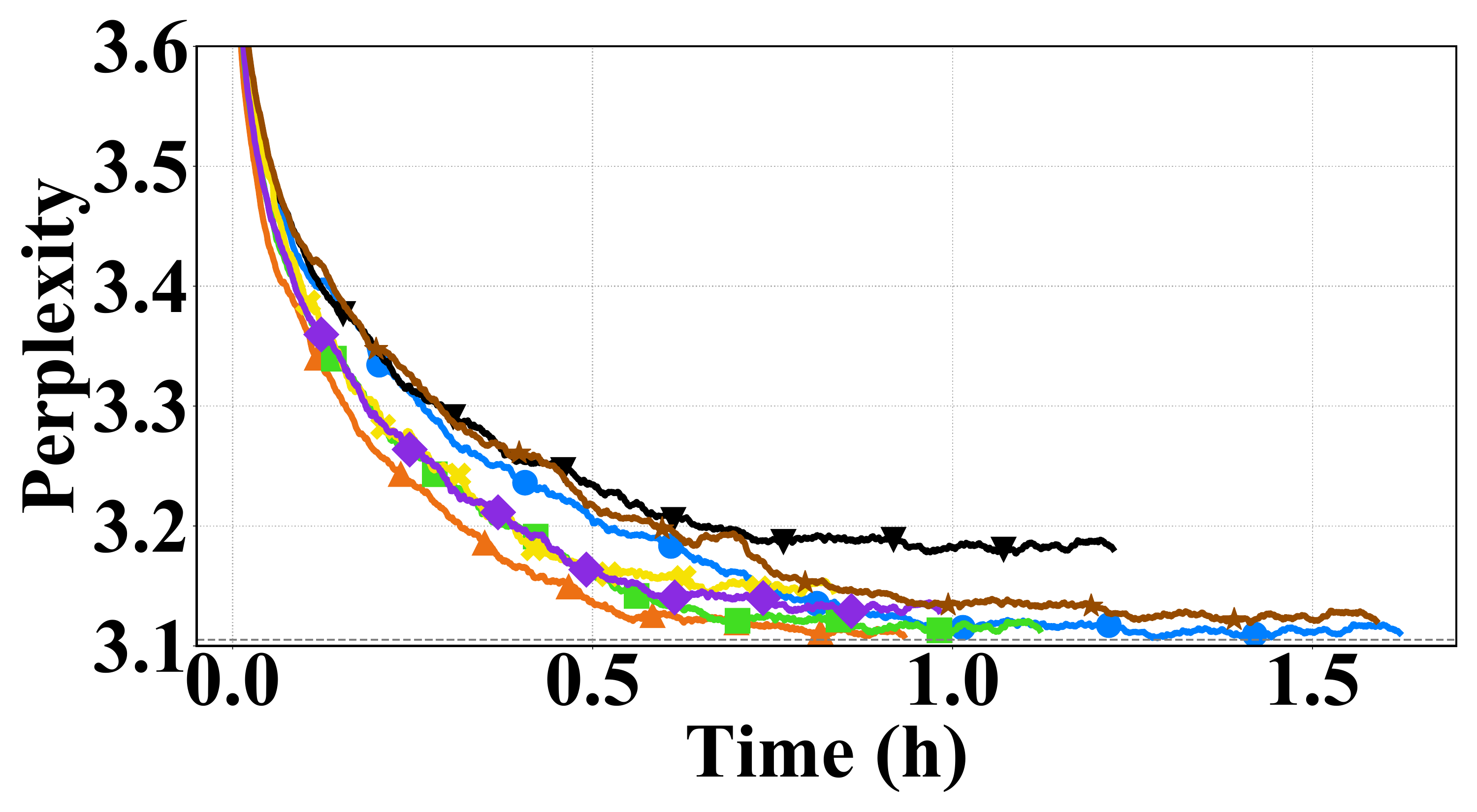}
		\end{center}
		}
		\label{subfig:e2e-bert-large-full}
		\end{minipage}
	}
        \hspace{-0.3cm}
        \subfigure[Gemma 1B Chat]{
		\begin{minipage}[t]{0.25\linewidth}{
		\vspace{-0.00in}
		\begin{center}
		\includegraphics[width=\textwidth, ]{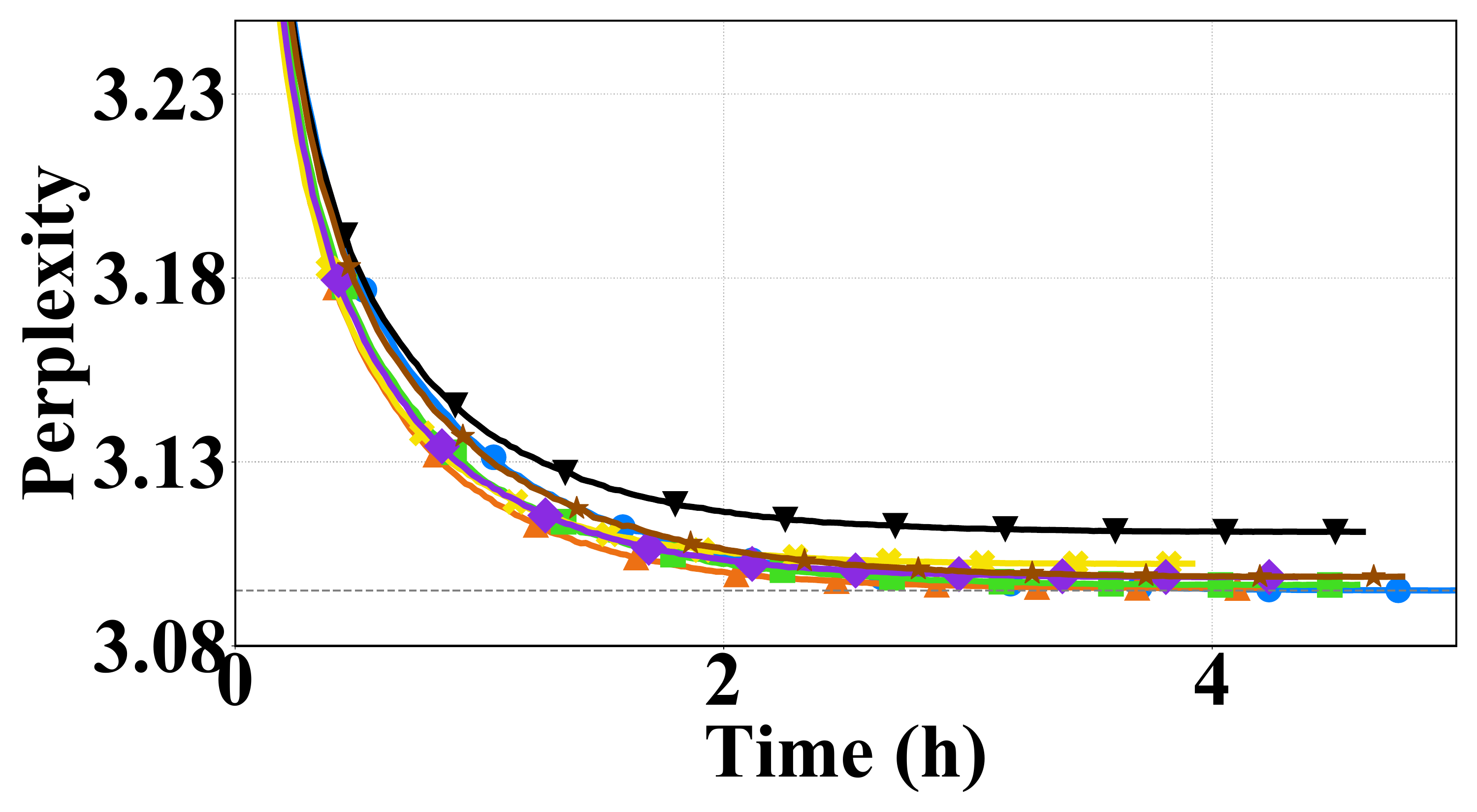}
		\end{center}
		}
		\label{subfig: e2e-gemma-full}
		\end{minipage}
	    }
    \hspace{-0.3cm}
	\subfigure[LLaMA 1B Chat]{
		\begin{minipage}[t]{0.25\linewidth}{
		\vspace{-0.00in}
		\begin{center}
		\includegraphics[width=\textwidth, ]{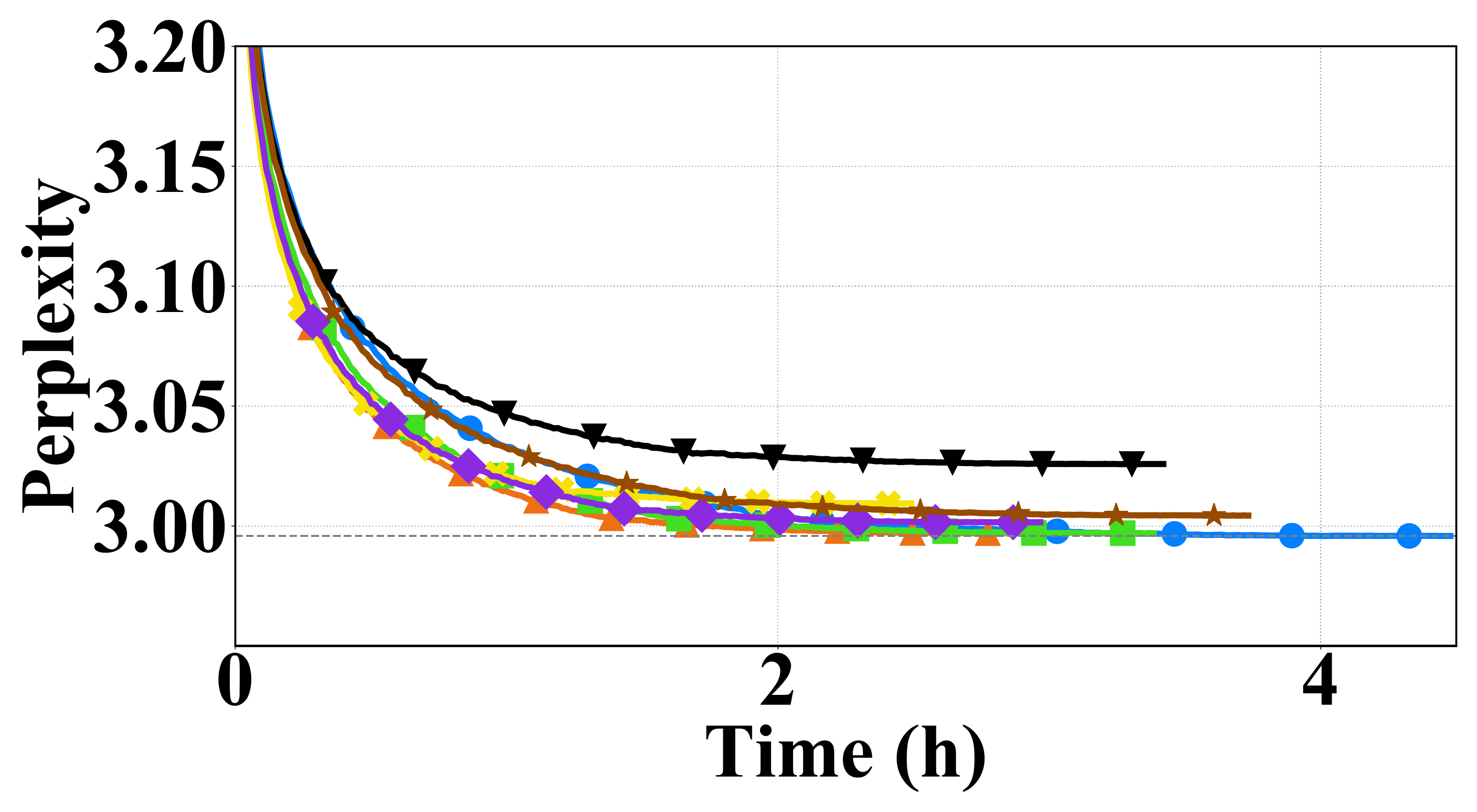}
		\end{center}
		}
		\label{subfig: e2e-LLaMA-full}
		\end{minipage}
	}
        \hspace{-0.3cm}
        \subfigure[LLaMA 1B MMLU]{
            \begin{minipage}[t]{0.25\linewidth}{
            \vspace{-0.00in}
            \begin{center}
            \includegraphics[width=\textwidth, ]{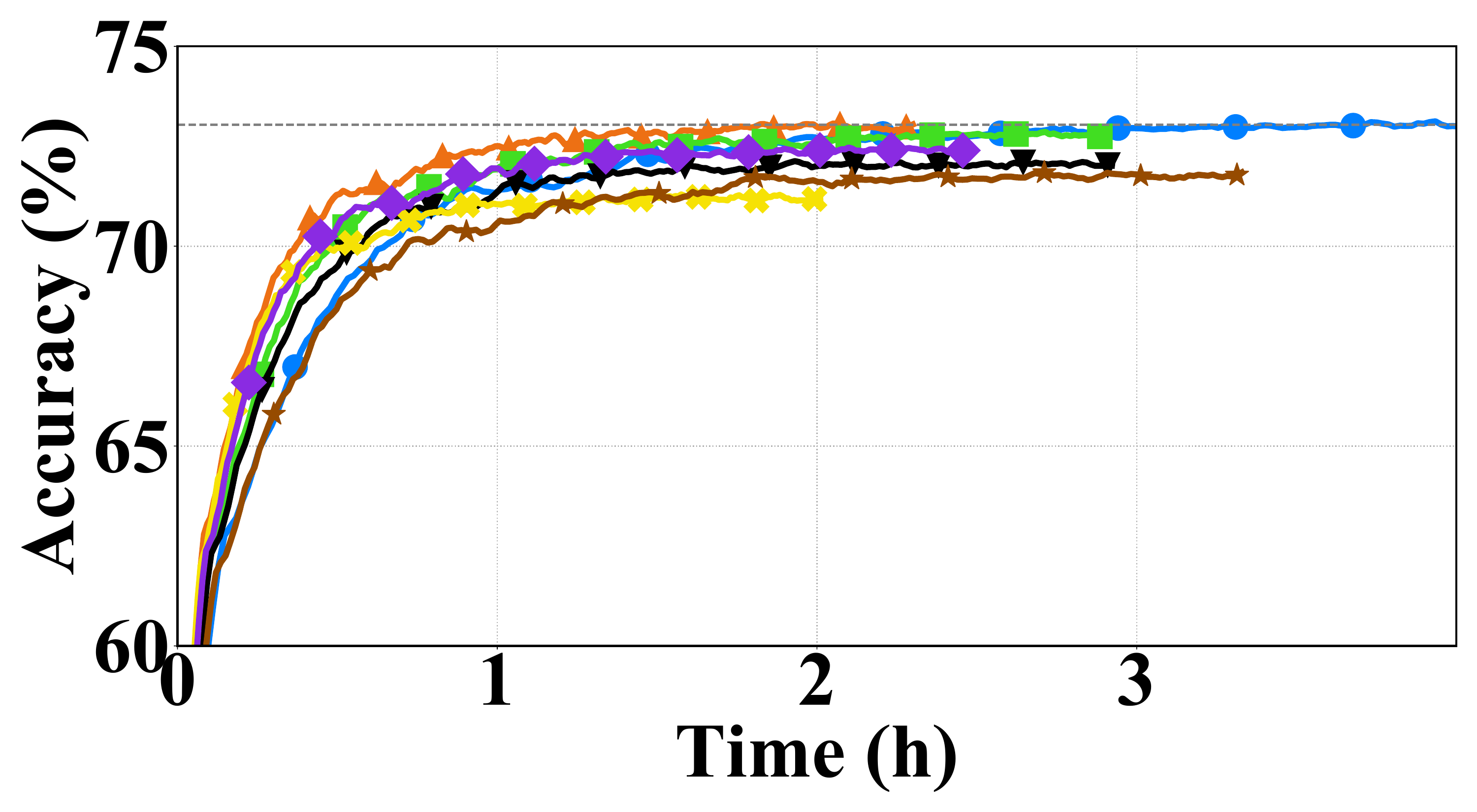}
            \end{center}
            }
            \label{subfig: e2e-mmlu-full}
            \end{minipage}
        }
        \vspace{-0.45cm}
	\caption{Zoomed-out TTA curves for LLM training
    and fine-tuning on an 8-GPU/4-worker testbed using ring
all-reduce.}
        \vspace{-0.35cm}
	\label{fig: e2e-tta-full}
\end{figure*}

Unlike many prior compression schemes~\cite{li2024thc, fei2021efficient, wang2023cupcake, terngrad} that are designed for parameter-server-based aggregation~\cite{li2014scaling}, \sysname naturally supports different multi-hop all-reduce topologies. That includes two well-known all-reduce topologies, namely ring all-reduce~\cite{ring} and butterfly all-reduce~\cite{thakur2005optimization} (also known as the recursive doubling~\cite{thakur2005optimization}, as illustrated in Figure~\ref{fig:butterfly-demo}). 
We remark that compared with ring all-reduce, butterfly all-reduce typically achieves lower tail latency in large-scale DDP training systems~\cite{desensi2024swing}.

We further observe that deploying \sysname to butterfly all-reduce also improves \textit{scalability} with respect to its compression error as $n$ grows larger. The intuition is that the compression error at each hop is proportional to the values of the partial sum being transmitted, which is in turn proportional to the size of the corresponding subtree if gradients on different workers follow the same distribution. Figure~\ref{fig:butterfly-demo} illustrates this claim: worker $3$, with a subtree size of $4$, compresses the partial sum of worker $0 \sim 3$'s gradients and transmits it to worker $7$, which holds the partial sum of workers $4 \sim 7$. 

We now heuristically analyze the compression error with ring all-reduce and butterfly all-reduce. For this analysis, we use the sum of the expected mean squared error (MSE) at each worker. We assume that the gradient data $X_{i,j}[k]$ indexed at $k$ of the $j$'th super-group at worker $i$ is bounded by $M = \max_{i,k} |X_{i,j}[k]|$. It can be derived that the MSE for compressing the partial sum gradient $s_{i, j}$ at worker $i$ is bounded by $\text{MSE} \le \epsilon S\max_k|s_{i, j}[k]|^2$. We note that $\max |s_{i, j}[k]| \le M|\text{subtree(i)}|$ where |subtree$(i)|$ is the subtree size rooted at worker $i$. Thus, with ring all-reduce, the expected worst-case MSE can be bounded by
$$\text{MSE}\le \sum_i \epsilon Si^2M^2 = O(\epsilon SM^2n^3),$$
while that of butterfly all-reduce is $$\text{MSE} \le \sum_{l \le \log n} \epsilon S(M2^l)^2 \cdot (n / 2^l) = O(\epsilon S M^2n^2).$$
That is, our upper bound on the MSE for the butterfly is a factor of $n$ less than that for the ring.

\begin{figure}[]

    \centering
    \begin{minipage}[t]{0.9\linewidth}{
		\vspace{-0.00in}
		\begin{center}
		\includegraphics[width=\textwidth, ]{exp_figures/bert_perp_legend_testing_tta_normal.pdf}
		\end{center}
		}
        \end{minipage}

        \hspace{-0.3cm}
        \subfigure[Gemma 1B Chat]{
		\begin{minipage}[t]{0.49\linewidth}{
		\vspace{-0.00in}
		\begin{center}
		\includegraphics[width=\textwidth, ]{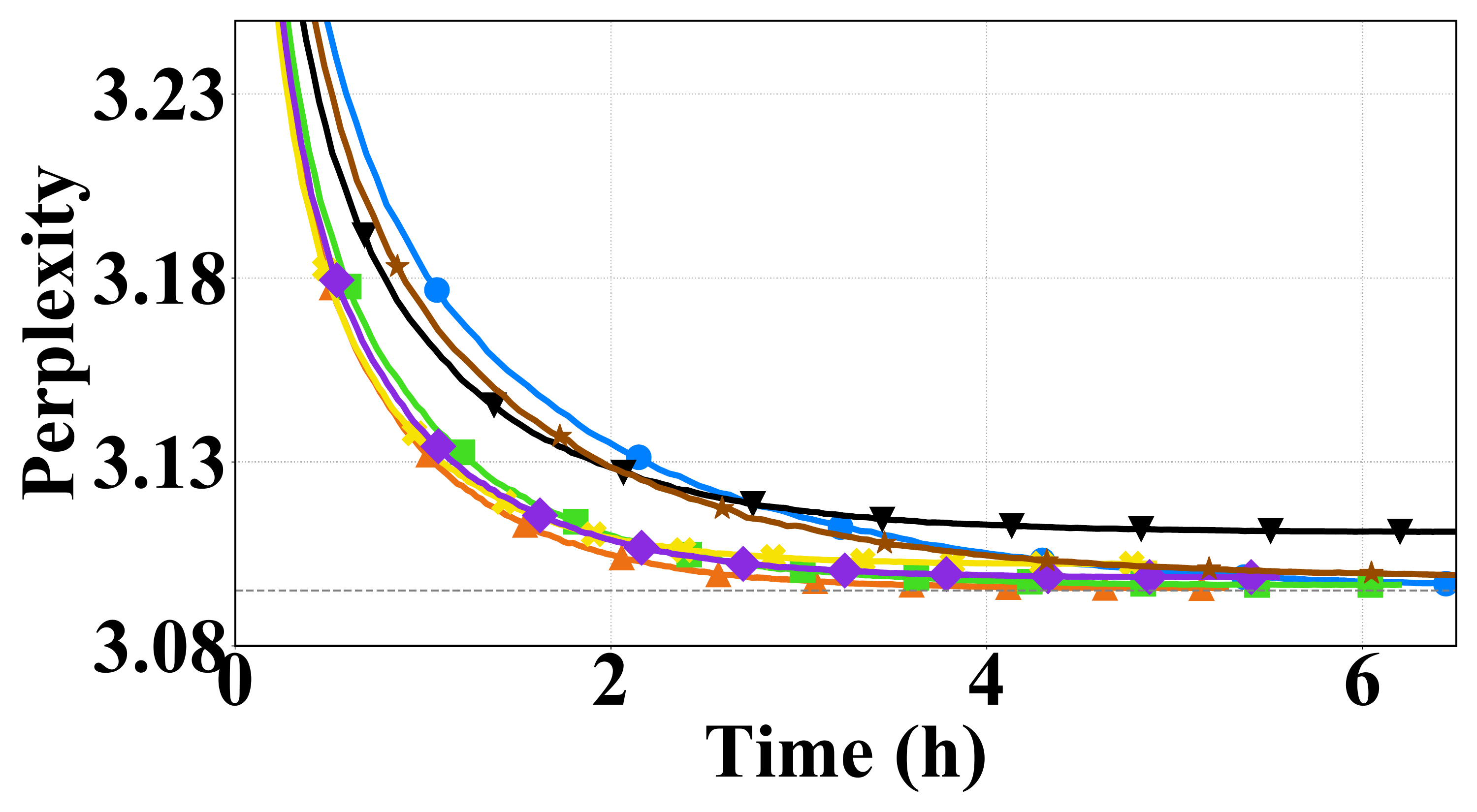}
		\end{center}
		}
		\label{subfig: e2e-gemma-flowcollision-full}
		\end{minipage}
	    }
        \hspace{-0.3cm}
        \subfigure[LLaMA 1B MMLU]{
            \begin{minipage}[t]{0.49\linewidth}{
            \vspace{-0.00in}
            \begin{center}
            \includegraphics[width=\textwidth, ]{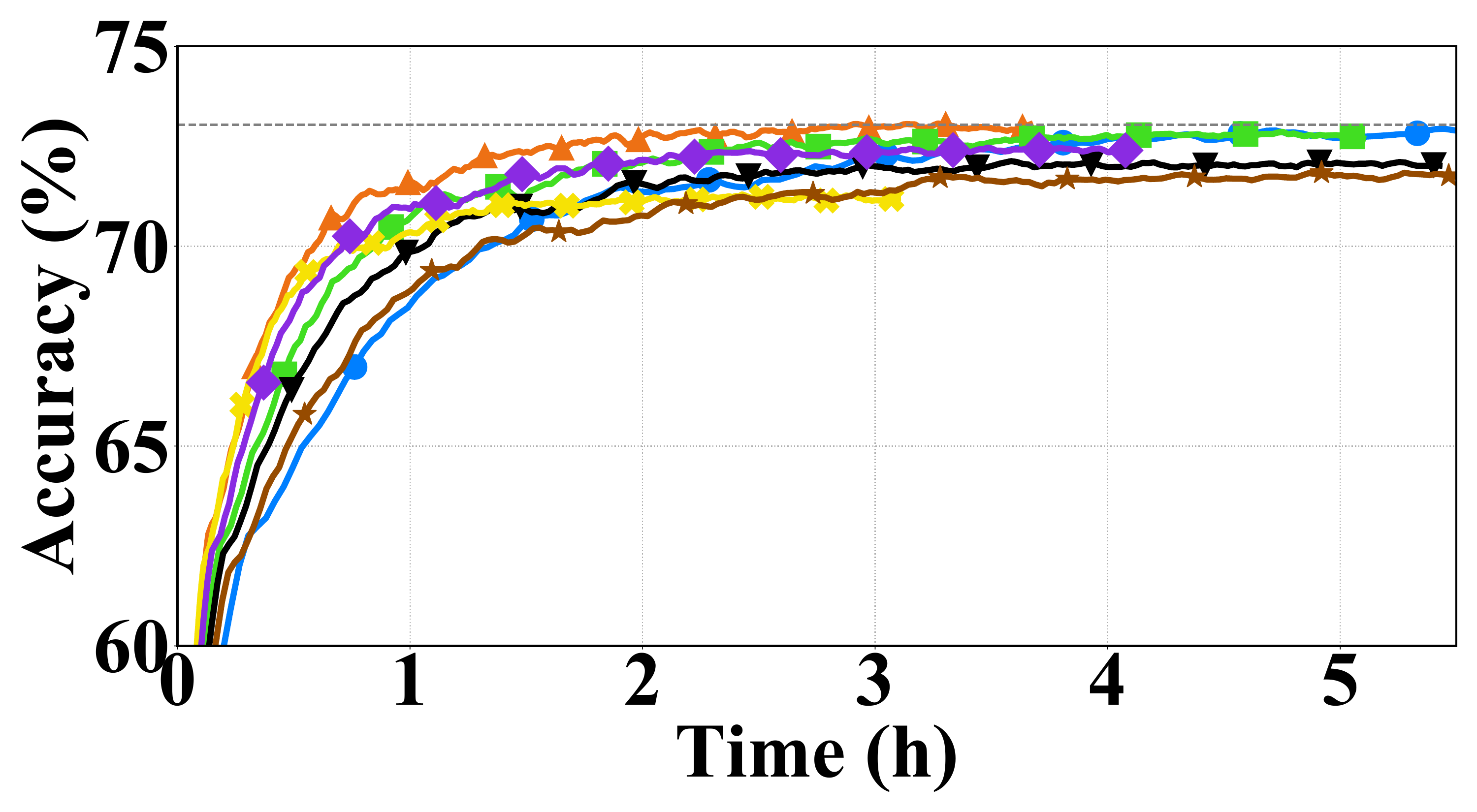}
            \end{center}
            }
            \label{subfig: e2e-mmlu-flowcollision-full}
            \end{minipage}
        }
	\caption{Zoomed-out TTA curves over a shared network.}
        \vspace{-0.25cm}
	\label{fig: e2e-tta-flowcollision-full}
\end{figure}

\begin{figure}[]
    \centering
    \begin{minipage}[t]{0.7\linewidth}{
		\vspace{-0.00in}
		\begin{center}
		\includegraphics[width=\textwidth, ]{exp_figures/legend_mmlu_butterfly_legend_zoomed_in.pdf}
		\end{center}
		}
        \end{minipage}

    \centering
    \hspace{-0.1in}
		\begin{minipage}[t]{0.7\linewidth}{
		\vspace{-0.00in}
		\begin{center}
		\includegraphics[width=\textwidth, ]{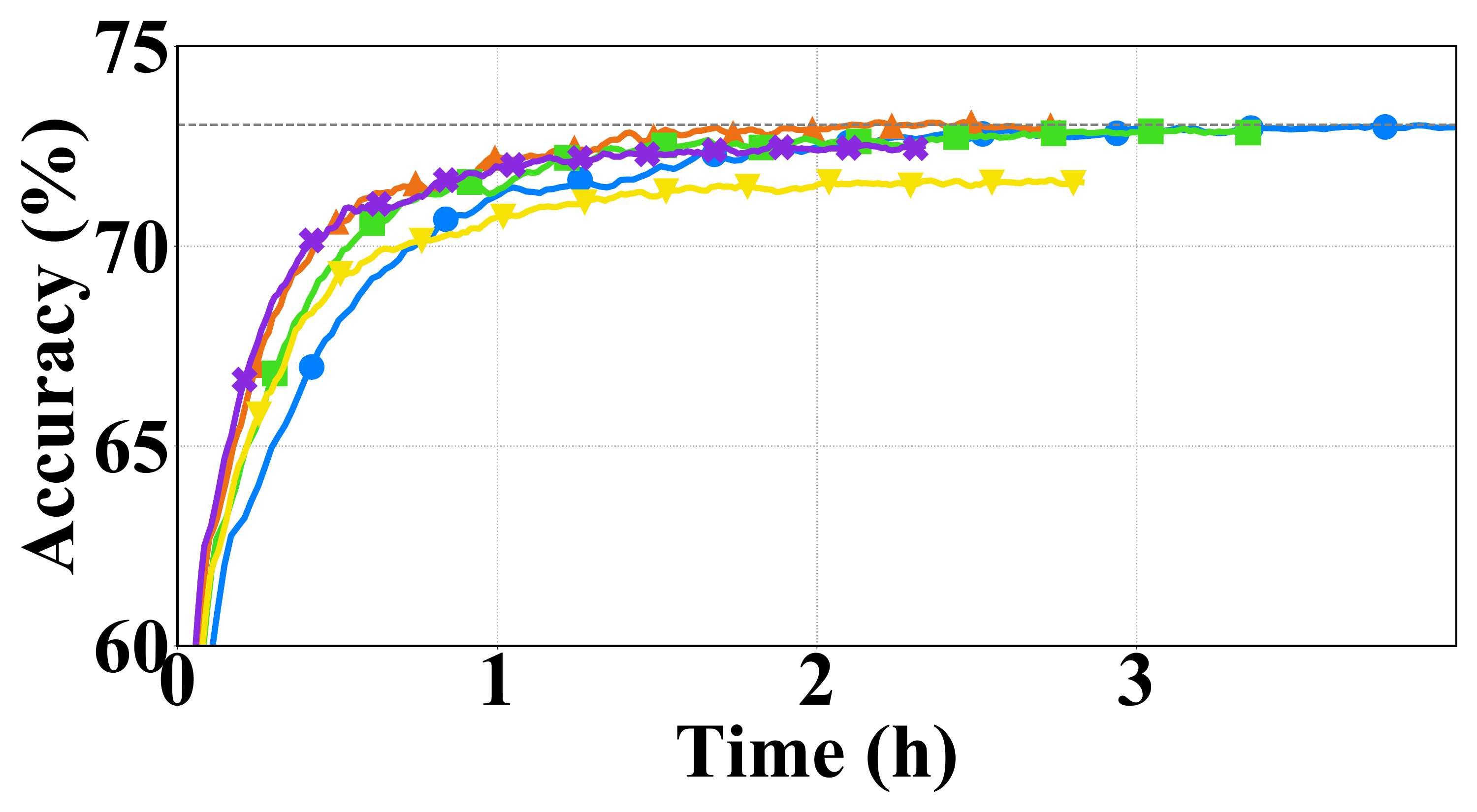}
		\end{center}
		}
		\label{subfig: lm_accuracy_butterfly_tta_full}
		\end{minipage}
\vspace{-2mm}
    \caption{Zoomed-out TTA (LLaMA 1B MMLU) for butterfly all-reduce.}
    \label{fig: overall-tta-butterfly}
    \vspace{-2mm}
\end{figure}

\begin{figure}
    \centering
    \includegraphics[width=0.7\linewidth]{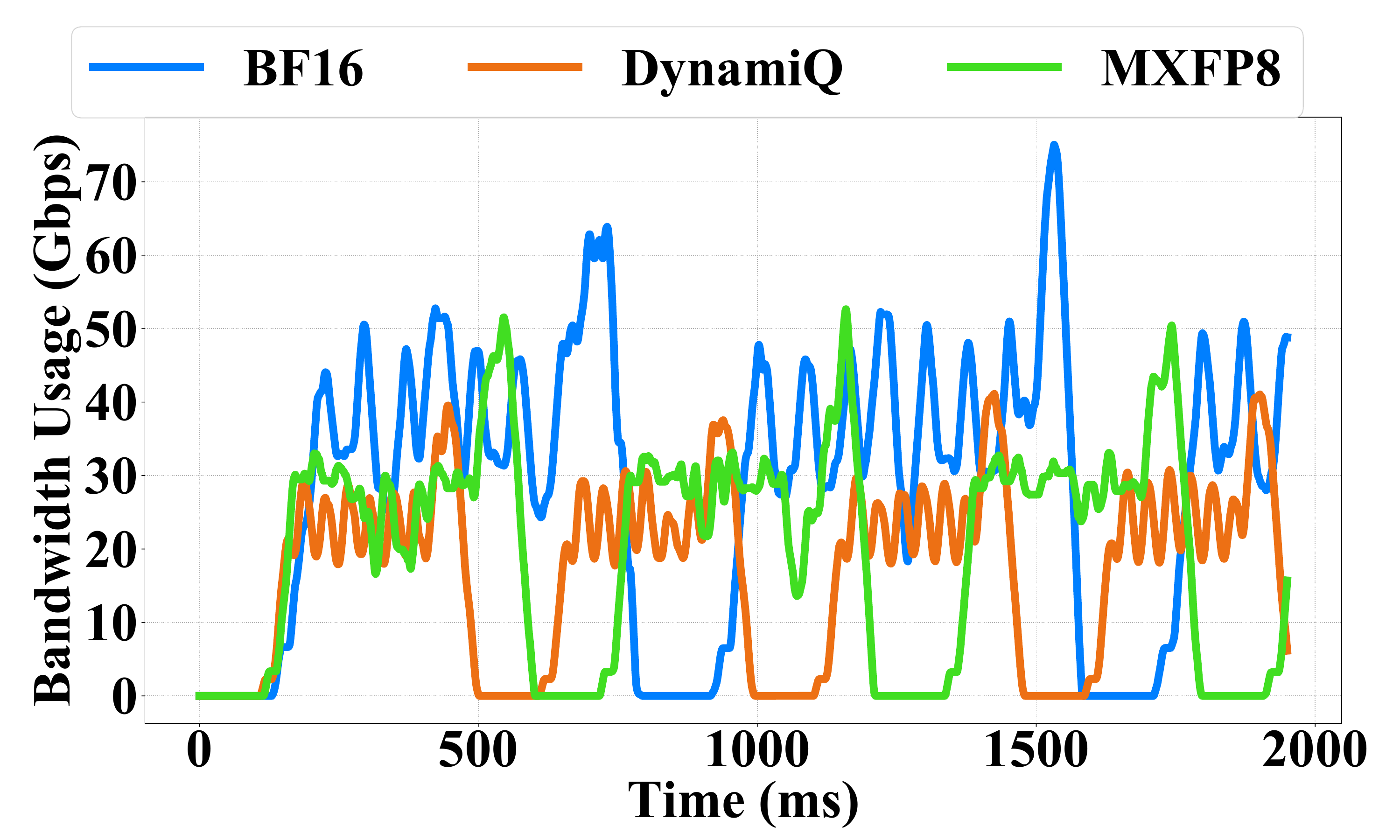}
    \vspace{-2mm}
    \caption{Bandwidth usage over time for the LLaMA 1B MMLU workload with ring all-reduce.}

    \label{fig:time-to-bandwidth}
\end{figure}

\begin{figure*}[]
    \centering
    \begin{minipage}[t]{0.61\linewidth}{
		\vspace{-0.00in}
		\begin{center}
		\includegraphics[width=\textwidth, ]{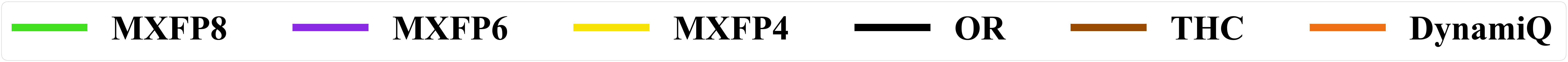}
		\end{center}
            
		}
        \end{minipage}
    
        \vspace{-2mm}

	\hspace{-0.3cm}
	    \subfigure[BERT-large MaskedLM]{
            \begin{minipage}[t]{0.24\linewidth}{
            \begin{center}
            \includegraphics[width=\textwidth, ]{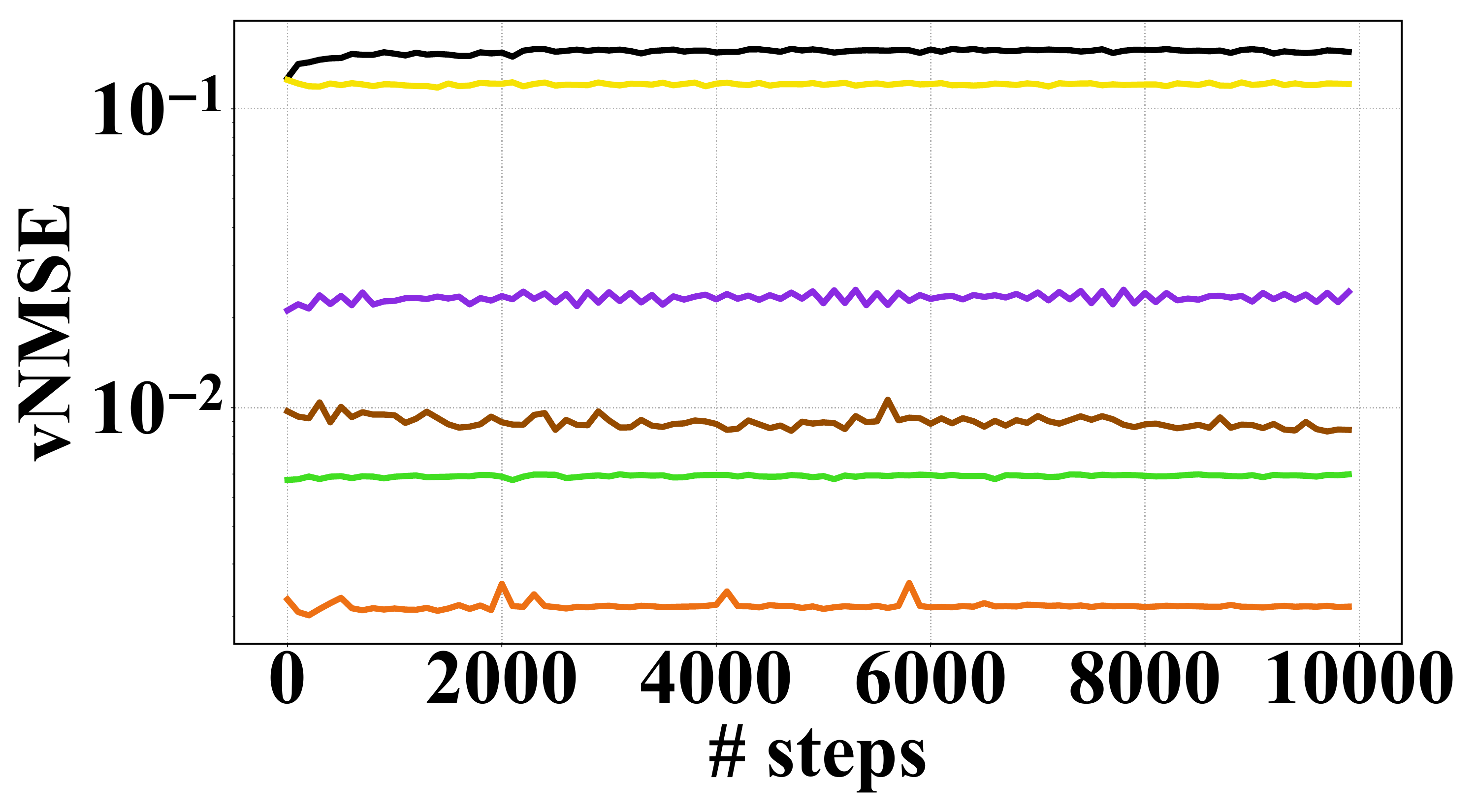}
            \end{center}
            }
            \vspace{-3mm}
            \label{subfig: vnmse-bert}
            \end{minipage}
        }
	%
        %
        \hspace{-0.3cm}
        \subfigure[Gemma 1B Chat]{
		\begin{minipage}[t]{0.24\linewidth}{
		\begin{center}
		\includegraphics[width=\textwidth, ]{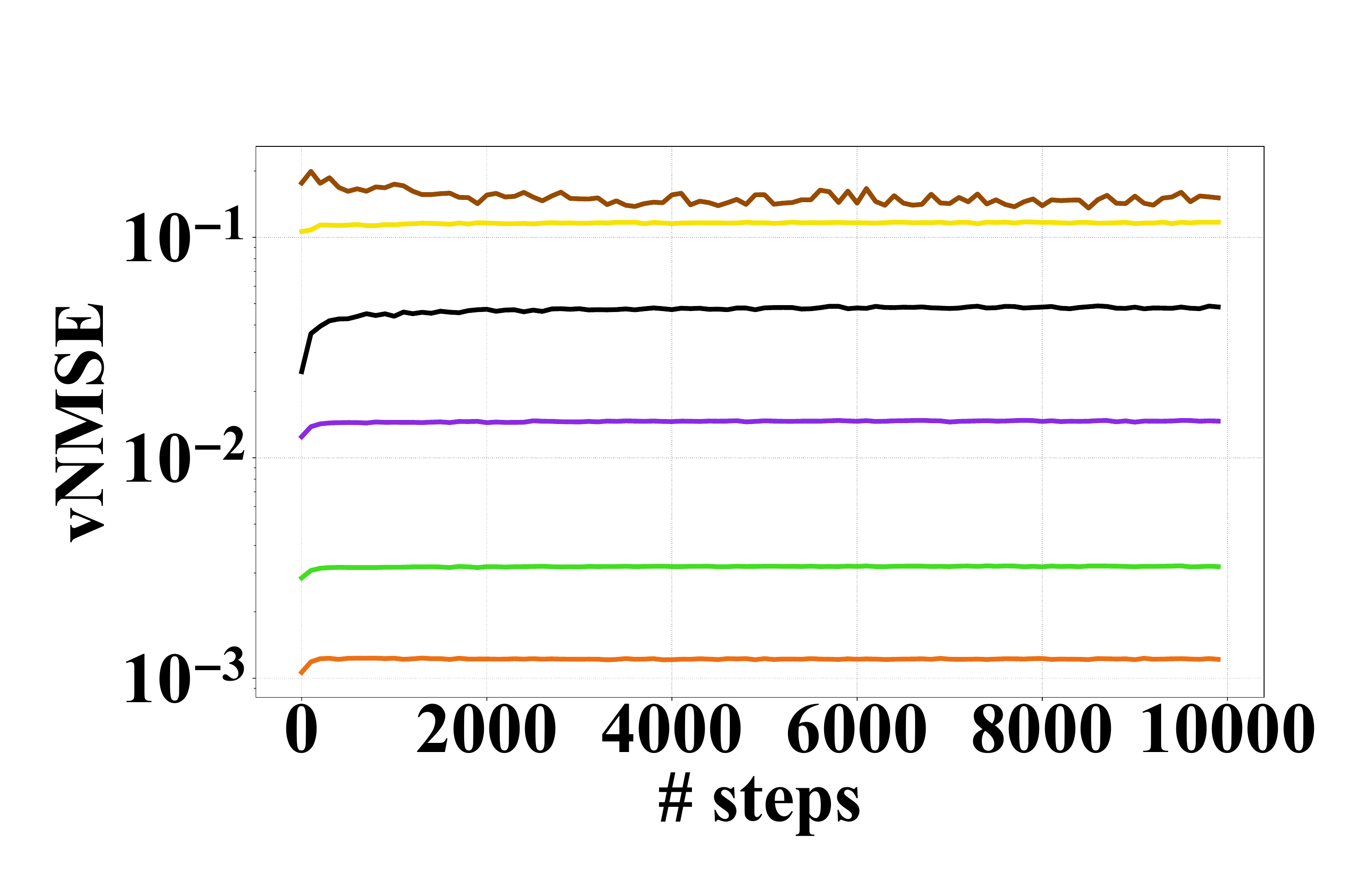}
		\end{center}
		}
            \vspace{-3mm}
		\label{subfig: vnmse-gemma}
		\end{minipage}
	}
        \hspace{-0.3cm}
	    \subfigure[LLaMA 1B Chat]{
            \begin{minipage}[t]{0.24\linewidth}{
            \begin{center}
            \includegraphics[width=\textwidth, ]{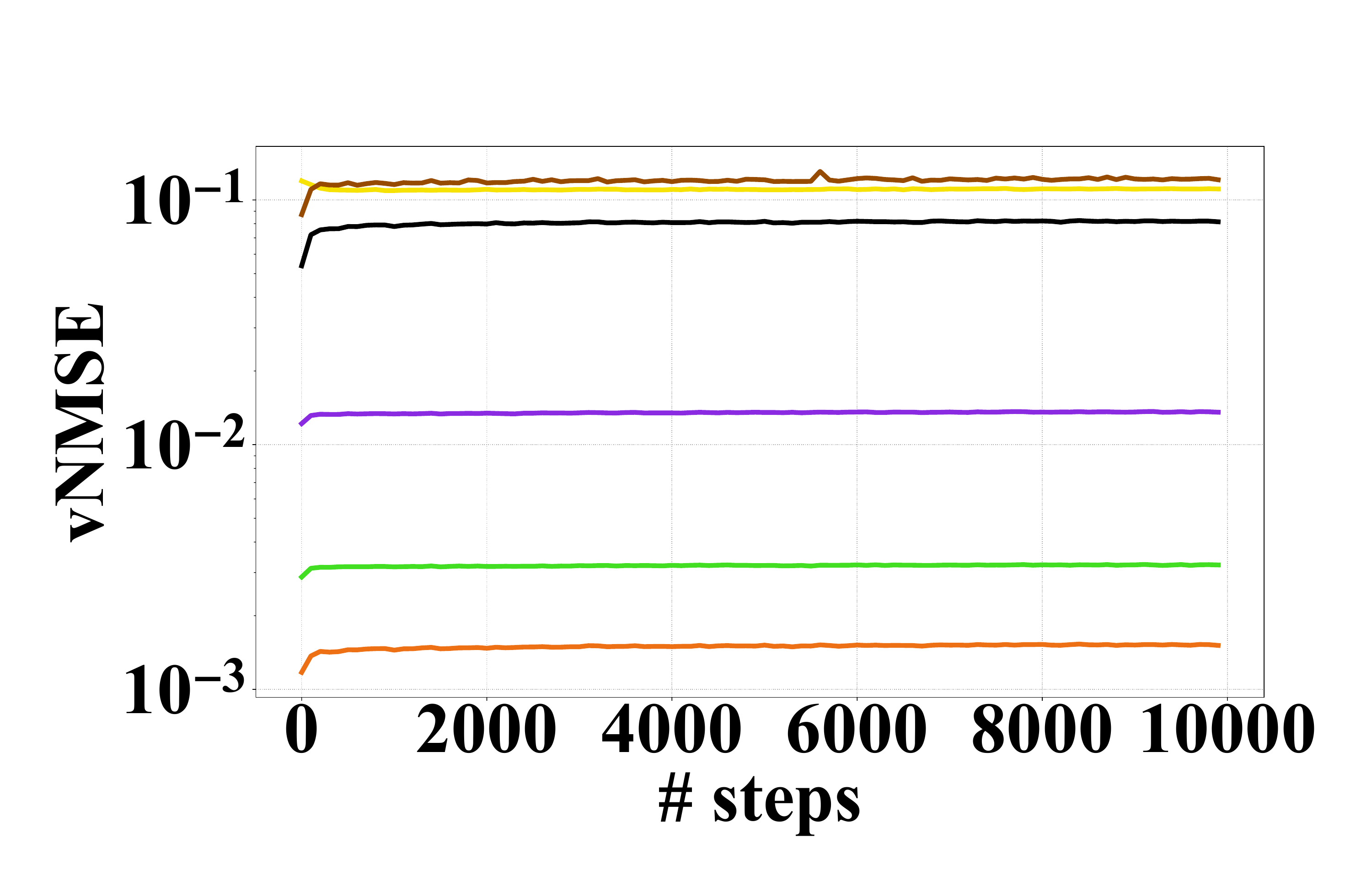}
            \end{center}
            }
            \vspace{-3mm}
                
            \label{subfig: vnmse-LLaMA}
            \end{minipage}
        }
    \hspace{-0.3cm}
        \subfigure[LLaMA 1B MMLU]{
		\begin{minipage}[t]{0.24\linewidth}{
		\begin{center}
		\includegraphics[width=\textwidth, ]{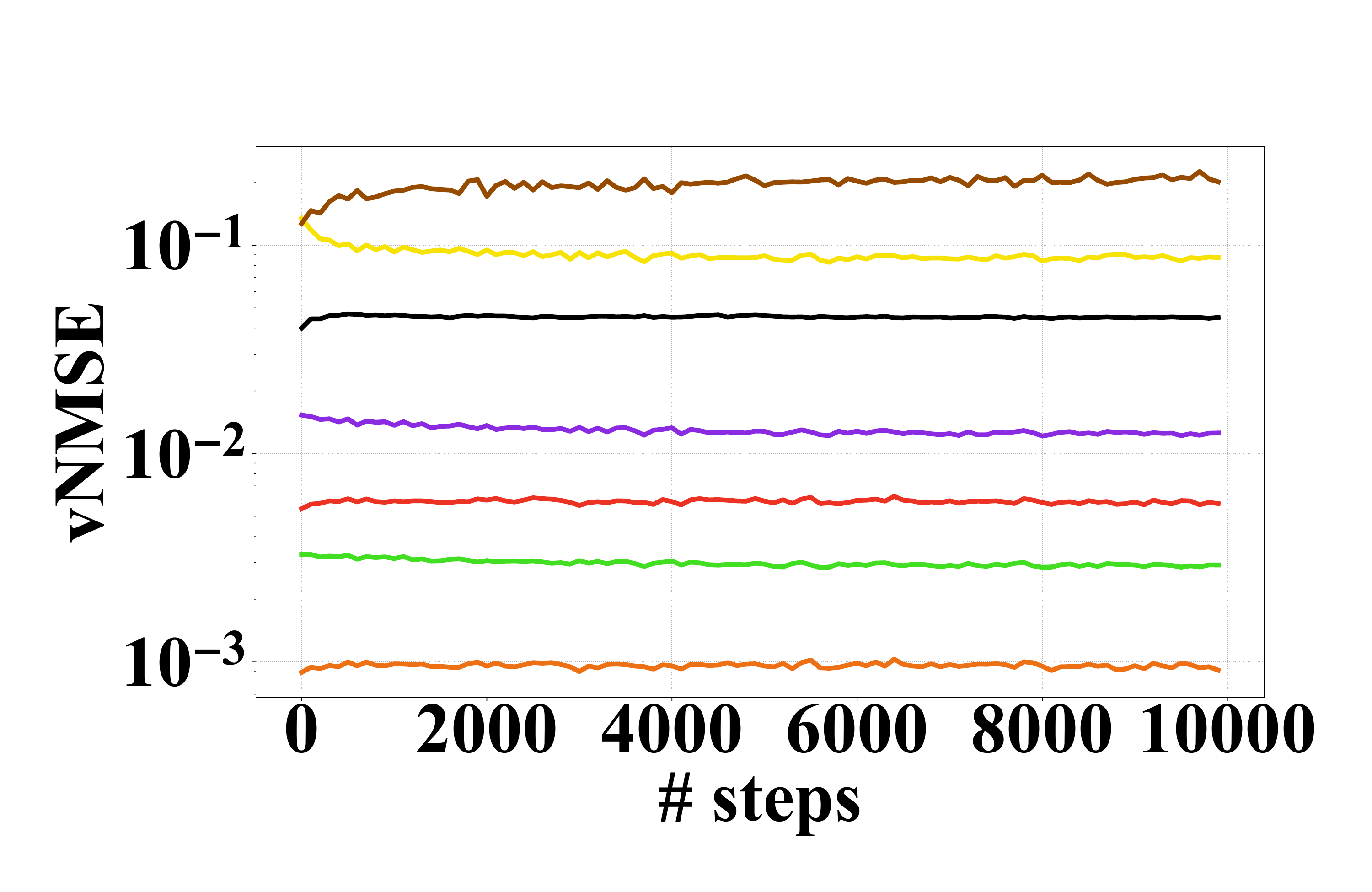}
		\end{center}
            \vspace{-3mm}
		}
            \vspace{-3mm}
		\label{subfig: vnmse-mmlu}
		\end{minipage}
	}
    
        \vspace{-0.3cm}

	\caption{vNMSE comparison over different workloads on our 8-GPU/4-worker testbed with ring all-reduce.}
	\label{fig: vnmse-e2e}
\end{figure*}


\section{Additional Experimental Setup}\label{appendix:exp_setup}

\subp{Adaptation of Omnireduce to ring all-reduce.} As noted, Omnireduce (OR) was originally tailored for parameter-server architectures. Specifically, OR employs chunked Top-$k$ compression, where each worker selects and aggregates its local top-$k$ gradient chunks. In a single-hop parameter-server architecture, this is easily achieved by having workers send their local top-$k$ chunks directly to the server. However, in multi-hop all-reduce, the local top-$k$ chunk indices can differ across workers; consequently, the number of aggregated chunks in an intermediate hop can exceed $k$. This leads to increased communication overhead, as more than $k$ chunks may need to be transmitted per hop. To address this, we propose an adaptation that computes the union of indices appearing in at least one worker’s local top-$k$ selection. We refer to these as the global top-$K$ chunks, where $K/n_{\text{chunks}} = b/16$ (i.e., both equal to the desired compression ratio). Given a fixed $K$, it is challenging to directly determine the required local $k$ because it varies dynamically with the gradient distribution. We therefore propose a heuristic to approximate $k$. In each round $t$, given $k_t$, we compute the actual number of global chunks, $K'_t$, obtained from the union of workers' local top-$k_t$ chunks. We then use the ratio $K/K'_t$ to adjust $k_{t+1}$ so that $K'_t$ matches the target $K$. We update $k_{t+1}$ according to the following momentum-based rule, where $0 \le \gamma \le 1$ represents the momentum (set to $\gamma=0.8$ in our experiments):
\begin{equation}
k_{t+1} = \gamma k_t + (1-\gamma)\cdot(K/K'_t)\cdot k_t
\end{equation}

\subp{Adaptation of microscaling floating-point compression (MXFPX) to all-reduce.} As the specification of microscaling floating-point formats (\ie, MXFP8, MXFP6, MXFP4)~\cite{opencompute} does not define the summation arithmetic required for all-reduce, we follow the FP8-LM~\cite{peng2023fp8} implementation to adapt MXFPX. The algorithm maintains a parameter $\mu$, initialized as $n$, which controls the scaling factors for quantizing BF16 gradients to MXFPX. In each round, we first compute, on each worker $i$ and for each chunk $j$, the maximum absolute value $m_{i,j}$ of the gradient chunk. These values are all-reduced across workers to obtain the global maximum $gm_j = \max_i(m_{i,j})$. We then determine the global scaling factor of the chunk as $s_j = \mu \cdot gm_j$, such that the original gradient $g_{i,j}$ is quantized as $g'_{i,j} = (g_{i,j} / s_j) \cdot \text{FPX\_MAX}$, where $\text{FPX\_MAX}$ is the largest value representable by MXFPX. The quantized $g'_{i,j}$ is then aggregated via all-reduce. The choice of $\mu$ is critical: a smaller $\mu$ leads to more overflows as $g'_{i,j}$ becomes larger, while a larger $\mu$ causes underflows. We thus adopt the automatic scaling technique proposed in FP8-LM to dynamically update $\mu$. If the overflow ratio exceeds a threshold $\epsilon$, $\mu$ is updated to $2\mu$ in the next training step. Conversely, if the overflow ratio remains smaller than $\gamma$, we decrease $\mu$ to $\gamma\mu$, where $0 < \gamma < 1$ and $\gamma$ is chosen to be close to $1$.

\section{Additional Evaluation Results}\label{appendix:exp_results}

\subp{Zoomed-out end-to-end TTA curves.} While Figures~\ref{fig: e2e-tta}, \ref{fig: e2e-tta-flowcollision}, and \ref{fig: butterfly_mmlu} in the main text present zoomed-in versions of the time-to-accuracy (TTA) curves for ring and butterfly all-reduce, we provide the full zoomed-out versions in Figures~\ref{fig: e2e-tta-full}, \ref{fig: e2e-tta-flowcollision-full}, and \ref{fig: overall-tta-butterfly}, respectively. These full-scale plots illustrate how each method progresses from an initial low accuracy toward the BF16 baseline (depicted by dashed horizontal lines) over time. Furthermore, the full versions confirm that the converged accuracies shown in the zoomed-in plots remain stable over extended periods without further improvement.

\subp{Bandwidth usage over time.} Figure~\ref{fig:time-to-bandwidth} illustrates bandwidth usage over time during the training of LLaMA 1B on MMLU. It distinguishes between active computation during the forward pass and network communication during overlapped backpropagation and gradient aggregation. The curves are periodic, with each period representing one training round. This clearly demonstrates that \sysname effectively reduces the time per training round (improving throughput) by minimizing communication overhead: while the computation intervals remain consistent across BF16, \sysname, and MXFP8, the communication intervals are significantly shortened, indicating faster aggregation.

\subp{Compression error curves over training steps.} We also examine the evolution of vNMSE as training progresses, plotted in Figure~\ref{fig: vnmse-e2e}. The results show that the vNMSE for \sysname and most baselines remains relatively steady, even as gradient distributions evolve during model convergence. Notably, Omnireduce (OR) shows an increase in vNMSE early in training. This suggests that gradient sparsity decreases as training progresses, making OR’s fixed-ratio sparsification less effective and leading to higher compression errors.

\section{Synthesizing Data for Large-scale Simulation}
\label{app:synthetic_data}

\begin{figure}[t!]
    \centering
    \begin{minipage}[t!]{1\linewidth}
        \centering
        \includegraphics[width=\textwidth]
        {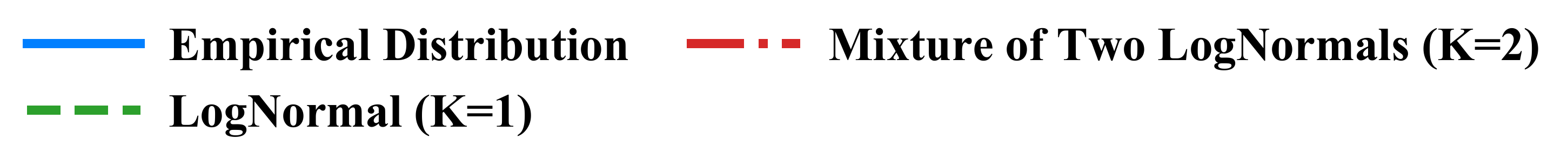}
    \end{minipage}

    \vspace{0.1cm}

    \begin{minipage}[t]{0.8\linewidth}
        \centering
        \includegraphics[width=\textwidth]
        {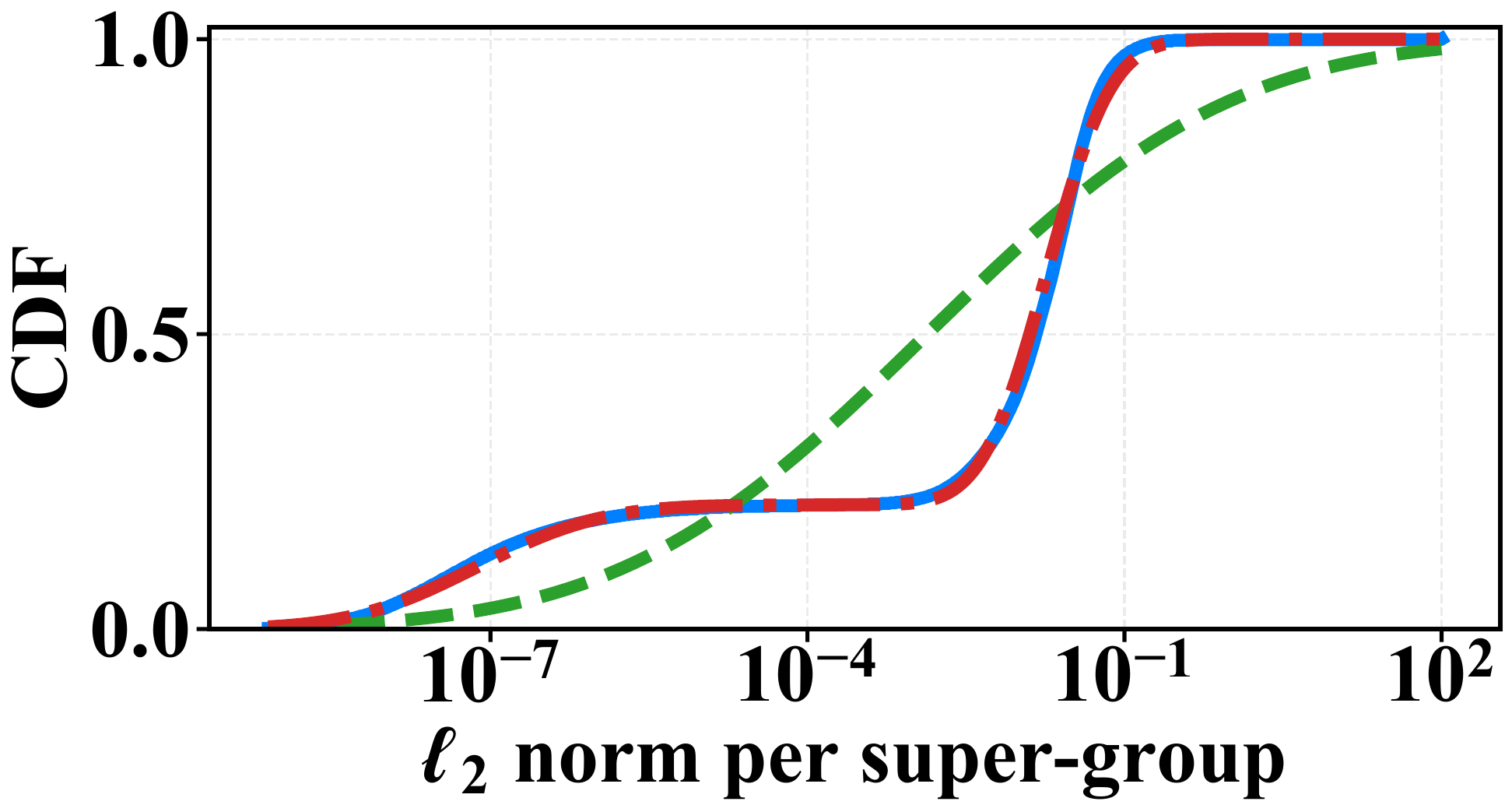}
    \end{minipage}

    \caption{\revise{Empirical distribution of per-super-group $\ell_2$ norms compared with CDFs fitted using mixtures of LogNormal
    distributions. Here for $K=1$: $\pi_1=1, \mu_1=-5.11, \sigma_1=3.14$ and for $K=2$: $\pi_1=0.79, \pi_2 = 0.21, \mu_1=-3.99, \mu_2 = -16.41, \sigma_1=1.22, \sigma_2=4.63$.}}
    \label{fig: real-dist}
\end{figure}

\revise{
We generate the magnitude $M_{\mathcal G}$ of each super-group $\mathcal G$ by fitting a mixture of LogNormal distributions to the empirical distribution of super-group $\ell_2$ norms shown in~\Cref{subfig:demo-original-CDF-per-supergroup-vs-shuffled}. Specifically, for a mixture with $K \in \mathbb{N}^{+}$ components, we define
\begin{equation}
\phi(x;\theta)
=
\sum_{k=1}^{K}
\pi_k \operatorname{LogNormal}(x \mid \mu_k,\sigma_k),
\end{equation}
where
\[
\theta
=
\left(
\{\pi_k\}_{k=1}^{K},
\{\mu_k\}_{k=1}^{K},
\{\sigma_k\}_{k=1}^{K}
\right),
\]
$\pi_k \geq 0$, $\sigma_k>0$, and $\sum_{k=1}^{K}\pi_k=1$. Thus, the mixture has $3K-1$ free parameters.}

\revise{
Because the empirical magnitudes span several orders of magnitude, we fit $\theta$ by minimizing the squared discrepancy between the empirical and fitted CDFs over a logarithmic scale:
\begin{equation}
\label{eq:discrepency}
\min_{\theta}
\int_{\log x_{\min}}^{\log x_{\max}}
\left[
\hat{F}(e^t)-F_{\theta}(e^t)
\right]^2
\,dt,
\end{equation}
where $\hat{F}$ is the empirical CDF, $F_{\theta}$ is the CDF corresponding to $\phi(\cdot;\theta)$, and $[x_{\min},x_{\max}]$ is the observed range of magnitudes.}

\revise{
\Cref{fig: real-dist} compares fits obtained using one and two LogNormal components. The two-component mixture closely reproduces the empirical distribution, so we use $K=2$ and sample $M_{\mathcal G} \sim \phi(\cdot;\theta)$ when generating the synthetic data. }

\end{document}

%% file: wenchen_macros.tex
\newcommand{\subp}{\noindent\textbf}

\newcommand{\ie}{\textit{i.e.}}
\newcommand{\eg}{\textit{e.g.}}

%% file: sigcomm.bib
@article{ben2026quantizing,
  title={Quantizing With Randomized Hadamard Transforms: Efficient Heuristic Now Proven},
  author={Ben-Basat, Ran and Kuszmaul, William and Mitzenmacher, Michael and Portnoy, Amit and Vargaftik, Shay},
  journal={arXiv preprint arXiv:2605.06014},
  year={2026}
}

@article{ben2026note,
  title={A note on turboquant and the earlier drive/eden line of work},
  author={Ben-Basat, Ran and Ben-Itzhak, Yaniv and Mendelson, Gal and Mitzenmacher, Michael and Portnoy, Amit and Vargaftik, Shay},
  journal={arXiv preprint arXiv:2604.18555},
  year={2026}
}

@article{warraich2025optinic,
  title={OptiNIC: A Resilient and Tail-Optimal RDMA NIC for Distributed ML Workloads},
  author={Warraich, Ertza and Imran, Ali and Zulfiqar, Annus and Vargaftik, Shay and Fahmy, Sonia and Shahbaz, Muhammad},
  journal={arXiv preprint arXiv:2512.22743},
  year={2025}
}

@inproceedings{dorfman2023docofl,
  title={DoCoFL: Downlink compression for cross-device federated learning},
  author={Dorfman, Ron and Vargaftik, Shay and Ben-Itzhak, Yaniv and Levy, Kfir Yehuda},
  booktitle={International Conference on Machine Learning},
  pages={8356--8388},
  year={2023},
  organization={PMLR}
}

@article{ben2024optimal,
  title={Optimal and approximate adaptive stochastic quantization},
  author={Ben-Basat, Ran and Ben-Itzhak, Yaniv and Mitzenmacher, Michael and Vargaftik, Shay},
  journal={Advances in Neural Information Processing Systems},
  volume={37},
  pages={94265--94291},
  year={2024}
}

@article{ben2025better,
  title={Better than Optimal: Improving Adaptive Stochastic Quantization Using Shared Randomness},
  author={Ben Basat, Ran and Ben-Itzhak, Yaniv and Mitzenmacher, Michael and Vargaftik, Shay},
  journal={Proceedings of the ACM on Measurement and Analysis of Computing Systems},
  volume={9},
  number={3},
  pages={1--44},
  year={2025},
  publisher={ACM New York, NY, USA}
}

@inproceedings{Hack,
  title={Hack: Homomorphic acceleration via compression of the key-value cache for disaggregated llm inference},
  author={Zhang, Zeyu and Shen, Haiying and Vargaftik, Shay and Basat, Ran Ben and Mitzenmacher, Michael and Yu, Minlan},
  booktitle={Proceedings of the ACM SIGCOMM 2025 Conference},
  pages={1245--1247},
  year={2025}
}

@article{zhao2023pytorch,
  title={Pytorch fsdp: experiences on scaling fully sharded data parallel},
  author={Zhao, Yanli and Gu, Andrew and Varma, Rohan and Luo, Liang and Huang, Chien-Chin and Xu, Min and Wright, Less and Shojanazeri, Hamid and Ott, Myle and Shleifer, Sam and others},
  journal={arXiv preprint arXiv:2304.11277},
  year={2023}
}

@inproceedings{rajbhandari2020zero,
  title={Zero: Memory optimizations toward training trillion parameter models},
  author={Rajbhandari, Samyam and Rasley, Jeff and Ruwase, Olatunji and He, Yuxiong},
  booktitle={SC20: International Conference for High Performance Computing, Networking, Storage and Analysis},
  pages={1--16},
  year={2020},
  organization={IEEE}
}

@InProceedings{pmlr-v258-tseng25a,
  title = 	 {Training LLMs with MXFP4},
  author =       {Tseng, Albert and Yu, Tao and Park, Youngsuk},
  booktitle = 	 {Proceedings of The 28th International Conference on Artificial Intelligence and Statistics},
  pages = 	 {1630--1638},
  year = 	 {2025},
}

@inproceedings{
zhao2024galore,
title={GaLore: Memory-Efficient {LLM} Training by Gradient Low-Rank Projection},
author={Jiawei Zhao and Zhenyu Zhang and Beidi Chen and Zhangyang Wang and Anima Anandkumar and Yuandong Tian},
booktitle={Forty-first International Conference on Machine Learning},
year={2024},
url={https://openreview.net/forum?id=hYHsrKDiX7}
}

@article{thakur2005optimization,
  title={Optimization of collective communication operations in MPICH},
  author={Thakur, Rajeev and Rabenseifner, Rolf and Gropp, William},
  journal={The International Journal of High Performance Computing Applications},
  volume={19},
  number={1},
  pages={49--66},
  year={2005},
  publisher={Sage Publications Sage CA: Thousand Oaks, CA}
}

@article{patarasuk2009bandwidth,
  title={Bandwidth optimal all-reduce algorithms for clusters of workstations},
  author={Patarasuk, Pitch and Yuan, Xin},
  journal={Journal of Parallel and Distributed Computing},
  volume={69},
  number={2},
  pages={117--124},
  year={2009},
  publisher={Elsevier}
}

@inproceedings{warraich2025optireduce,
  title={$\{$OptiReduce$\}$: Resilient and $\{$Tail-Optimal$\}$$\{$AllReduce$\}$ for Distributed Deep Learning in the Cloud},
  author={Warraich, Ertza and Shabtai, Omer and Manaa, Khalid and Vargaftik, Shay and Piasetzky, Yonatan and Kadosh, Matty and Suresh, Lalith and Shahbaz, Muhammad},
  booktitle={22nd USENIX Symposium on Networked Systems Design and Implementation (NSDI 25)},
  pages={685--703},
  year={2025}
}

@inproceedings{wang2023hi,
  title={Hi-Speed DNN Training with Espresso: Unleashing the Full Potential of Gradient Compression with Near-Optimal Usage Strategies},
  author={Wang, Zhuang and Lin, Haibin and Zhu, Yibo and Ng, TS Eugene},
  booktitle={Proceedings of the Eighteenth European Conference on Computer Systems},
  pages={867--882},
  year={2023}
}

@article{topk,
  title={Sparsified SGD with memory},
  author={Stich, Sebastian U and Cordonnier, Jean-Baptiste and Jaggi, Martin},
  journal={Advances in neural information processing systems},
  volume={31},
  year={2018}
}

@article{wang2018atomo,
  title={Atomo: Communication-efficient learning via atomic sparsification},
  author={Wang, Hongyi and Sievert, Scott and Liu, Shengchao and Charles, Zachary and Papailiopoulos, Dimitris and Wright, Stephen},
  journal={Advances in neural information processing systems},
  volume={31},
  year={2018}
}

@inproceedings{kim2019parallax,
  title={Parallax: Sparsity-aware data parallel training of deep neural networks},
  author={Kim, Soojeong and Yu, Gyeong-In and Park, Hojin and Cho, Sungwoo and Jeong, Eunji and Ha, Hyeonmin and Lee, Sanha and Jeong, Joo Seong and Chun, Byung-Gon},
  booktitle={Proceedings of the Fourteenth EuroSys Conference 2019},
  pages={1--15},
  year={2019}
}

@article{vargaftik2021drive,
  title={Drive: One-bit distributed mean estimation},
  author={Vargaftik, Shay and Ben-Basat, Ran and Portnoy, Amit and Mendelson, Gal and Ben-Itzhak, Yaniv and Mitzenmacher, Michael},
  journal={Advances in Neural Information Processing Systems},
  volume={34},
  pages={362--377},
  year={2021}
}

@article{alistarh2017qsgd,
  title={QSGD: Communication-efficient SGD via gradient quantization and encoding},
  author={Alistarh, Dan and Grubic, Demjan and Li, Jerry and Tomioka, Ryota and Vojnovic, Milan},
  journal={Advances in neural information processing systems},
  volume={30},
  year={2017}
}

@inproceedings{bernstein2018signsgd,
  title={signSGD: Compressed optimisation for non-convex problems},
  author={Bernstein, Jeremy and Wang, Yu-Xiang and Azizzadenesheli, Kamyar and Anandkumar, Animashree},
  booktitle={International Conference on Machine Learning},
  pages={560--569},
  year={2018},
  organization={PMLR}
}

@article{mee,
  title={Ice buckets: Improved counter estimation for network measurement},
  author={Einziger, Gil and Fellman, Benny and Friedman, Roy and Kassner, Yaron},
  journal={IEEE/ACM Transactions on Networking},
  volume={26},
  number={3},
  pages={1165--1178},
  year={2018},
  publisher={IEEE}
}

@Misc{memory-bound,
    year={2020},
    title={Accelerating HPC Applications with NVIDIA Nsight Compute Roofline Analysis},
    howpublished={\url{https://developer.nvidia.com/blog/accelerating-hpc-applications-with-nsight-compute-roofline-analysis}}
}

@Misc{bfloat16,
  year         = {2019},
  title        = {BFloat16: The secret to high performance on Cloud TPUs},
  howpublished = {\url{https://cloud.google.com/blog/products/ai-machine-learning/bfloat16-the-secret-to-high-performance-on-cloud-tpus}}
}

@inproceedings{cao2024crux,
  title={Crux: Gpu-efficient communication scheduling for deep learning training},
  author={Cao, Jiamin and Guan, Yu and Qian, Kun and Gao, Jiaqi and Xiao, Wencong and Dong, Jianbo and Fu, Binzhang and Cai, Dennis and Zhai, Ennan},
  booktitle={Proceedings of the ACM SIGCOMM 2024 Conference},
  pages={1--15},
  year={2024}
}

@inproceedings{panferovquest,
  title={QuEST: Training Accurate LLMs over Highly-Compressed Weights and Activation},
  author={Panferov, Andrei and Chen, Jiale and Tabesh, Soroush and Castro, Roberto L and Nikdan, Mahdi and Alistarh, Dan},
  booktitle={Sparsity in LLMs (SLLM): Deep Dive into Mixture of Experts, Quantization, Hardware, and Inference}
}

@inproceedings{10.5555/3692070.3694598,
author = {Zhao, Jiawei and Zhang, Zhenyu and Chen, Beidi and Wang, Zhangyang and Anandkumar, Anima and Tian, Yuandong},
title = {GaLore: memory-efficient LLM training by gradient low-rank projection},
year = {2024},
publisher = {JMLR.org},
booktitle = {Proceedings of the 41st International Conference on Machine Learning},
articleno = {2528},
numpages = {23},
location = {Vienna, Austria},
series = {ICML'24}
}

@Misc{blackwell,
    year={2025},
    title = "NVIDIA Blackwell Architecture
Technical Brief",
    url={https://resources.nvidia.com/en-us-blackwell-architecture}
}

@misc{open-source,
  author       = {Wenchen Han},
  title        = {Github code repository of DynamiQ artifact},
  year         = {2026},
  howpublished = {\url{https://github.com/CharlesHan24/DynamiQ}}
}

@inproceedings{hwang2021elastic,
  title={Elastic resource sharing for distributed deep learning},
  author={Hwang, Changho and Kim, Taehyun and Kim, Sunghyun and Shin, Jinwoo and Park, KyoungSoo},
  booktitle={18th USENIX Symposium on Networked Systems Design and Implementation (NSDI 21)},
  pages={721--739},
  year={2021}
}

@Misc{memory-bound2,
    year={2023},
    title={Mastering LLM Techniques: Inference Optimization},
    howpublished={\url{https://developer.nvidia.com/blog/mastering-llm-techniques-inference-optimization/}}
}

@article{castro2025quartet,
  title={Quartet: Native FP4 Training Can Be Optimal for Large Language Models},
  author={Castro, Roberto L and Panferov, Andrei and Tabesh, Soroush and Sieberling, Oliver and Chen, Jiale and Nikdan, Mahdi and Ashkboos, Saleh and Alistarh, Dan},
  journal={arXiv preprint arXiv:2505.14669},
  year={2025}
}

@Misc{kernel-fusion,
    year={2025},
    title={Advanced NVIDIA CUDA Kernel Optimization Techniques: Handwritten PTX},
    howpublished={\url{https://developer.nvidia.com/blog/advanced-nvidia-cuda-kernel-optimization-techniques-handwritten-ptx/}}
}

@inproceedings{wang2025optimizing,
  title={Optimizing large language model training using fp4 quantization},
  author={Wang, Ruizhe and Gong, Yeyun and Liu, Xiao and Zhao, Guoshuai and Yang, Ziyue and Guo, Baining and Zha, Zhengjun and Cheng, Peng},
  booktitle={Forty-second International Conference on Machine Learning},
  year={2025}
}

@article{efficientqat,
  title={EfficientQAT: Efficient Quantization-Aware Training for Large Language Models},
  author={Chen, Mengzhao and Shao, Wenqi and Xu, Peng and Wang, Jiahao and Gao, Peng and Zhang, Kaipeng and Qiao, Yu and Luo, Ping},
  journal={arXiv preprint arXiv:2407.11062},
  year={2024}
}

@misc{nvfp4,
    year={2025},
title={NVFP4 Trains with Precision of 16-Bit and Speed and Efficiency of 4-Bit},
howpublished={\url{https://developer.nvidia.com/blog/nvfp4-trains-with-precision-of-16-bit-and-speed-and-efficiency-of-4-bit/}}
}

@misc{why-elementwise-memory-bound,
    year={2023},
title={Memory-Limited Layers User's Guide},
howpublished={\url{https://docs.nvidia.com/deeplearning/performance/dl-performance-memory-limited/index.html}}
}

@article{gholami2024ai,
  title={Ai and memory wall},
  author={Gholami, Amir and Yao, Zhewei and Kim, Sehoon and Hooper, Coleman and Mahoney, Michael W and Keutzer, Kurt},
  journal={IEEE Micro},
  volume={44},
  number={3},
  pages={33--39},
  year={2024},
  publisher={IEEE}
}

@inproceedings{li2024thc,
  title={$\{$THC$\}$: Accelerating Distributed Deep Learning Using Tensor Homomorphic Compression},
  author={Li, Minghao and Basat, Ran Ben and Vargaftik, Shay and Lao, ChonLam and Xu, Kevin and Mitzenmacher, Michael and Yu, Minlan},
  booktitle={21st USENIX Symposium on Networked Systems Design and Implementation (NSDI 24)},
  pages={1191--1211},
  year={2024}
}

@misc{tang2025dreamddp,
      title={DreamDDP: Accelerating Data Parallel Distributed LLM Training with Layer-wise Scheduled Partial Synchronization}, 
      author={Zhenheng Tang and Zichen Tang and Junlin Huang and Xinglin Pan and Rudan Yan and Yuxin Wang and Amelie Chi Zhou and Shaohuai Shi and Xiaowen Chu and Bo Li},
      year={2025},
      eprint={2502.11058},
      archivePrefix={arXiv},
      primaryClass={cs.DC},
      url={https://arxiv.org/abs/2502.11058}, 
}

@article{powersgd,
  title={PowerSGD: Practical low-rank gradient compression for distributed optimization},
  author={Vogels, Thijs and Karimireddy, Sai Praneeth and Jaggi, Martin},
  journal={Advances in Neural Information Processing Systems},
  volume={32},
  year={2019}
}

@article{on-the-utility,
  title={On the utility of gradient compression in distributed training systems},
  author={Agarwal, Saurabh and Wang, Hongyi and Venkataraman, Shivaram and Papailiopoulos, Dimitris},
  journal={Proceedings of Machine Learning and Systems},
  volume={4},
  pages={652--672},
  year={2022}
}

@misc{ring,
    year={2017},
    title =        {Ring all reduce.},
    howpublished={\url{https://github.com/baidu-research/baidu-allreduce}}
    
}

@misc{topology-at-scale,
    year={2024},
    title = {How Meta trains large language models at scale},
    howpublished={\url{https://engineering.fb.com/2024/06/12/data-infrastructure/training-large-language-models-at-scale-meta/}}
}

@inproceedings{hammingmesh2022,
author = {Hoefler, Torsten and Bonato, Tommaso and De Sensi, Daniele and Di Girolamo, Salvatore and Li, Shigang and Heddes, Marco and Belk, Jon and Goel, Deepak and Castro, Miguel and Scott, Steve},
title = {HammingMesh: a network topology for large-scale deep learning},
year = {2022},
isbn = {9784665454445},
publisher = {IEEE Press},
booktitle = {Proceedings of the International Conference on High Performance Computing, Networking, Storage and Analysis},
articleno = {11},
numpages = {18},
location = {Dallas, Texas},
series = {SC '22}
}

@article{lee2024fp8,
  title={To FP8 and Back Again: Quantifying Reduced Precision Effects on LLM Training Stability},
  author={Lee, Joonhyung and Bae, Jeongin and Kim, Byeongwook and Kwon, Se Jung and Lee, Dongsoo},
  journal={arXiv preprint arXiv:2405.18710},
  year={2024}
}

@inproceedings{adam2014method,
  title={A method for stochastic optimization},
  author={Kinga, Diederik and Adam, Jimmy Ba and others},
  booktitle={International conference on learning representations (ICLR)},
  volume={5},
  number={6},
  year={2015},
  organization={California;}
}

@article{loshchilov2017decoupled,
  title={Decoupled weight decay regularization},
  author={Loshchilov, Ilya and Hutter, Frank},
  journal={arXiv preprint arXiv:1711.05101},
  year={2017}
}

@misc{collective-nccl,
    year={2024},
    title={Collective operations in NCCL.}, 
    howpublished={\url{https://docs.nvidia.com/deeplearning/nccl/user-guide/docs/usage/collectives.html}}
}

@misc{jax,
    year={2020},
    title={JAX: High performance array computing}, howpublished={\url={https://docs.jax.dev/en/latest/index.html}}
}

@inproceedings{li2014scaling,
  title={Scaling distributed machine learning with the parameter server},
  author={Li, Mu and Andersen, David G and Park, Jun Woo and Smola, Alexander J and Ahmed, Amr and Josifovski, Vanja and Long, James and Shekita, Eugene J and Su, Bor-Yiing},
  booktitle={11th USENIX Symposium on operating systems design and implementation (OSDI 14)},
  pages={583--598},
  year={2014}
}

@inproceedings{jiang2020unified,
  title={A unified architecture for accelerating distributed $\{$DNN$\}$ training in heterogeneous $\{$GPU/CPU$\}$ clusters},
  author={Jiang, Yimin and Zhu, Yibo and Lan, Chang and Yi, Bairen and Cui, Yong and Guo, Chuanxiong},
  booktitle={14th USENIX Symposium on Operating Systems Design and Implementation (OSDI 20)},
  pages={463--479},
  year={2020}
}

@inproceedings{fei2021efficient,
  title={Efficient sparse collective communication and its application to accelerate distributed deep learning},
  author={Fei, Jiawei and Ho, Chen-Yu and Sahu, Atal N and Canini, Marco and Sapio, Amedeo},
  booktitle={Proceedings of the 2021 ACM SIGCOMM 2021 Conference},
  pages={676--691},
  year={2021}
}

@inproceedings{bai2021gradient,
  title={Gradient compression supercharged high-performance data parallel dnn training},
  author={Bai, Youhui and Li, Cheng and Zhou, Quan and Yi, Jun and Gong, Ping and Yan, Feng and Chen, Ruichuan and Xu, Yinlong},
  booktitle={Proceedings of the ACM SIGOPS 28th Symposium on Operating Systems Principles},
  pages={359--375},
  year={2021}
}

@misc{grattafiori2024llama3herdmodels,
      title={The Llama 3 Herd of Models}, 
      author={Grattafiori et al.},
      year={2024},
      eprint={2407.21783},
      archivePrefix={arXiv},
      primaryClass={cs.AI},
      url={https://arxiv.org/abs/2407.21783}, 
}

@misc{nccl,
year={2024},
title={NVIDIA Collective Communications Library (NCCL).}, howpublished={\url{https://developer.nvidia.com/nccl}}
}

@misc{GGUF,
year={2024},
title={GGUF format}, howpublished={\url{https://github.com/ggml-org/ggml/blob/master/docs/gguf.md}}
}

@article{li2020pytorch,
  title={Pytorch distributed: Experiences on accelerating data parallel training},
  author={Li, Shen and Zhao, Yanli and Varma, Rohan and Salpekar, Omkar and Noordhuis, Pieter and Li, Teng and Paszke, Adam and Smith, Jeff and Vaughan, Brian and Damania, Pritam and others},
  journal={arXiv preprint arXiv:2006.15704},
  year={2020}
}

@article{devlin2018bert,
  title={Bert: Pre-training of deep bidirectional transformers for language understanding},
  author={Devlin, Jacob and Chang, Ming-Wei and Lee, Kenton and Toutanova, Kristina},
  journal={arXiv preprint arXiv:1810.04805},
  year={2018}
}

@article{wikitext,
  title={Pointer sentinel mixture models},
  author={Merity, Stephen and Xiong, Caiming and Bradbury, James and Socher, Richard},
  journal={arXiv preprint arXiv:1609.07843},
  year={2016}
}

@article{terngrad,
  title={Terngrad: Ternary gradients to reduce communication in distributed deep learning},
  author={Wen, Wei and Xu, Cong and Yan, Feng and Wu, Chunpeng and Wang, Yandan and Chen, Yiran and Li, Hai},
  journal={Advances in neural information processing systems},
  volume={30},
  year={2017}
}

@article{hedayat1978hadamard,
  title={Hadamard matrices and their applications},
  author={Hedayat, A and Wallis, Walter Dennis},
  journal={The annals of statistics},
  pages={1184--1238},
  year={1978},
  publisher={JSTOR}
}

@inproceedings{wang2023cupcake,
  title={CUPCAKE: A Compression Optimizer for Scalable Communication-efficient Distributed Training},
  author={Wang, Zhuang and Wu, Xinyu Crystal and Xu, Zhaozhuo and Ng, TS Eugene},
  booktitle={Proceedings of the Sixth Conference on Machine Learning and Systems (MLSys' 23)},
  year={2023},
  organization={Proceedings of the Sixth Conference on Machine Learning and Systems (MLSys' 23)}
}

@article{m2021efficient,
  title={An efficient statistical-based gradient compression technique for distributed training systems},
  author={M Abdelmoniem, Ahmed and Elzanaty, Ahmed and Alouini, Mohamed-Slim and Canini, Marco},
  journal={Proceedings of Machine Learning and Systems},
  volume={3},
  pages={297--322},
  year={2021}
}

@article{ding2023enhancing,
  title={Enhancing Chat Language Models by Scaling High-quality Instructional Conversations},
  author={Ding, Ning and Chen, Yulin and Xu, Bokai and Qin, Yujia and Zheng, Zhi and Hu, Shengding and Liu, Zhiyuan and Sun, Maosong and Zhou, Bowen},
  journal={arXiv preprint arXiv:2305.14233},
  year={2023}
}

@article{hendrycks2020measuring,
  title={Measuring massive multitask language understanding},
  author={Hendrycks, Dan and Burns, Collin and Basart, Steven and Zou, Andy and Mazeika, Mantas and Song, Dawn and Steinhardt, Jacob},
  journal={arXiv preprint arXiv:2009.03300},
  year={2020}
}

@inproceedings{NIPS2012_6aca9700,
 author = {Dean, Jeffrey and Corrado, Greg and Monga, Rajat and Chen, Kai and Devin, Matthieu and Mao, Mark and Ranzato, Marc\textquotesingle aurelio and Senior, Andrew and Tucker, Paul and Yang, Ke and Le, Quoc and Ng, Andrew},
 booktitle = {Advances in Neural Information Processing Systems},
 editor = {F. Pereira and C.J. Burges and L. Bottou and K.Q. Weinberger},
 pages = {},
 publisher = {Curran Associates, Inc.},
 title = {Large Scale Distributed Deep Networks},
 url = {https://proceedings.neurips.cc/paper_files/paper/2012/file/6aca97005c68f1206823815f66102863-Paper.pdf},
 volume = {25},
 year = {2012}
}

@inproceedings{sapio2021scaling,
  title={Scaling distributed machine learning with $\{$In-Network$\}$ aggregation},
  author={Sapio, Amedeo and Canini, Marco and Ho, Chen-Yu and Nelson, Jacob and Kalnis, Panos and Kim, Changhoon and Krishnamurthy, Arvind and Moshref, Masoud and Ports, Dan and Richt{\'a}rik, Peter},
  booktitle={18th USENIX Symposium on Networked Systems Design and Implementation (NSDI 21)},
  pages={785--808},
  year={2021}
}

@inproceedings{karimireddy2019error,
  title={Error feedback fixes signsgd and other gradient compression schemes},
  author={Karimireddy, Sai Praneeth and Rebjock, Quentin and Stich, Sebastian and Jaggi, Martin},
  booktitle={International Conference on Machine Learning},
  pages={3252--3261},
  year={2019},
  organization={PMLR}
}

@article{li2024accelerating,
  title={Accelerating Distributed Deep Learning using Lossless Homomorphic Compression},
  author={Li, Haoyu and Xu, Yuchen and Chen, Jiayi and Dwivedula, Rohit and Wu, Wenfei and He, Keqiang and Akella, Aditya and Kim, Daehyeok},
  journal={arXiv preprint arXiv:2402.07529},
  year={2024}
}

@inproceedings{vargaftik2022eden,
  title={Eden: Communication-efficient and robust distributed mean estimation for federated learning},
  author={Vargaftik, Shay and Basat, Ran Ben and Portnoy, Amit and Mendelson, Gal and Itzhak, Yaniv Ben and Mitzenmacher, Michael},
  booktitle={International Conference on Machine Learning},
  pages={21984--22014},
  year={2022},
  organization={PMLR}
}

@inproceedings{benaccelerating,
  title={Accelerating Federated Learning with Quick Distributed Mean Estimation},
  author={Ben-Basat, Ran and Portnoy, Amit and Einziger, Gil and Ben-Itzhak, Yaniv and Mitzenmacher, Michael},
  booktitle={ICML},
  year={2024}
}

@article{wang2018glue,
  title={GLUE: A multi-task benchmark and analysis platform for natural language understanding},
  author={Wang, Alex and Singh, Amanpreet and Michael, Julian and Hill, Felix and Levy, Omer and Bowman, Samuel R},
  journal={arXiv preprint arXiv:1804.07461},
  year={2018}
}

@article{jiao2019tinybert,
  title={Tinybert: Distilling bert for natural language understanding},
  author={Jiao, Xiaoqi and Yin, Yichun and Shang, Lifeng and Jiang, Xin and Chen, Xiao and Li, Linlin and Wang, Fang and Liu, Qun},
  journal={arXiv preprint arXiv:1909.10351},
  year={2019}
}

@inproceedings{ben2020send,
  title={How to Send a Real Number Using a Single Bit (And Some Shared Randomness)},
  author={Ben Basat, Ran and Mitzenmacher, Michael and Vargaftik, Shay},
  booktitle={48th International Colloquium on Automata, Languages, and Programming (ICALP 2021)},
  year={2021},
}

@inproceedings{chen2024justintime,
  title={{When ML Training Cuts Through Congestion: Just-in-Time Gradient Compression via Packet Trimming}},
  author={ Chen, Xiaoqi and Vargaftik, Shay and Ben-Basat, Ran},
  booktitle={Hotnets},
  year={2024}
}

@misc{opencompute,
year={2023},
title={OCP Microscaling Formats (MX) Specification}, howpublished={\url{https://www.opencompute.org/documents/ocp-microscaling-formats-mx-v1-0-spec-final-pdf}
}
}

@article{rouhani2023microscaling,
  title={Microscaling data formats for deep learning},
  author={Rouhani, Bita Darvish and Zhao, Ritchie and More, Ankit and Hall, Mathew and Khodamoradi, Alireza and Deng, Summer and Choudhary, Dhruv and Cornea, Marius and Dellinger, Eric and Denolf, Kristof and others},
  journal={arXiv preprint arXiv:2310.10537},
  year={2023}
}

@article{peng2023fp8,
  title={Fp8-lm: Training fp8 large language models},
  author={Peng, Houwen and Wu, Kan and Wei, Yixuan and Zhao, Guoshuai and Yang, Yuxiang and Liu, Ze and Xiong, Yifan and Yang, Ziyue and Ni, Bolin and Hu, Jingcheng and others},
  journal={arXiv preprint arXiv:2310.18313},
  year={2023}
}

@article{team2025gemma,
  title={Gemma 3 technical report},
  author={Team, Gemma and Kamath, Aishwarya and Ferret, Johan and Pathak, Shreya and Vieillard, Nino and Merhej, Ramona and Perrin, Sarah and Matejovicova, Tatiana and Ram{\'e}, Alexandre and Rivi{\`e}re, Morgane and others},
  journal={arXiv preprint arXiv:2503.19786},
  year={2025}
}

@article{lee2023training,
  title={Training with mixed-precision floating-point assignments},
  author={Lee, Wonyeol and Sharma, Rahul and Aiken, Alex},
  journal={arXiv preprint arXiv:2301.13464},
  year={2023}
}

@inproceedings{suresh2022correlated,
  title={Correlated quantization for distributed mean estimation and optimization},
  author={Suresh, Ananda Theertha and Sun, Ziteng and Ro, Jae and Yu, Felix},
  booktitle={International Conference on Machine Learning},
  pages={20856--20876},
  year={2022},
  organization={PMLR}
}

@inproceedings {desensi2024swing,
author = {Daniele De Sensi and Tommaso Bonato and David Saam and Torsten Hoefler},
title = {Swing: Short-cutting Rings for Higher Bandwidth Allreduce},
booktitle = {21st USENIX Symposium on Networked Systems Design and Implementation (NSDI 24)},
year = {2024},
isbn = {978-1-939133-39-7},
address = {Santa Clara, CA},
pages = {1445--1462},
url = {https://www.usenix.org/conference/nsdi24/presentation/de-sensi},
publisher = {USENIX Association},
month = apr
}

@inproceedings{hotnets2499problems,
author = {Gherghescu, Alexandru M. and B\u{a}doiu, Vlad-Andrei and Agache, Alexandru and Dumitru, Mihai-Valentin and Vasilescu, Iuliu and Mantu, Radu and Raiciu, Costin},
title = {I've Got 99 Problems But FLOPS Ain't One},
year = {2024},
isbn = {9798400712722},
publisher = {Association for Computing Machinery},
address = {New York, NY, USA},
url = {https://doi.org/10.1145/3696348.3696893},
doi = {10.1145/3696348.3696893},
booktitle = {Proceedings of the 23rd ACM Workshop on Hot Topics in Networks},
pages = {195–204},
numpages = {10},
location = {Irvine, CA, USA},
series = {HotNets '24}
}

@inproceedings{han24hotnets,
author = {Han, Wenchen and Vargaftik, Shay and Mitzenmacher, Michael and Karp, Brad and Basat, Ran Ben},
title = {Beyond Throughput and Compression Ratios: Towards High End-to-end Utility of Gradient Compression},
year = {2024},
isbn = {9798400712722},
publisher = {Association for Computing Machinery},
address = {New York, NY, USA},
url = {https://doi.org/10.1145/3696348.3696857},
doi = {10.1145/3696348.3696857},
booktitle = {Proceedings of the 23rd ACM Workshop on Hot Topics in Networks},
pages = {186–194},
numpages = {9},
location = {Irvine, CA, USA},
series = {HotNets '24}
}

@inproceedings{narayanan2021efficient,
  author    = {Deepak Narayanan and Mohammad Shoeybi and Jared Casper and Patrick LeGresley and Mostofa Patwary and Vijay Korthikanti and Dmitri Vainbrand and Prethvi Kashinkunti and Julie Bernauer and Bryan Catanzaro and Amar Phanishayee and Matei Zaharia},
  title     = {Efficient Large-Scale Language Model Training on {GPU} Clusters Using {Megatron-LM}},
  booktitle = {Proceedings of the International Conference for High Performance Computing, Networking, Storage and Analysis},
  series    = {SC '21},
  year      = {2021},
  publisher = {Association for Computing Machinery},
  doi       = {10.1145/3458817.3476209}
}

@article{llama3herd,
  author        = {{Llama Team, AI @ Meta}},
  title         = {The Llama 3 Herd of Models},
  journal       = {arXiv preprint arXiv:2407.21783},
  year          = {2024},
  eprint        = {2407.21783},
  archivePrefix = {arXiv},
  primaryClass  = {cs.AI},
  doi           = {10.48550/arXiv.2407.21783}
}

@inproceedings{chu2025scalingllama3,
  author    = {Weiwei Chu and Xinfeng Xie and Jiecao Yu and Jie Wang and Amar Phanishayee and Chunqiang Tang and Yuchen Hao and Jianyu Huang and Mustafa Ozdal and Jun Wang and Vedanuj Goswami and Naman Goyal and Abhishek Kadian and Andrew Gu and Chris Cai and Feng Tian and Xiaodong Wang and Min Si and Pavan Balaji and Ching-Hsiang Chu and Jongsoo Park},
  title     = {Scaling Llama 3 Training with Efficient Parallelism Strategies},
  booktitle = {Proceedings of the 52nd Annual International Symposium on Computer Architecture},
  series    = {ISCA '25},
  year      = {2025},
  publisher = {Association for Computing Machinery},
  doi       = {10.1145/3695053.3731410}
}

@inproceedings{fishman2025scalingfp8,
  title     = {Scaling FP8 Training to Trillion-Token LLMs},
  author    = {Fishman, Maxim and Chmiel, Brian and Banner, Ron and Soudry, Daniel},
  booktitle = {International Conference on Learning Representations},
  year      = {2025}
}

@misc{fujii2024balancing,
  title         = {Balancing Speed and Stability: The Trade-offs of FP8 vs. BF16 Training in LLMs},
  author        = {Fujii, Kazuki and Nakamura, Taishi and Yokota, Rio},
  year          = {2024},
  eprint        = {2411.08719},
  archivePrefix = {arXiv},
  primaryClass  = {cs.LG}
}

@article{shazeer2017outrageously,
  title={Outrageously large neural networks: The sparsely-gated mixture-of-experts layer},
  author={Shazeer, Noam and Mirhoseini, Azalia and Maziarz, Krzysztof and Davis, Andy and Le, Quoc and Hinton, Geoffrey and Dean, Jeff},
  journal={arXiv preprint arXiv:1701.06538},
  year={2017}
}

@misc{nvidia_gb200_nvl72,
  title        = {{NVIDIA GB200 NVL72}},
  author       = {{NVIDIA}},
  howpublished = {NVIDIA Data Center Product Page},
  year         = {2026},
  url          = {https://www.nvidia.com/en-us/data-center/gb200-nvl72/},
  note         = {Accessed: 2026-06-17}
}

@misc{nvidia_connectx7_400g_adapters,
  title        = {{NVIDIA ConnectX-7 400G Adapters}: Datasheet},
  author       = {{NVIDIA}},
  howpublished = {NVIDIA Datasheet},
  year         = {2026},
  url          = {https://www.nvidia.com/content/dam/en-zz/Solutions/networking/infiniband/connectx-7-datasheet.pdf},
  note         = {Accessed: 2026-06-17}
}

@misc{nvidia_connectx7_user_manual,
  title        = {{NVIDIA ConnectX-7 Adapter Cards User Manual}},
  author       = {{NVIDIA}},
  howpublished = {NVIDIA Networking Documentation},
  year         = {2026},
  url          = {https://networking-docs.nvidia.com/connectx7hw},
  note         = {Accessed: 2026-06-17}
}
